\documentclass{article}
\usepackage{graphicx}
\usepackage[numbers]{natbib}

\usepackage{booktabs}
\usepackage{subcaption}
\usepackage{cprotect}
\usepackage{amssymb,amsmath,amsthm,mathtools}
\usepackage{epstopdf}
\usepackage{bbm}
\usepackage[marginratio=1:1,height=600pt,width=460pt,tmargin=100pt]{geometry}
\usepackage{layout, color}
\usepackage{bbm}
\usepackage{enumerate}
\usepackage{authblk}
\usepackage{algorithm}
\usepackage{algpseudocode}
\usepackage{setspace}
\usepackage{hyperref}
\usepackage{url}
\usepackage{mdwlist}
\usepackage{multirow}
\usepackage{amsthm}
\usepackage{color}
\newcommand{\R}{\mathbb{R}}

\newcommand{\calN}{\mathcal{N}}

\newtheorem{theorem}{Theorem}[section]
\newtheorem{proposition}[theorem]{Proposition}
\newtheorem{assumption}{Assumption}

\newtheorem{lemma}[theorem]{Lemma}
\newtheorem{definition}[theorem]{Definition}

\theoremstyle{remark}
\newtheorem{remark}{Remark}
\newtheorem{example}{Example}
\newtheorem{conjecture*}{Conjecture}
\theoremstyle{plain}

\usepackage{xcolor}

\newcommand{\Sigmat}{\Sigma_{t}}

\newcommand{\B}{\mathcal{B}}

\newcommand{\EE}{\mathbb{E}}

\newcommand{\omegatd}{\omega_t^{-d}}

\newcommand{\Femp}{F_{1:N}}

\newcommand{\bTheta}{\boldsymbol \Theta}

\newcommand{\ciNNI}{cINN-MMD}
\newcommand{\ciNNII}{cINN-Flow}
\newcommand{\ciNNIIplus}{cINN-Flow+}

\newlength{\tempdima}
\newcommand{\rowname}[1]
{\rotatebox{90}{\makebox[\tempdima][c]{\textbf{#1}}}}
\renewcommand{\thesubfigure}{\alph{subfigure}}
\newcommand{\mycaption}[1]
{\refstepcounter{subfigure}\textbf{(\thesubfigure) }{\ignorespaces #1}}

\usepackage{array}
\newcolumntype{P}[1]{>{\centering\arraybackslash}p{#1}}

\title{Invertible Neural Networks for Graph Prediction}

\author[1]{Chen Xu
}
\author[2]{Xiuyuan Cheng
}
\author[1]{Yao Xie\footnote{Email: yao.xie@isye.gatech.edu}}
\affil[1]{{\small H. Milton Stewart School of Industrial and Systems Engineering, Georgia Institute of Technology}}
\affil[2]{{\small Department of Mathematics, Duke University}}

\date{}

\newcommand{\rev}[1]{{\color{black}#1}}

\begin{document}
\maketitle

\begin{abstract}
Graph prediction problems prevail in data analysis and machine learning. The inverse prediction problem, namely to infer input data from given output labels, is of emerging interest in various applications. 
In this work, we develop \textit{invertible graph neural network} (iGNN),
a deep generative model to tackle the inverse prediction problem on graphs by casting it as a conditional generative task. 
The proposed model consists of an invertible sub-network that maps one-to-one from data to an intermediate encoded feature, 
which allows forward prediction by a linear classification sub-network as well as efficient generation from output labels via a parametric mixture model.
The invertibility of the encoding sub-network is ensured by a Wasserstein-2 regularization which allows free-form layers in the residual blocks. The model is scalable to large graphs by a factorized parametric mixture model of the encoded feature and is computationally scalable by using GNN layers.
The existence of invertible flow mapping is backed by theories of optimal transport and diffusion process, and we prove the expressiveness of graph convolution layers to approximate the theoretical flows of graph data. 
The proposed iGNN model is experimentally examined on synthetic data, including the example on large graphs, and the empirical advantage is also demonstrated on real-application datasets of solar ramping event data and traffic flow anomaly detection.
\end{abstract}

\section{Introduction}

Graph prediction is an important topic in statistics, machine learning, and signal processing, 
and is motivated by various applications, e.g., protein-protein interaction networks \cite{yang2020graph}, wind power prediction \cite{yu2020superposition}, and user behavior modeling in social networks \cite{beutel2015graph}. 
There can be various versions of graph prediction problems.
We are interested in the scenario where one observes nodal (multi-dimensional) features $X$, nodal responses $Y$, and graph topology information.
The prediction algorithm typically will then leverage graph topology information to facilitate the learning of predicting $Y$ given $X$.
In practice, Graph Neural Network (GNN) models \cite{wu2020comprehensive} have become a popular choice due to their expressiveness power and scalable computation. 
In the so-called inverse of a graph prediction problem, one would like to infer the input graph nodal features $X$ given an outcome response $Y$. 
Such a problem is of interest in various real-world applications,
 for instance, in molecular design \citep{SnchezLengeling2018InverseMD}, scientists want to infer features of the molecule that lead to certain outcomes; 
 in power outage analysis \citep{Zong2018DeepAG,AlShaalan2020ReliabilityEO}, we are interested in finding features (e.g., weather or past power output by sensors) that leads to outages for future prevention.
The inverse graph prediction problem is the focus of the current work, for which we develop an invertible deep model that can be efficiently applied to graph data. 
 
\begin{figure}[t]
    \centering
    \includegraphics[width=0.9 \linewidth]{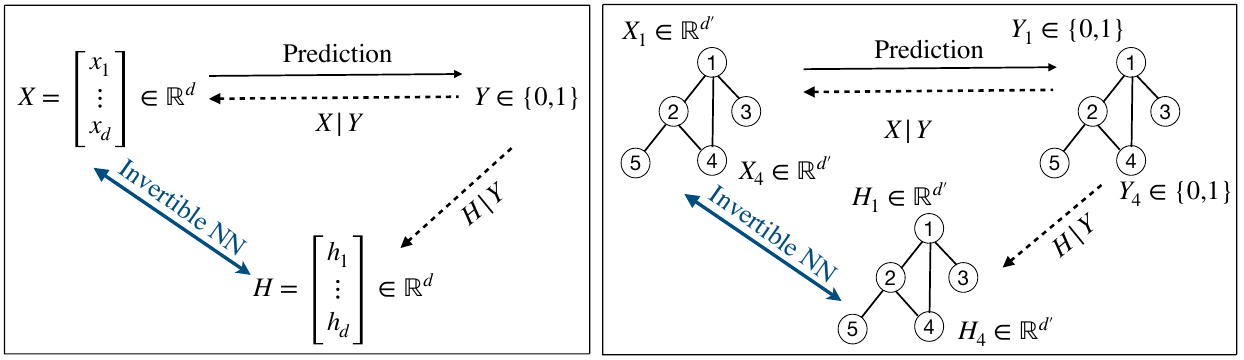}
    \caption{ 
Illustration of the proposed iGNN model applied to general data (left) and graph data (right).
On graph data, for $v \in \{1, \cdots, 5 \}$, 
$X_v$ is the nodal feature, and $Y_v$ is the nodal label.
The inverse prediction problem is to generate the conditional distribution $X|Y$, which is a one-to-many mapping from $Y$ to $X$ (indicated by dash lines). 
In our approach, $Y$ is first mapped to an intermediate feature $H|Y$ (one-to-many) and then through an invertible neural network to $X|Y$ (one-to-one). 
}
\label{motivation}
\end{figure}

The prediction and inverse prediction problems are illustrated in Fig. \ref{motivation}.
The left plot shows the case of general non-graph data, and the right plot shows the case with graph data $X$ and graph label $Y$. 
As an illustrative example, 
suppose the graph is a power grid, where each node $v \in \{1,\ldots,5\}$ denotes a power generator, 
$X_v \in \R^{d'}$ denotes historical power outputs from generator $v$, 
and $Y_v$ denotes the status of the generator $v$
(e.g., $Y_v=0$ for functioning normally, and 1 for anomaly). 
For system monitoring, it is useful to 
predict the status of generators given historical observations (the forward prediction, indicated by solid black arrows),
as well as to generate unobserved possible circumstances  given generator status (the inverse, indicated by dashed black arrows)
for cause analyses.

To formalize the notion of inverse prediction, we adopt some probability terminologies. 
The forward prediction problem is to predict $Y$ from input $X$, which can be formulated as learning the conditional probability of $p(Y|X)$.
The inverse prediction problem is to learn the conditional probability of $p(X|Y)$ and to {\it generate} samples $X$ from it. 
Note that a discriminative task seeks to estimate the posterior $p(X|Y)$ at a given (test) point $X$, which can be done by a conventional classification model. The generative task is different in that one asks to sample $X$ according to $p(X|Y)$, that is, given a label $Y$, to produce samples $X$ which do not exist in the training data nor any provided test set. The problem is challenging when data $X$ is in high dimensional space, where a grid of $X$ can not be efficiently constructed.
In particular,  when $X$ is graph nodal feature data, the dimensionality of $X$ scales linearly with the graph size $N$ (the number of nodes in the graph).
In the case of categorical response, the graph label $Y$ assigns one of the $K$-class labels to each node, which makes the total possible outcomes $K^N$ many.
Thus the inverse prediction problem on graph data poses both modeling and computational challenges when scalability to large graphs is needed.

\begin{figure*}[t]
    \centering
    \includegraphics[width=\linewidth]{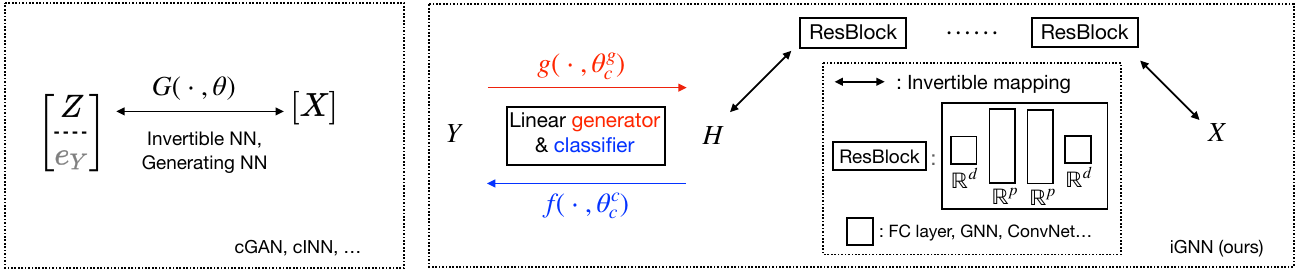}
    \vspace{-2pt}
    \caption{
    Comparison of existing conditional generative neural network models (left) and the proposed iGNN model (right). 
    Most current approaches concatenate the 
    encoded prediction label $Y$ (e.g. one-hot encoding)
    as an additional input to the generative network $G$. Our model takes a two-step approach:
    a one-to-many mapping $g$ from label $Y$ to intermediate feature $H$ by a Gaussian mixture model (which allows classification from $H$ to $Y$ by $f$),
    and a one-to-one mapping from $H$ to input data $X$.
    \vspace{-10pt}
    }
    \label{fig:comparison}
\end{figure*}

In this work, we develop a deep generative model for the conditional generation task of the inverse prediction problem on graphs. 
While deep generative models like
generative adversarial networks (GAN) \cite{GAN,WassersteinGAN}, 
 variational auto-encoders (VAE) \cite{VAE,VAE_review} and normalizing flow networks \cite{nflow_review} have been intensively developed in recent years,
the conditional generation task given categorial prediction labels is less studied. 
We further review the literature and comment on related works in Section \ref{subsec:literature}. 
Unlike previous conditional generative models \citep{cGAN_old,cGAN,cinn1,cinn2,cinn2+}, 
which typically concatenate the prediction label $Y$ with the random code $Z$ as input 
or rely on curated forms of neural network (NN) layers to ensure invertibility,
we propose to encode the input data $X$ one-to-one by an invertible network to an intermediate feature $H$, 
from which the label $Y$ can be predicted using a linear classifier, and in the other direction $H$ can also be generated from $Y$ by a parametric mixture model. 
The framework of our model is shown in Fig. \ref{fig:comparison}. 
Because the general (non-graph) data case can be seen as a special case of graph data where the graph only has one node, 
we call the proposed model invertible Graph Neural Network (iGNN), as a unified name.

To efficiently handle the up to exponentially many outcomes of graph nodal labels, we introduce a factorized mixture distribution $H | Y$ over graphs.
We theoretically show that the capacity to learn the conditional distribution $X|Y$ over a graph can be achieved when the per-node Gaussian mixture model $H_v | Y_v$ has well-separated components, and the need separation increases with graph size $N$ only mildly (an additive ${\log N}$ factor), cf. Proposition \ref{prop:separate}.
Under this formulation, the $Y$-to-$H$ sub-network is light in model and computation. 
The major part of the model capacity of the iGNN model is in the $H$-to-$X$ sub-network, 
which is an invertible flow model based on Residual Network (ResNet) framework \cite{iResnet}.
The computational scalability to large graphs is tackled by adopting GNN layers, 
which are not compatible with normalizing flow models having constrained formats in architecture \citep{RNVP,Wehenkel2019UnconstrainedMN},
and would also complicate the spectral normalization technique in \cite{iResnet} to ensure network invertibility. 

To overcome these issues, we propose Wasserstein-2 regularization of the ResNet, which in the limit of a large number of residual blocks
recovers the transport cost in the dynamic formulation of optimal transport (OT) \cite{villani2009optimal,benamou2000computational,villani2021topics}. 
Based on known results of OT theory, the Wasserstein-2 regularization leads to smooth trajectories of the transported densities.
The invertibility of the induced flow map can then be guaranteed with a sufficient number of residual blocks.
We empirically verify the invertibility of the flow model in experiments. 
This allows our model to use free-form neural network layers inside the ResNet blocks, including any GNN layer types. 
Theoretically, we study the existence and invertibility of flow mapping based on theories of OT and diffusion process.
We also prove the expressiveness of GNN layers to approximate the velocity field in the theoretical flows when data is a Gaussian field on the graph.

We examine the performance of the proposed iGNN model on simulated graph data as well as data from real-world graph prediction applications.
In summary, the contributions of the work are
%

$\bullet$ We propose a two-step procedure, $Y$-to-$H$ and $H$-to-$X$, to tackle the generative task of inverse prediction problem viewed as a conditional generation problem
and develop an invertible flow model consisting of two subnetworks accordingly. The model is made scalable to graph data by a factorized formulation of the parametric mixture model $H|Y$ and the GNN layers in the invertible flow network between $H$ and $X$. 

$\bullet$ We introduce Wasserstein-2 regularization of the invertible flow network, which is computationally efficient and compatible with general free-form layer types, particularly the GNN layers in the flow model. The effect on preserving invertibility is backed by OT theory and verified in practice. 

$\bullet$ The existence of invertible flow maps is analyzed theoretically. Based on the theoretical flows, we analyze the expressiveness of spectral and spatial graph convolution layers applied to graph data.

$\bullet$ The proposed iGNN model is applied to both simulated and real-data examples, showing improved generative performance over alternative conditional generation models. 

\noindent
{\it  Notation.} We denote by $[n] = \{1,\cdots, n \}$ the integer set.
$\mathbb{E}_{x\sim p}$ denotes the expectation over $x \sim p$, that is, $\mathbb{E}_{x\sim p} f(x) = \int f(x) p(x) dx$. 
When a probability distribution has density, we use the same notation, e.g., $p$, to denote the distribution and the density. 
For real-diagonalizable matrix $A$, and function $f: \R \to \R$, we denote by $f(A)$ the function of the matrix $A$. 
For $T: \R^d \to \R^d$ and $p$ a distribution on $\R^d$, 
we denote by $T_\# p $ the push-forward of distribution $p$ by $T$, i.e., $(T_\# p)(\cdot) = p( T^{-1} (\cdot))$.

\section{Background}\label{sec:background}

\subsection{Related works}\label{subsec:literature}

\noindent
{\it Generative deep models and normalizing flow.} At present, generative adversarial networks (GAN) \citep{GAN, WassersteinGAN} and variational auto-encoders (VAE) \citep{VAE,VAE_review} are two of the most popular frameworks that have achieved various successes \citep{Makhzani2015AdversarialA,Zhu2017UnpairedIT,Ledig2017PhotoRealisticSI}. However, they also suffer from clear limitations such as 
notable difficulties in training, such as mode collapse \citep{Salimans2016ImprovedTF} and posterior collapse \citep{Lucas2019UnderstandingPC}. On the other hand, normalizing flows (see \citep{nflow_review} for a comprehensive review) estimate arbitrarily complex densities via the maximum likelihood estimation (MLE), and they transport original random features $X$ into distribution that are easier to sample from (e.g., standard multivariate Gaussian) through invertible neural networks. Flow-based models can be classified into two broad classes: the discrete-time models (some of which include coupling layers \citep{RNVP}, autoregressive layers \citep{Wehenkel2019UnconstrainedMN} and residual networks \citep{iResnet,ResFlow}), and the continuous-time models as exemplified by neural ODE \citep{FFJORD,OT-Flow}. Most normalizing flow methods focus on unconditional generation with little development in a conditional generation. In addition, to achieve numerically reliable training, regularization of the density transport trajectories in flow networks are necessary but remain a challenge. In this work, we propose Wasserstein-2 regularization motivated by the transport cost in the dynamic formula of optimal transport (OT) theory. 
In Remark \ref{rk:spectral-normalization},  we compare with the original spectral normalization in iResNet \cite{iResnet} in more detail.

\noindent
{\it Conditional generation networks.}
The conditional generation versions of GAN (cGAN) have been studied in several places \citep{cGAN_old,cGAN},
where the  prediction outcome $Y$ (one-hot encoded as $e_Y$) are typically concatenated with random noise $Z$ and taken as input to the generator network,
as illustrated in the left of Fig. \ref{fig:comparison}. 
The one-hot encoding of categorical $Y$ concatenated with Gaussian $Z$ poses challenges in training cGAN models,
in addition to known issues of their unconditional counterparts, such as mode collapse, posterior collapse, and failures to provide exact data likelihood. 
When the prediction label $Y$ is lying on a graph having $N$ nodes, 
the one-hot coding will increase up to $O(N)$ more coordinates to the input $(Z,e_Y)$, 
which significantly increases the model complexity and computational load with large graphs.  
Conditional invertible neural network (cINN) model was developed in \citep{cinn1} for analyzing inverse problems.
The model inherits the approach of cGAN models to concatenate one-hot encoded prediction label $e_Y$ with normal code $Z$
while using Real-NVP layers \cite{RNVP} to ensure neural network invertibility. 
In terms of training objective, \cite{cinn1} proposed to use maximum mean discrepancy (MMD) losses to encourage 
both the matching of the input data distribution 
and the independence between label $Y$ and normal code $Z$. 
We call the method in \cite{cinn1} \ciNNI.
Replacing the MMD losses with a flow-based objective,
\cite{cinn2,cinn2+} extended the invertible network approach in \cite{cinn1} and applied to image generation problems.
In the models in \cite{cinn2} and \cite{cinn2+},
which we call \ciNNII \ and \ciNNIIplus\ respectively, 
the inputs to the Real-NVP layers contain encoded information of the prediction label $Y$ so as to learn label-conditioned generation,
see more in Fig. \ref{fig:comparison_appendix}.
In experiments, we compare with cGAN and cINN models on both non-graph and graph data.

\subsection{Transport cost regularization}\label{subsec:transport-cost}

We review needed background on the transport cost and the dynamic formula of Wasserstein-2 optimal transport (OT). 
The transport cost has been proposed to regularize the normalizing flow neural network models, e.g., in \cite{OT-Flow}. Consider the trajectory $x(t)$ satisfying $ \dot{x}(t) = v( x(t), t)$,
that is, $v(x,t)$ is the velocity field,
where $x(0) \sim p$ and $t \in [0,1]$.
The transport cost is defined as
\begin{equation}\label{eq:def-transport-cost}
   \mathcal{T} := 
    \int_0^1 \mathbb{E}_{x \sim \rho(\cdot,t) } \| {v}(x,t) \|^2 dt,
\end{equation}
where $\rho(x,t)$ is the marginal distribution of $x(t)$ after pushing forward the initial distribution $\rho(\cdot, 0) = p$
by $v(x,t)$ for time $t$. 
The cost \eqref{eq:def-transport-cost} can be viewed as taking the kinetic energy in computing the Lagrangian action along the trajectory. A more general form of cost under the framework of Mean-field Games has been proposed in the current and independent work \cite{huang2022bridging} to regularize trajectories of normalizing flows. 
It is also known that minimizing the cost $\mathcal{T}$ under the constraint of transporting from fixed $p = \rho(\cdot, 0)$ to $q = \rho(\cdot, 1)$ leads to the dynamic formulation of OT, i.e.,  the Benamou-Brenier formula \cite{villani2009optimal,benamou2000computational}.
Specifically, the solution of the minimization 
\begin{equation}\label{eq:Benamou-Brenier}
\begin{split}
&\inf_{\rho, v} \int_0^1 \mathbb{E}_{x \sim \rho( \cdot, t)} \| {v}(x,t) \|^2 dt  \\
& s.t.  \quad
    \partial_t \rho + \nabla \cdot (\rho v) = 0, 
\quad \rho( \cdot, 0) = p,    
\quad \rho( \cdot, 1) = q,    
\end{split}
\end{equation}
recovers the Wasserstein-2 OT from $p$ to $q$,
that is, at the minimizer $v(x,t)$, 
$\rho(\cdot, 1) = q$ and 
$\mathcal{T} = W_2^2(p,q)$.
The minimizer $v(x,t)$ of \eqref{eq:Benamou-Brenier} can be interpreted as the optimal control of the transport problem from $p$ to $q$.

In terms of neural network implementation, OT-Flow \cite{OT-Flow} used the potential model based on OT theory and parametrized the potential $\Phi( x,t)$ by a neural network, 
where 
$\nabla \Phi( x,t)$ gives the velocity $v(x,t)$. 
Our method adopts a ResNet base model, where the invertibility is fulfilled by a Wasserstein-2 regularization,
which can be viewed as a finite-step discrete-time counterpart of the transport cost $\mathcal{T}$.
 The proposed Wasserstein-2 regularization recovers $\mathcal{T}$ with a large number of steps, yet does not require integrating the continuous-time flow with accuracy for all time, cf. Remark \ref{rk:W2-reg-finite-L}.

\section{Methods}\label{sec:setup}

Given data-label pairs $\{X_i, Y_i\}$, we first describe our approach when $X_i$ is a sample in $\R^d$, and $Y_i$ is the categorical label in $K$ classes, i.e., $Y_i \in [K]$.
The case where both $X_i$ and $Y_i$ lie on a graph is addressed in Section \ref{subsec:scale-to-graph} where we make the approach scalable to large graphs. 
In Section \ref{subsec:X-H}, we introduce Wasserstein-2 regularization to preserve the invertibility of the trained flow network with free-form residual block layer types.
All proofs are in Appendix \ref{sec:proof}.

\subsection{Inverse of prediction as conditional generation}

The overall framework is to (end-to-end) train a network consisting of two sub-networks:  the first sub-network maps invertibly from $X$ to an intermediate representation $H$,
and the second sub-network maps from $H$ to label $Y$, which is a classifier and loses information. Specifically, 
%
    
$\bullet$    
    $H$-$Y$ classification sub-network.
    We model $H|Y$ by a Gaussian mixture model with ``well-separated'' means (detailed in Section \ref{subsec:H-Y}).
    The parametric form of $H|Y$ contains trainable parameter $\theta_c^{g}$,
    and the generation of $H|Y$ is by sampling the corresponding mixture component accordingly. 
    The prediction of label $Y$ from $H$ can be conducted by a linear classifier parametrized by $\theta_c^{c}$.
    The trainable parameters in the classification sub-network are denoted as $\theta_c=(\theta_c^{g},\theta_c^{c})$. 
    
$\bullet$   $X$-$H$ invertible sub-network.
    The invertible mapping from $X$ to $H$ is by a flow ResNet in $\R^d$ (detailed in Section \ref{subsec:X-H}).
    The sub-network parameters are denoted as $\theta = \{\theta_l, l=1,\cdots, L\}$, where $L$ is the number of residual blocks, and the network mapping is denoted as $F_\theta$.
    The prediction of label $Y$ from input $X$ is by first computing $H = F_\theta(X)$ and then applying the $H$-to-$Y$ classifier sub-network parametrized by $\theta_c^c$.
    The generation of $X|Y$ is by inversely mapping $X = F_\theta^{-1}(H)$ once $H$ is sampled according to $p(H|Y)$ parametrized by $\theta_c^g$.

\begin{remark}[Encoding-decoding perspective]
 The design of the network can be regarded as an encoding scheme that maps $X|Y$ to $H|Y$, and in the encoded domain $H|Y$ are ``noisy codes'' represented by separable isotropic Gaussian that can be linearly classified and generated (through the $H$-$Y$ sub-network). 
 In this sense, the invertible neural network $F_\theta(X)$ can be viewed as an invertible encoder that preserves information content of $ X | Y$ in the encoding domain $F_\theta(X)|Y$.
\end{remark}

The end-to-end training objective of the proposed network can be written as
\begin{equation}\label{end-to-end-obj}
     \underset{\{\theta, \theta_c\}}{\min}
     \ \mathcal{L}_g + \mu \mathcal L_c + \gamma \mathcal{W} ,
\end{equation}
where $\mathcal{L}_g$, $\mathcal{L}_c$  and $\mathcal{W}$
are the generative loss,
the classification loss 
and the Wasserstein-2 regularization, respectively.
The scalars $\mu, \gamma \ge 0$ are penalty factors.
We will explain the choice of $\mu$ later in this subsection, and the choice of $\gamma$ is explained after the derivation in Section \ref{subsec:X-H}.

Given $n_{\rm tr}$ many data-label pairs  $\{X_i, Y_i\}$, 
the generative loss is defined as 
$\mathcal{L}_g =  \frac{1}{n_{\rm tr}} \sum_{i=1}^{n_{\rm tr}} \ell_g ( X_i, Y_i)$ 
where
\begin{align}
   \hspace{-0.1in} - \ell_g ( X_i, Y_i)
    & =\log p_{H|Y_i}(F_{\theta}(X_i)) +\log |\det J_{F_{\theta}}(X_i)|.\label{Nflow_adaptive}
\end{align}
Because $p_{H|Y}$ is a mixture model parametrized by $\theta_c^g$, the per-sample loss $\ell_g ( X_i, Y_i)$ is determined by both $\theta$ and $\theta_c^g$. 
We specify the construction and training of $p_{H|Y}$ in Section \ref{subsec:H-Y}.
The mapping $F_\theta$ is by an invertible ResNet.
The construction of $F_\theta$, including the computation of $\log |\det J_{F_{\theta}}|$  and the regularization loss $\mathcal{W}$, will be explained in Section \ref{subsec:X-H}.

Finally, the classification loss is defined as
$\mathcal{L}_c =  \frac{1}{n_{\rm tr}} \sum_{i=1}^{n_{\rm tr}} \ell_c ( F_\theta(X_i), Y_i)$, 
where $\ell_c( H_i, Y_i)$ is the per-sample  $K$-class classification \rev{cross-entropy} loss computed by softmax. 
We use a linear classifier to predict $Y$ given $H$, parametrized by $\theta_c^c$. 
When $F_\theta(X) | Y$ is close in distribution to the Gaussian mixture model $H|Y$ specified by $\theta_c^g$, one may also use $\theta_c^g$ to construct the linear classifier from $H$ to $Y$, that is, to tie the parameters $\theta_c^c$ and $\theta_c^g$. Here we separately parametrize the forward and inverse prediction parameters in $H$-$Y$ sub-network so as to facilitate optimization since, in both directions, the model is light.  
For the penalty factor $\mu$, we find the experimental results insensitive to the choice. We use $\mu = 1$ in all experiments.

\begin{figure*}[t]
\centering
    \includegraphics[width=\linewidth]{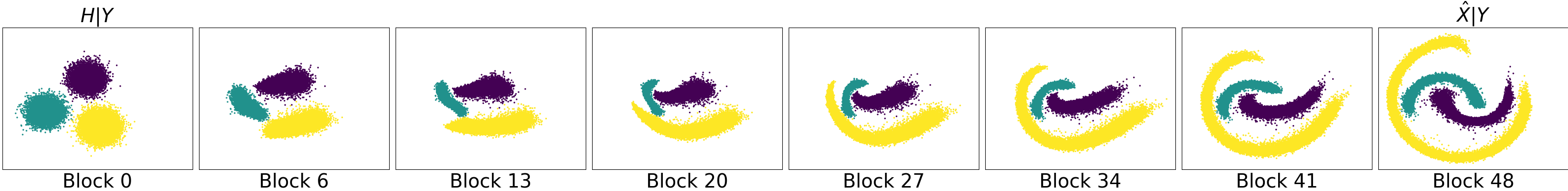}
    \includegraphics[width=\linewidth]{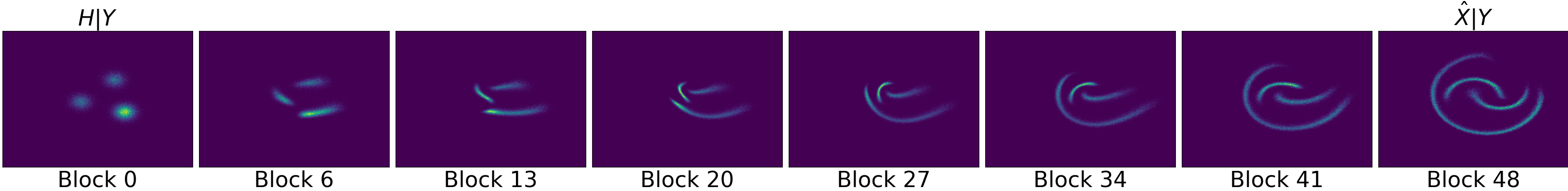}
\caption{ 
Flow map learned by iGNN model
that transports three-class data in $\R^2$ to a three-component Gaussian mixture and back. 
The distribution $X|Y$ has $(0,22, 0.22, 0.56)$ fractions in each classes respectively.
The ResNet has 48 blocks.
The transported data samples (upper panel) 
and distribution (lower panel) 
of the three-class data are illustrated along the trained invertible flow network. 
\vspace{-10pt}}
\label{three_moon}
\end{figure*}

\subsection{Mixture model of $H|Y$ and the shared flow}
\label{subsec:H-Y}

We parametrize $p(H|Y)$ by a Gaussian mixture model 
as 
\begin{equation}\label{eq:H|Y-one-node}
H | Y=k \sim \calN( \mu_k, \sigma^2 I_d), \quad k=1,\cdots, K,    
\end{equation}
where
$\sigma > 0$ is a prefixed parameter. 
The $K$ mean vectors $\mu_k$ will be trainable,
and we use isotropic Gaussian with the same covariance matrix for simplicity,
which may be generalized.
Note that we only parametrize the locations of means of the Gaussian components, and there is no need to specify the weights (fraction of samples in each component) because the fractions will be determined by the data.
Fig. \ref{three_moon} provides an example where the fractions in the three classes differ.

We will initialize and train the mean vectors $\mu_k$ to be sufficiently separated to ensure the Gaussian components have almost non-overlapping supports. 
Formally, we say that a collection of objects $S$ in $\R^d$ are $\delta$-separated if for any two distinct elements $s, s' \in S$, $d(s, s') \ge \delta$. 
When $S$ is a collection of points in $\R^d$,
$x, x' \in S$, then $d(x,x') = \| x-x'\|$.
When $S$ is a collection of sets in $\R^d$,
and $A, A' \in S$, we define
$d(A,A') = \inf_{x \in A, \, x' \in A'} \| x-x'\|$.
A set $S\subset \R^d$ is called  an $\epsilon$-support of a probability distribution $p$ if $p(S^c) \le \epsilon$.
When the means of Gaussian components are sufficiently separated, 
one would expect that there are $\epsilon$-supports of the $K$ components that are mutually separated. 
Technically, we have that $3\rho$-separation of the means 
guarantees $\rho$-mutual separation of $K$ $\epsilon$-supports, where $\rho \sim \sigma \sqrt{\log (1/\epsilon)}$.
The formal statement is proved in the following lemma. 
\begin{lemma}[Separation of $K$ components]\label{lemma:separate-Rd}
Let $\rho = \sigma \sqrt{2 \log (K/\epsilon)}$ for $ 0 < \epsilon < 1/2$. 
If the $K$ mean vectors $\{ \mu_k\}_{k=1}^K$ of the Gaussian mixture distribution  of  $H|Y$ are $  3 \rho$-separated in $\R^{d}$,
then each component $q_k = \calN( \mu_k, \sigma^2 I_d)$ in  \eqref{eq:H|Y-one-node} 
has an $\epsilon$-support $\Omega_k$ such that for $k \neq k'$,
$d( \Omega_k, \Omega_{k'} ) \ge \rho$. 
\end{lemma}

The separation condition of the mixture model,
especially the high dimensional counterpart in Proposition \ref{prop:separate}, may potentially be connected to the coding theory of Gaussian channel \cite{thomas2006elements}. We further comment on this in the discussion section. 
The purpose of having separated $\epsilon$-supports of the $K$ components in $H|Y$ is for the construction of an invertible flow map from $H$ to $X$ that transports each class of samples correspondingly.
Intuitively, the separated $\epsilon$-supports $\Omega_k^{(H)}$ of the $K$ components of $p(H|Y)$ 
allow to partition $\R^d$ into $K$ domains with smooth boundary where each domain contains $\Omega_k^{(H)}$.
If the $K$-class distributions $p(X|Y)$ also have separated $\epsilon$-supports $\Omega_k^{(X)}$,
then one can try to construct an invertible flow mapping in $\R^d$ which transports $p_{X|Y=k}$ to $p_{H|Y=k}$ by transporting from $\Omega_k^{(X)}$ to $\Omega_k^{(H)}$ respectively. 
While the transport from each $p_{X|Y=k}$ to $p_{H|Y=k}$ can be constructed individually, 
the simultaneous transports of the $K$ components by a  {\it shared} flow would need the transported components to stay separated from each other along the flow,
and this may not always be possible, e.g., due to topological constraints. 
For the shared invertible flow to exist, we introduce the following assumption:
\begin{assumption}[Shared flow]
\label{assump:same-flow}
There exist $\delta > 0$ and constant $C$ such that
for any $\epsilon < 1/2$, 
if the $K$ components $q_k = p_{H|Y=k} (\cdot)$ of the Gaussian mixture of $p(H|Y)$ have $\epsilon$-supports which are mutually $\delta$-separated in $\R^d$, 
then there exists a smooth invertible flow mapping $F :\R^d \to \R^d$  induced by velocity field $v(x,t)$ for $t \in [0,T]$
(i.e. $F(x) = x(T)$ from $x(0)=x$,  and $x(t)$ solves the ODE of $\dot{x}(t) = v(x(t),t)$)
such that the transported conditional distributions 
$F_\# p_{X|Y=k}(\cdot)  = \tilde{q}_k$ satisfy that 
 $W_2( \tilde{q}_k,  q_k) \le C \epsilon$  for $k=1,\cdots, K$. 
\end{assumption}
While theoretical justification of Assumption  \ref{assump:same-flow} goes beyond the scope of the paper and  is postponed here, 
we demonstrate the validity of the assumption empirically.
An example of the shared invertible flow on a 3-class data in $\R^2$ is illustrated in Fig. \ref{three_moon}.
In all our experiments, we find that the trained invertible ResNet successfully transports from $p(X|Y)$ to $p(H|Y)$ for all $K$ classes.
In the special case of having $K=1$ class, the conditional generation problem reduces to an unconditional one
and the shared flow is no longer an issue. 
In this case, one can set the distribution of $H$ to be a standard normal, and the desired transport map is a normalizing flow. 
In Section \ref{sec:theory-flow} we further study the existence and theoretical properties of the flow mapping when $K=1$, for general data and graph data.

Assumption \ref{assump:same-flow} guarantees the existence of a shared flow mapping which transports the $K$-class conditional densities up to $O(\epsilon)$ error
as long as the components of the Gaussian mixture distribution $p(H|Y)$ have $\delta$-separated $\epsilon$-supports. 
Meanwhile, by Lemma \ref{lemma:separate-Rd}, $p(H|Y)$ will have $\rho$-separated $\epsilon$-supports if the means $\mu_k$ are separated as therein.
Since $\rho \sim \sigma \sqrt{\log (1/\epsilon)}$, it will exceed the $O(1)$ constant $\delta$ required by Assumption \ref{assump:same-flow} when $\epsilon$ is small.
This suggests keeping the mean vectors $\mu_k$ to be separated at the order of $\sigma\sqrt{\log (1/\epsilon)}$.
In practice, we initialize  $\mu_k$ to be separated at such a scale and preserve the separation during training via a barrier penalty on the distances $\|\mu_k - \mu_{k'}\|$.

Note that Assumption \ref{assump:same-flow} ensures the existence of a continuous-time flow, 
and we will use a ResNet model to represent the flow by finite-step composed transports which are regularized by summed discrete-time transport cost,
to be detailed in Section \ref{subsec:X-H}.

\vspace{-5pt}
\subsection{Scalable conditional generation on graph data}
\label{subsec:scale-to-graph}

The mixture modeling of $H|Y$ described in the previous section is scalable to large graphs, as we explain in this subsection.
Consider the graph data where both the data vector $X$ and the label $Y$ are defined on each node in a graph.
In this subsection, the subscript $_v$ indicates graph node $v$. 
Suppose the graph $(V,E)$ has $N$ nodes in $V$, and the edge set $E$ is specified by an adjacency matrix $A$. 
We denote a graph data sample $X$ and a graph label $Y$ as 
$X = [ X_1, \cdots, X_N]^T \in \R^{N\times d'}$, 
$X_v \in \R^{d'}$,
$Y  = [ Y_1, \cdots, Y_N]^T \in \R^N$, 
$Y_v \in [K]$,  $v \in [N]$,
where $d'$ is the dimension of node feature. 
One may view $X$ as a vector in $\R^{d}$, $d= d'N$, 
and label vector $Y$ taking $K^N$ many possibilities,
and then apply the approach for non-graph data to the $K^N$-class prediction problem in $\R^d$. 
However, this may lead to exponentially many classes when the graph size $N$ is large, leading to difficulty in modeling $H|Y$ by a mixture distribution as well as large model complexity.

To make our approach scalable to large graphs, we adopt two techniques:
(i) we introduce a factorized form of $H | Y$ which is independent and homogeneous over graph;
(ii) we propose to use GNN layers in the invertible flow ResNet, which makes the neural network computation scalable. 
The scalability of our iGNN approach is demonstrated on a larger graph with $N=500$ nodes, cf. Fig. \ref{large_graph_cond_gen}.

(i)  {\it Factorized $H|Y$ over graph.}
Recall that $H$ is of same dimension as $X$ and also defined on $V$, 
\[
 H = [ H_1, \cdots, H_N]^T \in \R^{N\times d'}, 
    \ H_v \in \R^{d'}.
\]
Suppose we have a $K$-component Gaussian mixture distribution of means $\mu_k$ in $\R^{d'}$, 
we specify the graph $H|Y$ as 
\begin{equation}\label{eq:H|Y-factored}
p ( H | Y) = \prod_{v=1}^N p(H_v | Y_v),
\quad H_v | Y_v \sim \calN( \mu_{Y_v}, \sigma^2 I_{d'}), 
\end{equation}
that is, the joint distribution of $H|Y$ consists of 
independent and identical $K$-component Gaussian mixture distribution of $H_v|Y_v$ in $\R^{d'}$ across the graph. 
As a result, on the graph sample-label pair $\{ X,Y \}$, 
the $\log p(H|Y)$ term in the  generative loss  \eqref{Nflow_adaptive} can be computed as 
\begin{equation}\label{Nflow_graph}
    \log p_{H|Y}( F_\theta(X)) 
    = \sum_{v=1}^N \log p_{H_1| Y_1} ( (F_\theta(X))_v),
\end{equation}
where $p_{H_1| Y_1} $ is specified by a Gaussian mixture model in $\R^{d'}$. 

The factorized form of $H|Y$ reduces the complexity of modeling $H|Y$ in $\R^{d'N}$ to that of modeling a $K$-class mixture model in $\R^{d'}$. 
At the same time, one would prefer that such a reduction does not prevent the mixture distribution $H|Y$ from presenting any sufficiently separated conditional distribution $X|Y$ 
via an invertible mapping in $\R^{d'N}$. 
Under Assumption \ref{assump:same-flow}, the desired shared flow in $\R^{d'N}$ exists
as long as the $O(1)$-separation between some $\epsilon$-supports of the $K^N$ components of $p(H|Y)$ can be achieved. 
Due to that  $p(H|Y)$ 
is $\calN( \mu_Y, \sigma^2 I_{d' N})$ 
($\mu_Y$ is concatenated from $\mu_{Y_v}$, $v\in [N]$),
and the length of 
$H-\mu_Y$ is of order $\sigma \sqrt{d'N}$,
this raises the question of how the separation of the Gaussian means of $H_1|Y_1$ in $\R^{d'}$ needs to scale with large $N$. 
The following proposition shows that, theoretically,
compared to Lemma \ref{lemma:separate-Rd} 
only an additional $O(\sigma \sqrt{\log N  })$ separation is needed. 

\begin{proposition}[Separation of $K^N$ components]\label{prop:separate}
Let $\rho_N = \sigma \sqrt{2 \log (2NK/\varepsilon)}$ for $ 0 < \varepsilon < 1/2$. 
If the $K$ mean vectors $\{ \mu_k\}_{k=1}^K$ of the Gaussian mixture distribution  of  $H_1|Y_1$ are $  3 \rho_N$-separated in $\R^{d'}$,
then the $K^N$ components of the Gaussian mixture distribution of $H|Y$ in $\R^{d'N}$ satisfy that each component $q_Y$ has an $\varepsilon$-support $\Omega_Y$
and for any $Y \neq Y'$, $d(  \Omega_Y, \Omega_{Y'}) \ge \rho_N$. 
\end{proposition}

The proposition suggests that the needed separation of Gaussian mixture means in $\R^{d'}$ increases mildly when the graph size $N$ increases. 
In experiments, we find that using the same algorithmic setting to preserve mean vectors' separation in $\R^{d'}$ suffices for graph data experiments where $N$ is up to a few hundreds.

(ii) {\it Scalable computation on the graph.}
As will be introduced in Section \ref{subsec:X-H}, 
the residual block in the proposed iGNN model can take a general form
(the invertibility will be enforced by a Wasserstein transport regularization). 
This allows using any GNN layer type in the residual block, which can be made computationally scalable to large graphs \citep{wu2020comprehensive}.

Consider input graph data $X$ in $\R^{ N \times C}$, where $N$ is the number of nodes, and $C$ is the dimension of node features. 
We denote by $Y$ the output graph data in $\R^{ N \times C'}$ after the graph convolution, 
and we adopt spectral and spatial graph convolution layers in this work. 
Because $X$ and $Y$ can be hidden-layer features inside the residual block, the dimension $C$ and $C'$ may be larger than the data node feature dimension $d'$.
Let $ \tilde{L} \in \R^{N \times N}$ be the graph Laplacian matrix, possibly normalized so that the spectrum lines on a unit interval. 
Omitting the bias term and the non-linear node-wise activation function, the GNN layers used in this work are

\noindent
$\bullet$ ChebNet \cite{Chebnet}, where $Y = \sum_{k=1}^K T_k(\tilde{L}) X  \Theta_k$, parametrized by $\Theta_k \in \R^{C \times C'}$, and  $T_k$ is the Chebyshev polynomial of degree $k$.

\noindent
$\bullet$ L3Net \cite{cheng2021graph}, where  $Y = \sum_{k=1}^K B_kXA_k$, parametrized by $B_k$ which are local filters on graph and $A_k \in \R^{C \times C'}$. 

The benefit of using GNN layer lies in the reduced model and computational complexities:  
the number of trainable parameters in a ChebNet layer is  $O(KCC')$, 
and that in an L3Net layer is  $O(K(CC'+Nv))$
  where $v$ denotes average local patch size \citep{cheng2021graph}.
  In contrast, a fully-connected layer mapping from $X$ to $Y$ would have $O(N^2CC')$ many parameters, which is not scalable to large graphs. 
  We include the fully-connected layer baseline in experiments when the graph size is small and compare performance with the GNN layers.

\vspace{-5pt}
\subsection{Invertible flow network with Wasserstein-2 regularization}\label{subsec:X-H}

We use an invertible flow network to construct the encoding
which maps one-to-one between $X$ and $H$. 
The base model follows the framework of \citep{iResnet}. 
The ResNet has $L$ residual blocks, and the $l$-th block residual mapping $f(x, \theta_l)$ is parametrized by $\theta_l$. 
The overall ResNet mapping from $X$ to $H$ can be expressed as $H=F_\theta(X) = x_L$ where 
\begin{align}\label{eq:resnet}
 x_{l} & = x_{l-1} + f(x_{l-1}, \theta_l), \quad l=1, \cdots, L, \quad x_0 = X.
\end{align}
In each of the $L$ residual blocks, we use a shallow network with one or two hidden layers for $f(x,\theta_l)$. The parameter $\theta_l$ consists of the weights and bias vectors of the shallow network. 
The ResNet architecture is illustrated on the right of Fig. \ref{fig:comparison}.
The architecture of the residual block is free-form: 
one can use any layer type inside the residual blocks and even different layer types in different blocks. 
In this work, we use a fully-connected layer for Euclidean data and GNN layers for graph data. 
To compute the log determinant in \eqref{Nflow_adaptive}, we express the quantity as a power series and adopt the technique in \citep{ResFlow} to obtain an unbiased estimator. Further details can be found in Appendix \ref{sec:setup_details}.

The invertibility of the ResNet is to be fulfilled by using sufficiently large $L$ together with the Wasserstein-2 regularization in \eqref{end-to-end-obj},
which we introduce here.
The regularization $\mathcal{W}$ takes the form as
\[
\mathcal{W} = \frac{1}{n_{\rm tr}} \sum_{i=1}^{n_{\rm tr}} \ell_w (X_i),
\quad \ell_w (X_i)= \sum_{l=1}^L \|  x_{l} - x_{l-1} \|_2^2,
\]
where $x_0 = X_i$ and $x_l$ is defined as in \eqref{eq:resnet}. 
Replacing the empirical measure of $n_{\rm tr}$ training samples by the data density $p$ gives the population counterpart of $\mathcal{W}$ as 
$W := \mathbb{E}_{x_0 \sim p}  \sum_{l=1}^L \| x_{l} -  x_{l-1}  \|_2^2$.
We denote by $T_l$ the transport map from $x_{l-1}$ to $x_l$, that is,
\begin{equation}\label{eq:def-Tl-resnet}
T_{l}(x ) := x + f(x, \theta_l), \quad T_l: \R^d \to \R^d, 
\end{equation}
and then $x_l = T_l (x_{l-1})$. Define the composite of $T_l$'s as $F_l$, which is the transport mapping from $x_0$ to $x_l$, namely,
$F_l := T_l\circ \cdots \circ T_1$,
$x_l = F_l( x_0)$, for $l=1,\cdots, L$,
and $F_0 = \rm{Id}$.
We also define $\rho_l = (F_l)_{\#} p$ which is the marginal distribution of $x_l$, and $\rho_0 = p$. 
Then 
$W$
can be equivalently written as
 \begin{align}\label{eq:W-reg-2}
 W 
 =  \sum_{l=1}^L \mathbb{E}_{x \sim \rho_{l-1}}  \|  T_{l}( x )  - x \|_2^2.
 \end{align}
We will show in Section \ref{subsec:W2-interprete} that regularizing by $W$  is equivalent to adding
\[
W':= \sum_{l=1}^L  W_2( \rho_{l-1}, \rho_l )^2
\]
to the minimizing objective. We thus call the proposed regularization the ``Wasserstein-2 regularization,''
because
$W'$
sums over the step-wise (squared) Wasserstein-2 distance over the $L$ steps.

In the limit of large $L$, the proposed regularization by $W$ serves to penalize the transport cost. 
When $K=1$, the optimal flow induced by the minimizer gives the Wasserstein geodesic from $p$ to the normal density, cf. Section \ref{sec:invertibility}.
In our problem of conditional generation, when there are $K>1$ classes, 
the limiting continuous time flow differs from the Wasserstein geodesic from $\rho(\cdot, 0) = p(X|Y)$ to $\rho(\cdot, 1) = p(H|Y)$ for fixed $Y$,
but is expected to give a shared flow in $\R^d$ from $X$ to $H$, cf. Assumption \ref{assump:same-flow} and 
Fig. \ref{three_moon}.
Under regularity conditions, one would expect the transport-cost regularized continuous-time flow to also be regular
(the $K=1$ case is proved in Proposition \ref{prop:BB-flow}).
As a result, the velocity field $v(x,t)$ would have a finite $x$-Lipschitz constant $B$ on any bounded domain. 
Then, the transport from $x(t_{l-1})$ to $x(t_l)$ induced by $v(x,t)$ on the interval $[t_{l-1}, t_l]$ is invertible when $B \Delta t  < 1$, which holds when $L > B$. 
In this case, 
also assuming that the solved discrete-time transport map $T_l$ is close to that induced by $v(x,t)$, 
one can expect the invertibility of $T_l$ to hold for each $l$, 
and then the composed transport $F_L = F_\theta$ is also invertible.

We numerically verify the invertibility of the trained $T_l$ in ResNet in experiments, cf. Table \ref{W2_inv_err} and Appendix \ref{sec:exp_append}. 
Our analysis in Section \ref{subsec:W2-interprete}
also suggests that the regularization factor $\gamma$ should scale with $L = 1/\Delta t$, the number of residual blocks.
In experiments, we find that the invertibility of the ResNet can be guaranteed over a range of choices of  $\gamma$, and the quality of generation is insensitive to the choice, see more in
Appendix \ref{append_hyperparam}.
The number of blocks $L$ to use depends on the complication of the data distribution, and in all our experiments, we find a few tens to be enough, including the large graph experiment.

\begin{remark}[Comparison to spectral normalization]
\label{rk:spectral-normalization}
iResNet \citep{iResnet} proposed spectral normalization to ensure the invertibility of each residual block. 
Given a weight matrix $W \in \R^{C\times C'}$ in a fully-connected layer, 
the method first computes an estimator $\tilde{\sigma}$ of the spectral norm $\|W\|_2$ by power iteration \citep{gouk2021regularisation},
and then modify the weight matrix $W$ to be 
$ c W / \tilde{\sigma}$ if $c / \tilde{\sigma}<1$,
where $c < 1$ is a pre-set scaling parameter. 
It was proposed to apply the procedure to all weight matrices in all ResNet blocks in every stochastic gradient descent (SGD) step with mini-batches. 
When the number of blocks $L$ is large, this involves expensive computation, especially if the hidden layers are wide, i.e., $C$ and $C'$ being large. 
In addition, while spectral normalization of fully-connected layers  together with contractive nonlinearities (e.g., ReLU, ELU, Tanh) ensures invertibility, 
it may not be directly applicable to other layer types, e.g., GNN layers. 
In contrast, the proposed Wasserstein-2 regularization 
are obtained from the forward passes of the residual blocks on mini-batches of training samples without additional computation. It is also generally compatible with free-form neural network layer types. 
\end{remark}

\section{Theory: Invertible flow}\label{sec:theory-flow}

In this section, we first interpret the proposed  Wasserstein-2 regularization in view of transport cost in Section \ref{subsec:W2-interprete}.
In Section \ref{sec:invertibility},  utilizing theories of diffusion process and the dynamic formulation of OT,
we theoretically study the existence of invertible flow in the simplified case where $K=1$.
All proofs are in Appendix \ref{sec:proof}. 

Throughout the section, we consider flow mapping induced by a velocity field $v(x,t)$, that is, the continuous-time flow is represented by an initial value problem (IVP) of ODE
\begin{equation}\label{eq:ode-Xt}
\dot{x} (t) = v( x(t), t), \quad  x(0) \sim p.
\end{equation}
The transport in $\R^d$ is the solution mapping from $x(0)=x$ to $x(t)$ at some time $t > 0$.

\subsection{Interpretation of Wasserstein-2 regularization}\label{subsec:W2-interprete}

The proposed regularization by $W$ in Section \ref{subsec:X-H}
can be interpreted using the dynamic formula of the Wasserstein-2 transport,
where we recall the notations in Section \ref{subsec:transport-cost}.
Suppose the time interval $[0,1]$ is divided into $L$ time steps, $\Delta t = t_{l+1} - t_l  = {1}/{L}$.
With the optimal velocity field $v(x,t)$ and the corresponding $\rho(x,t)$ that minimize the transport cost in \eqref{eq:Benamou-Brenier},
on every time subinterval $[t_{l-1}, t_l]$, 
$\mathbb{E}_{x(t_{l-1})\sim \rho(\cdot, t_{l-1})}
\int_{t_{l-1}}^{t_l}  \| v( x(t), t) \|^2  dt
= { W_2( \rho( \cdot, t_{l-1} ), \rho( \cdot, t_{l} ) )^2 }/{\Delta t}$.
By the definition of the transport cost \eqref{eq:def-transport-cost}, we have
\begin{align}
\mathcal{T}  
  = & \sum_{l=1}^L 
  { W_2( \rho( \cdot, t_{l-1} ), \rho( \cdot, t_{l} ) )^2 }/{\Delta t}. 
  \label{eq:T-by-rhol}
\end{align}
Note that the right hand side (r.h.s.) only depends on $\rho(\cdot, t)$ at the time stamps $t_l$.
We define $\rho_l: = \rho(\cdot, t_l)$, and denote by $T_l$ the transport map from $x(t_{l-1})$ to $x(t_l)$ induced by the ODE of $x(t)$ on the time interval $[t_{l-1}, t_l]$, i.e.,
$T_l(x) = x + \int_{t_{l-1}}^{t_l} v(x(s),s) ds$ starting from $x(t_{l-1}) = x$.
About the notation, $\rho_l$ and $T_l$ here are determined by the continuous-time flow,
and the notations coincide with those in \eqref{eq:def-Tl-resnet} and \eqref{eq:W-reg-2} which are determined by the finite-step ResNet.
We use the same notations here since $T_l$ will be the variables to minimize the training objective, and thus ready to be parametrized by a residual block $f(\cdot, \theta_l)$.

Specifically, when the other part in the loss, denoted as $\mathcal{L}$, only depends on the terminal time density $\rho(\cdot, 1)=\rho(\cdot, t_L)$,
one can use the r.h.s. of \eqref{eq:T-by-rhol} as the regularization term added to $\mathcal{L}$, 
making  the overall objective as 
\begin{equation}\label{eq:L+T}
 \mathcal{L} +  \mathcal{T}
=  \mathcal{L} [ \rho_L ] +  \frac{1}{\Delta t} \sum_{l=1}^L  W_2( \rho_{l-1}, \rho_l )^2,  
\end{equation}
where  $\rho_l = (T_l)_{\#} \rho_{l-1}$, $\rho_0 = p$.
Note that for a given $p$, the densities  $\rho_1, \cdots, \rho_L$ are determined by the $L$ transport maps $T_l$'s,
this means that $T_1, \cdots, T_L$ can be used as the variables to minimize \eqref{eq:L+T}.

The following proposition shows that this minimization is equivalent to 
using the proposed Wasserstein-2 regularization in Section \ref{subsec:X-H}
(specifically, using $1/\Delta t$ times the r.h.s. of \eqref{eq:W-reg-2})
in solving for $T_l$'s, 
and there is no need to solve for the OT distance $W_2( \rho_{l-1}, \rho_l )^2$ with additional computation.

\begin{proposition}[Equivalent form of $W$]
\label{prop:W2-reg}
Minimizing \eqref{eq:L+T}  over $T_1, \cdots, T_L$ is equivalent to 
\begin{equation}\label{eq:L+T-reduce}
\min_{T_1, \cdots, T_l}  \mathcal{L} [ \rho_L ] +  \frac{1}{\Delta t} \sum_{l=1}^L  \mathbb{E}_{x \sim \rho_{l-1} } \|  T_l(x) -x  \|^2.
\end{equation}
\end{proposition}

\begin{remark}\label{rk:W2-reg-finite-L}
The regularization \eqref{eq:W-reg-2} can also be viewed as a first-order approximation of the transport cost \eqref{eq:def-transport-cost} using the Forward Euler scheme, that is, by (replacing the integral with finite summation at $t_l$ and) the approximation that $\|v(x(t_l), t_l) \| \approx { \|x(t_{l+1}) - x(t_l) \|}/{\Delta t}$.
Note that our derivation is based on \eqref{eq:L+T}, which coincides with the continuous-time transport cost in the limit of large $L$, but is well-defined for finite $L$. The objective \eqref{eq:L+T} only involves the discrete-time transported densities $\rho_l$'s, which are determined by the $L$ transport maps $T_l$'s,
and does not require modeling the continuous-time trajectory $x(t)$ for $t \in [t_{l-1}, t_l]$ with numerical accuracy. In principle, this allows to use a smaller number of residual blocks and model size 
to achieve an approximation of $T_l$'s only (and guarantees transport invertibility) than to approximate and solve for $v(x,t)$ and $x(t)$ for all $t$.
\end{remark}

\subsection{Existence and invertibility of flow} \label{sec:invertibility}

The transport mapping by the IVP \eqref{eq:ode-Xt} is invertible as long as \eqref{eq:ode-Xt} is well-posed.
For the $K$-class case, Assumption \ref{assump:same-flow} assumes the existence of the smooth velocity field $v(x,t)$ as well as the shared flow mapping. 
Here we consider the special case of $K=1$ (the unconditional generation problem), 
that is, the source density $p$ is the data distribution in $\R^d$ 
and the target density $q$ is standard normal $\calN(0, I_d)$.  
Though the $K$-class conditional generation problem is of the main interest of the paper and has called for additional requirement of the share flow,
our analysis of the $K=1$ case provides theoretical insights,
especially for the expressiveness of GNN layers in the flow model for graph data in Section \ref{sec:expressiveness}. 

The normalizing flow induced by $v(x,t)$ from $p$ to normal $q$ is typically not unique. 
We introduce two constructions here that are related to the proposed Wasserstein-2 regularization of the flow model.

\vspace{5pt}
{\it (i) Flow by Benamou-Brenier formula.}
Consider \eqref{eq:ode-Xt} on $t\in [0,1]$, the flow $F$ maps from $x=x(0)$ to $x(1)$.
Let $\rho(\cdot, t)$ be the density of $x(t)$, $\rho(\cdot, 0)= p$ and $\rho(\cdot, 1) = F_\# p$. 
We consider the flow induced by the optimal $v(x,t)$ in the Benamou-Brenier formula \eqref{eq:Benamou-Brenier},
where the regularity of $v$ and $\rho$ follows from classical OT theory:
\begin{proposition}[\cite{caffarelli1996boundary, villani2021topics}]
\label{prop:BB-flow}
Suppose $p$ is smooth on $\R^d$ with finite moments,
then the optimal velocity field $v(x,t)$ that minimizes \eqref{eq:Benamou-Brenier} is smooth,
the induced IVP \eqref{eq:ode-Xt} is well-posted on $\R^d \times [0,1]$ and the flow mapping is smooth and invertible. 
\end{proposition}

We now show that our optimization objective \eqref{end-to-end-obj} is equivalent to the action minimization in the Benamou-Brenier formula 
when $L$ is large so that the discrete-time transports approximate the continuous-time limit. 
When $K=1$, the generalization loss  (in population form) in \eqref{end-to-end-obj} reduces to 
\[
\mathcal{L}_g 
= - \mathbb{E}_{ x \sim p} \log (  (F^{-1})_\# q ( x ) ).
\]
By the relation \cite{papamakarios2021normalizing}
\begin{equation}\label{eq:reverse-KL}
{\rm KL}(  F_\# p || q) = {\rm KL}(   p || (F^{-1})_\# q),
\end{equation}
and that ${\rm KL}(   p || (F^{-1})_\# q) = \mathcal{L}_g + \int p \log p$, 
we have that 
$\mathcal{L}_g + \int p \log p = {\rm KL}(  F_\# p || q)$,
where $ \int p \log p$ is a constant independent of the model. 
The analysis in Section \ref{subsec:X-H} gives that $\mathcal{W}$ (in population form) is equivalent to $\frac{1}{L} \mathcal{T}$ with large $L$. 
Thus our minimizing objective $\mathcal{L}_g  + \gamma \mathcal{W}$ is equivalent to 
\begin{equation}\label{eq:BB-loss}
\mathcal{T} +  \tilde{\gamma} {\rm KL}(  \rho( \cdot, 1) || q )
\end{equation}
with some positive scalar $\tilde{\gamma}$.
Compared to the Benamou-Brenier formula \eqref{eq:Benamou-Brenier},
the objective \eqref{eq:BB-loss} relaxes the terminal condition that $\rho(\cdot, 1) = q$ to be the KL divergence, 
which does not change the solution of the optimal $v$.

\vspace{5pt}
{\it (ii) Flow by Fokker-Planck equation.}
Because the transport cost $\mathcal{T}$ equals squared Wasserstein-2 distance between $\rho(\cdot,0)=p$ and $\rho(\cdot,1)$ at optimal $v$,
the objective \eqref{eq:BB-loss}  is closely related to the problem of 
$\min_{\tilde{q}} {\rm KL}(  \tilde{q} || q ) + \frac{1}{\tilde{\gamma}}  W_2( p, \tilde{q})^2$.
When $\tilde{\gamma}$ is small, 
the problem 
is the first step of the Jordan-Kinderleherer-Otto (JKO) scheme \cite{jordan1998variational} 
to solve the Fokker-Planck equation of a stochastic diffusion process toward the equilibrium $q$. 
Because $q$ is standard normal, the stochastic process is an Ornstein-Uhlenbeck (OU) process. 
Without going to further connections to the JKO scheme, 
here we use the Fokker-Planck equation to provide another theoretical construction of the invertible normalizing flow. 
For the OU process in $\R^d$, the Fokker-Planck equation can be written as
\begin{equation}\label{eq:fokker-planck}
\partial_{t}\rho=\nabla\cdot(\rho\nabla V + \nabla \rho), 
\,  V(x)={|x|^{2}}/{2},
\, \rho(x,0) = p(x),
\end{equation}
where $\rho(x,t)$ represents the probability density of the OU process at time $t$. 
The Liouville equation of \eqref{eq:ode-Xt} is $\partial_{t}\rho=-\nabla\cdot(\rho v)$, where $\rho(x,t)$ represents the density of $x(t)$.
Comparing to \eqref{eq:fokker-planck}, 
we see that the density evolution can be made the same if  the velocity field $v(x,t)$ is set to satisfy
\begin{equation}\label{eq:f-jko-flow}
- v (x,t) = \nabla V(x) + \nabla \log \rho( x,t) = x + \nabla \log \rho( x,t).
\end{equation}
The smoothness of $v$ follows from that of $\rho(x,t)$,
which has an explicit expression as the solution of \eqref{eq:fokker-planck}.

\begin{proposition}\label{prop:velocity-OU-smooth}
Let $\rho( x,t)$ be the solution to \eqref{eq:fokker-planck} from $\rho(x,0) = p$, 
then the IVP \eqref{eq:ode-Xt} induced by velocity field $v(x,t)$ as in \eqref{eq:f-jko-flow} is well-posted on $\R^d \times (0,T)$ for any $T>0$
and the flow mapping is smooth and invertible. 
\end{proposition}

While theoretically the density $\rho(x,t)$ in \eqref{eq:fokker-planck} converges to the normal equilibrium $q$ in infinite time,
the convergence is exponentially fast \citep{bolley2012convergence}.
Thus the transported density $F_\# p = \rho(\cdot, T)$ for a finite $T \sim \log (1/\epsilon)$ can be $\epsilon$-close to $q$,
which then guarantees the closeness of $(F^{-1})_\# q$ to $p$ by \eqref{eq:reverse-KL}.

\section{Theory: Expressing flow of graph data}\label{sec:expressiveness}

We analyze the expressiveness of graph convolution layers to approximate the theoretical flow maps identified in Section \ref{sec:theory-flow}. 
All proofs are in Appendix \ref{sec:proof}. 

\vspace{-5pt}
\subsection{Theoretical flows of graph Gaussian field data}

Taking the continuous-time formulation, the goal is to represent the velocity field $v(x,t)$ by a GNN layer where $x$ is graph data. Assuming that the discrete-time flow can approximate the continuous-time counterpart when the number of residual blocks is large, 
the successful approximation of $v (x,t)$ by GNN layers indicates that the flow can be constructed by the invertible GNN flow network. 

For simplicity, we consider $x(0) \sim \calN( 0, \Sigma)$, which is a Gaussian field on the graph having $N$ nodes (The node feature dimension $d'=1$, and data dimension $d=N$).
The $N$-by-$N$ covariance matrix $\Sigma$ is positive semi-definite, and we further assume it is invertible. 
Suppose PSD matrix $A = VDV^T$ is the eigen-decomposition, where $V$ is an orthogonal matrix, 
define $A^{1/2} = V D^{1/2} V^T$. 
In this special case of Gaussian data, we have explicit expressions of the velocity field of the flow induced by 
Benamou-Brenier formula and by Fokker-Planck equation,
as characterized by the following two lemmas.

\begin{lemma}[$v(x,t)$ of Fokker-Planck flow]
\label{lem:force}
Suppose $x(t)$ solves \eqref{eq:ode-Xt} for $t \in [0,\infty)$
where the velocity field $v(x,t)$ is as in \eqref{eq:f-jko-flow},
and  $x(0) \sim \calN(0,\Sigma)$,  then, define $s :=1 - e^{-2t}$, 
\begin{equation} \label{force_general_form}
    v (x,t)=-(I -  ( s I+ (1-s) \Sigma )^{-1})x =: T_t^{\rm FK} x.
\end{equation}
\end{lemma}

In particular, at $t=0$, $T_0^{\rm FK} + I = \Sigma^{-1}$.
As $t \to \infty$, $T_t^{\rm FK} \to 0$ exponentially fast.

\begin{lemma}[$v(x,t)$ of Benamou-Brenier flow]
\label{lem:force-BB}
Suppose $x(t)$ solves \eqref{eq:ode-Xt} for $t \in [0,1]$
where the velocity field $v(x,t)$ is the optimal $v$ in \eqref{eq:Benamou-Brenier},
and $x(0) \sim \calN(0,\Sigma)$, 
then 
\begin{equation}\label{eq:v-BB-lemma}
 v (x,t)= (t I + (1-t) \Sigma^{1/2})^{-1} (I - \Sigma^{1/2}) x =: T_t^{\rm BB} x.
\end{equation}
\end{lemma}

In particular, $T_0^{\rm BB} = \Sigma^{-1/2} - I$,
and $T_1^{\rm BB} = I - \Sigma^{1/2}$.

The velocity field $v(x,t)$ in both cases is a linear mapping of $x$ depending on $t$, and thus it can always be exactly expressed by a fully-connected layer. 
For GNN layers, the problem is to represent the matrix $T_t$ in both cases by a graph convolution.
Because the desired operator itself is linear, we only consider the graph convolution in space 
(there is no channel mixing parameter because the node feature dimension $d'=1$ here),
omitting the bias vector and the non-linear activation.

\subsection{Spectrum of $\Sigma$ and an approximation lemma}

The approximation error will depend on the condition number of $\Sigma$, 
which reveals the fundamental difficulty of approximation when $\Sigma$ is near singular 
(the operators $ T_t^{\rm FK}$ and $ T_t^{\rm BB}$
 involves $\Sigma^{-1}$ at $\Sigma^{-1/2}$ at $t=0$ respectively). 
Define $S:= \Sigma^{-1}$ as the precision matrix of the distribution of $x(0)$.
Denote the condition number of $\Sigma$ as 
$\kappa := {\rm Cond}(\Sigma) < \infty$.
$\kappa = d_{\rm max}/d_{\rm min} \ge 1$, where $d_{\rm max}$ and $d_{\rm min}$ are the largest and smallest eigenvalues of $\Sigma$.
Suppose the data is properly normalized and without loss of generality, we divide $\Sigma$ by the constant $\sqrt{  d_{\rm max} d_{\rm min} }$, 
which makes the smallest and largest eigenvalue equal $1/\sqrt{\kappa}$ and $\sqrt{\kappa}$ respectively. 
Denote the spectrum of a real symmetric matrix $A$  (i.e., the set of $n$ eigenvalues) as ${\rm spec}(A)$.
Because the eigenvalues of $S$ are the reciprocal of those of $\Sigma$, we have made
\begin{equation}\label{eq:spec-Sig-S}
{\rm spec}( \Sigma ),  {\rm spec}( S ) \subset [ 1/\sqrt{\kappa}, \sqrt{\kappa} ].
\end{equation}
Meanwhile, we also have
\begin{equation}\label{eq:spec-Sig-S-1/2}
{\rm spec}( \Sigma^{1/2} ),  {\rm spec}( S^{1/2} ) \subset [ 1/\sqrt{\kappa'}, \sqrt{\kappa'} ], \quad \kappa' := \sqrt{\kappa}.
\end{equation}

We introduce a function approximation lemma which will be used in the analysis of both $T_t^{\rm FK}$ and $T_t^{\rm BB}$.  Define
 \begin{equation}\label{eq:def-fs-gs}
 \begin{split}
 f_s(x) &: = \left( s + (1-s) x \right)^{-1},\\
 g_s(x) &: =  \left( s + (1-s) x^{-1} \right)^{-1},
 \end{split}
 \quad  0 \le s  \le 1.
 \end{equation}

\begin{lemma}\label{lemma:fs-gs-approx}
For any $a \ge 1$,  let $b = ( 1/a + a )/2$, which satisfies $1\le b \le a$. 

(i) Short time. 
For all $s  \in [0, 1/2] $, and $n \ge 1$, 
there is a polynomial $Q_s^{(n+1)}(x)$ of degree at most $(n+1)$, where coefficients depending on $s$,  s.t. 
$\forall s \in [0, 1/2]$, 
\begin{equation}
\sup_{x \in [ 1/a, a ]} | g_s(x) -  Q_s^{(n+1)}(x) | \le a  b  \left(\frac{b}{b+1} \right)^n.
\end{equation}

(ii) Long time. For all $ s \in [1/2, 1]$ and $n \ge 1$,  
there is a polynomial $P_s^{(n)}(x)$ of degree $n$, where coefficients depending on $s$,  s.t. $\forall s \in [1/2, 1]$,
\begin{equation}
\sup_{x \in [ 1/a, a ]} | f_s(x) -  P_s^{(n)}(x) | \le  b  \left(\frac{b}{b+1} \right)^n.
\end{equation}
\end{lemma}
In below, we first introduce the approximation of $T_t^{\rm FK}$ by spectral and spatial graph convolutions.
The approximation of $T_t^{\rm BB}$ uses similar techniques and will be included afterward.

\subsection{Approximating $T_t^{\rm FK}$ by spectral graph convolution}

By Lemma \ref{lem:force}, with $s =1 - e^{-2t} \in [0, 1)$, we have
$ T_t^{\rm FK} + I = ( s I+ (1-s) \Sigma )^{-1} =: { T}_t'$, and it is equivalent to approximate ${ T}_t'$ by a graph convolution. 
By definition,
\begin{equation}\label{eq:Tt-by-fs-gs}
{ T}_t'  =  f_s(\Sigma) = g_s (S).
\end{equation}
Denote by $\tilde{L}$ the (possibly normalized) graph Laplacian matrix, which is a real symmetric PSD  matrix.

\begin{assumption}[$\delta$-approximation by spectral convolution]\label{spectral_assume}
Assuming \eqref{eq:spec-Sig-S}, 
there exist polynomial $p$ and $q$ of degree at most $n_0$, 
s.t. $\tilde{\Sigma} = p(\tilde{L})$ and  $\tilde{S} = q(\tilde{L})$ satisfy that

(C1) $ {\rm spec}( \tilde{\Sigma} ),  {\rm spec}( \tilde{S} ) \subset [ 1/\sqrt{\kappa}, \sqrt{\kappa} ]$. 

(C2)  $\| \tilde{\Sigma} - {\Sigma} \|_{op},  \| \tilde{S} - S \|_{op} \le \delta \sqrt{\kappa}$.
\end{assumption}

The assumption means that both $\Sigma$ and $S=\Sigma^{-1}$ can be approximated by some low degree polynomial of $\tilde{L}$ up to relative error $\delta$. 
This can be the case, for example, when $\Sigma$ shares the eigenvectors with $\tilde{L}$, and the polynomial $p$ and $q$ can be constructed to fit the eigenvalues. 
The condition (C1) may be relaxed by enlarging the interval $[ 1/\sqrt{\kappa}, \sqrt{\kappa} ]$ to be $[ 1/a, a ]$ where $a$ is a multiple of $\sqrt{\kappa}$,
and then construct the polynomial approximation of $g_s$ and $f_s$ on $[ 1/a, a ]$ as in Lemma \ref{lemma:fs-gs-approx}, which results in an $O(1)$ constant factor in the bound. In this case, (C1) may also be induced by (C2) with $O(1)$ constant $\delta$. We keep the current form of (C1)(C2)  for simplicity. 

The following theorem constructs the throughout-time approximation of ${ T}_t^{\rm FK}$ by using the composed polynomials
$Q_s^{(n+1)} \circ q$ and  $P_s^{(n)}\circ p$ on short and long times respectively.

\begin{theorem}[Spectral-convolution approximation]\label{thm:spectral_sigma}
Under Assumption \ref{spectral_assume}, $n_0$ and $\delta$ as therein, and $t_0 = \log \sqrt{2}$.

(i) For all $ t \in [0, t_0]$ and $n \ge 1$, 
there is a polynomial $q_t (x)$ of degree at most $(n_0+n+1)$, where coefficients depending on $t$,  s.t. 
\begin{equation}\label{eq:bound-short-time}
\| { T}_t' -  q_t( \tilde{L}) \|_{op} 
\le 2 \delta\sqrt{\kappa} +  \kappa  \exp\{ - \frac{n}{\sqrt{\kappa}+1} \}.
\end{equation}

(ii) For all $ t \ge t_0$ and $n \ge 1$, 
there is a polynomial $p_t(x)$ of degree at most $(n_0+n)$, where coefficients depending on $t$,  s.t. 
\begin{equation}\label{eq:bound-long-time}
\| { T}_t' -   p_t(\tilde{L}) \|_{op} 
\le 2 \delta \sqrt{\kappa} + \sqrt{ \kappa }  \exp\{ - \frac{n}{\sqrt{\kappa}+1} \}.
\end{equation}
\end{theorem}
The theorem suggests that when the covariance matrix $\Sigma$ is is well-conditioned, 
then small approximation error can be achieved using a low-degree $n$.

\vspace{-5pt}
\subsection{Approximating $T_t^{\rm FK}$ by spatial graph convolution}

We consider approximating operator $T_t'$ by spatially local graph filters, 
such as L3Net layer \citep{cheng2021graph} as introduced in Section \ref{subsec:scale-to-graph}. 
We start by introducing the notion of locality on graph. 

\begin{definition}[$v$-locality]\label{def:v-local}
A matrix $A$ is $v$-local on the graph for integer $v \geq 1$ if
$\rev{A_{ij}}=0$ when $j \notin \mathcal{N}^{(v)}_i$,
where $\mathcal{N}^{(v)}_i$ denotes the set of $v$-th neighbors of node $i$ (assuming $i$ is in the first neighborhood of itself).
We say that a diagonal matrix is 0-local as a convention. 
\end{definition}
The neighborhood $\mathcal{N}^{(v)}_i$ denotes all nodes accessible from node $i$ within $v$ steps along adjacent nodes. 
By definition, if a matrix $B$ is $v$-local, then $B^k$ is ($kv$)-local. 

\begin{assumption}[$\delta$-approximation by local filter]\label{spatial_assume}
Assuming \eqref{eq:spec-Sig-S}, 
there exist $v$-local matrices $\tilde{\Sigma} $ and  $\tilde{S} $ which satisfy conditions (C1)(C2) of Assumption \ref{spectral_assume} with some $\delta$.
\end{assumption}

The spatial graph convolution operator can be written as  $ \sum_{r=1}^R c_r B_r$,
where $B_r$ are local filters on the graph, and $c_r$ are coefficients. 
The following theorem proves the approximation of $T_t'$,
achieving the same bound as in Theorem \ref{thm:spectral_sigma}.
The construction is by using powers of the local filters $\tilde{\Sigma} $ and  $\tilde{S} $ as $B_r$.

\begin{theorem}[Spatial-convolution approximation]
\label{thm:local_sigma} 
Under Assumption \ref{spatial_assume}, $v$ and $\delta$ as therein, and  $t_0 = \log \sqrt{2}$.

(i) For all $ t \in [0, t_0]$ and $n \ge 1$, 
there is a spatial graph convolution filter of rank at most $(n+2)$
with basis filter $B_k$ having $(kv)$-locality and coefficients $c_k(t)$, 
$k=0,\cdots, n+1$, s.t. 
$\| T_t' -  \sum_{k=0}^{n+1} c_k(t) B_k  \|_{op} $  satisfies the same bound as in \eqref{eq:bound-short-time}.

(ii) For all $ t \ge t_0$ and $n \ge 1$, 
there is a spatial graph convolution filter of rank at most $(n+1)$
with basis filter $B_k$ having $(kv)$-locality and coefficients $c_k(t)$, $k=0,\cdots, n$, s.t. 
$\| T_t' -  \sum_{k=0}^{n} c_k(t) B_k  \|_{op} $ satisfies the same bound as in \eqref{eq:bound-long-time}.
\end{theorem}

Note that the construction has time-independent $B_r$, which means that in the L3Net GNN layers, 
we can potentially share the basis filter $B_r$ across residual blocks throughout the $L$ blocks, which will further reduce model complexity.

\subsection{Approximating $T_t^{\rm BB}$ by graph convolutions}

We take $s = t$ the time on $[0,1]$.  By Lemma \ref{lem:force-BB},  $T_t^{\rm BB}$ has the following short and long time representation, 
\[
\begin{split}
 T_t^{\rm BB} + \frac{1}{1-t} I &= \frac{1}{1-t} g_t( S^{1/2}),  \quad t \in [0,1/2], \\
 T_t^{\rm BB}  - \frac{1}{t}  I  &=  - \frac{1}{t}  \bar{f}_t (\Sigma^{1/2}), \quad t \in [1/2,1],
\end{split}
\]
where $\bar{f}_s (x) : = x  f_s(x)$, and $f_s$, $g_s$ are defined same as before \eqref{eq:def-fs-gs}. 
Let $ \tilde{\Sigma}^{1/2} $ and $ \tilde{S}^{1/2} $ are the approximators 
by some low-degree polynomials of $\tilde{L}$ in spectral convolution,
and by $v$-local graph filters in spatial convolution, respectively. 
Following \eqref{eq:spec-Sig-S-1/2}, 
we replace $\kappa$ to be $\kappa'$ in the analysis,
and the conditions (C1)(C2) in Assumptions \ref{spectral_assume}-\ref{spatial_assume} become

\vspace{5pt}
(C1') $ {\rm spec}( \tilde{\Sigma}^{1/2} ),  {\rm spec}( \tilde{S}^{1/2} ) \subset [ 1/\sqrt{\kappa'}, \sqrt{\kappa'} ]$. 

(C2')  $\| \tilde{\Sigma}^{1/2} - {\Sigma}^{1/2} \|_{op},  \| \tilde{S}^{1/2} - S^{1/2} \|_{op} \le \delta \sqrt{\kappa'}$.
\vspace{5pt}

Noting that in the short and long time expression of $T_t^{\rm BB}$, the constant factor $1/(1-t)$ and $1/t$ are bounded by 2 respectively,
thus we aim to approximate $g_t( S^{1/2})$ and $\bar{f}_t (\Sigma^{1/2})$. 
In applying Lemma \ref{lemma:fs-gs-approx}(ii), 
the bound becomes $| \bar{f}_s(x) - x P_s^{(n)}(x) | \le a b (b/(b+1))^n$,
which raises the polynomial degree to be $(n+1)$, 
and has another factor of $a = \sqrt{\kappa'}$ in the bound. 
The analysis proceeds using the same technique, and the details are omitted.
The final approximation bounds in both the short-time ($t \in [0,1/2]$)
and long-time ($t \in [1/2,1]$) cases are 
\[
4 \delta\sqrt{\kappa'} +  2 \kappa'  \exp\{ - \frac{n}{\sqrt{\kappa'}+1} \} 
\sim  \delta \kappa^{1/4} + \kappa^{1/2}  \exp\{ - \frac{n}{ \kappa^{1/4}+1} \},
\]
which improves the dependence on condition number from $\kappa^{1/2}$ to $\kappa^{1/4}$.

\subsection{Comparison of spectral and spatial graph convolutions}\label{sec_compare_spectral_spatial}

The analysis above shows that,
for Gaussian field data $X$ on graph,
the ability of a GNN flow network to learn the normalizing flow
is determined by the ability of the graph convolution filters to approximate the covariance matrix and precision matrix (or their square roots) of $X$. 
In terms of the expressiveness of spectral and spatial filters,
first note that when the spectral graph convolution filters are local (e.g., ChebNet filters of low polynomial degree), they become a special case of the spatial filters. In this case, spatial filters are always more expressive.
Meanwhile, there can be local spatial filters that cannot be expressed by spectral filters \cite{cheng2021graph}. 
Here we provide an example of covariance matrix $\Sigma $ on a three-node graph
that cannot be represented by spectral graph filter due to constructed symmetry.

\begin{example}
\label{ex:cannot_learn}
Consider a graph with three nodes $\{1, 2, 3\}$ and two edges $\{(1,2), (2,3)\}$ between nodes. Self-loops at each node are also inserted. 
Let the covariance matrix $\Sigma$ and permutation matrix $\pi$ take the form 
\[
\Sigma:=\begin{bmatrix}
1 & \rho & 0\\
\rho & 0 & \rho_1\\
0 & \rho_1 & 1
\end{bmatrix}, \pi=\begin{bmatrix}
0 & 0 & 1\\
0& 1 & 0\\
1 & 0 & 0
\end{bmatrix}.
\]
One can see that $\pi \Sigma \pi^T \neq \Sigma$ if $\rho \neq \rho_1$. On the other hand, $\pi A \pi^T =A$ at this chosen permutation $\pi$, so that any spectral graph filter $f (A)$ as a matrix function of $A$ satisfies that $\pi f(A) \pi^T=f(A)$. As a result, $f(A)$ for any $f$ will make an $O(1)$  error in approximating $\Sigma$. 
Because (possibly normalized) graph Laplacian is either a polynomial of $A$ or preserves the same symmetry pattern as $A$, 
the issue happens with any spectral convolutional filter. 
\end{example}

This example will be empirically examined in 
Section \ref{sec:theory_connect} where we train iGNN models on simulated graph data, see 
\rev{Fig. \ref{example_one_verify}}. 
We find that for this example, using ChebNet layer fails to learn the generation flow, and switching to the L3Net layer resolves the issue. 
In other situations where spectral graph filters have sufficient expressiveness for graph data, the spectral GNN has the advantage of a lighter model size.

\section{Experiment}\label{sec:experiment}

We first examine the proposed iGNN model on simulated data, including large-graph data, in Section \ref{sec:simple_example}. 
We then apply the iGNN model to real graph data 
(solar ramping event data and traffic flow anomaly detection) in Section \ref{sec:complex_example}. 
In Section \ref{sec:theory_connect}, we study the effect of using different GNN layers in the iGNN model.
The experimental setup is introduced in Section \ref{subsec:exp-setup},
and further details and additional results are provided in Appendix \ref{append:exp_result}.
The code  is available at \url{https://github.com/hamrel-cxu/Invertible-Graph-Neural-Network-iGNN}.

\vspace{-5pt}
\subsection{Experiment setup}\label{subsec:exp-setup}

\noindent \textit{Baselines and evaluation metrics.} 
We consider three competing conditional generative models: 
(a) Conditional generative adversarial network (cGAN) \citep{cGAN}.
(b) Conditional invertible neural network with maximum mean discrepancy (\ciNNI) \citep{cinn1}.
(c) Conditional invertible neural network using normalizing flow (\ciNNII) \citep{cinn2}.
To quantify performance, we measure the difference between two distributions ($X|Y$ versus $\hat{X}|Y$ at different $Y$) by kernel maximum mean discrepancy (MMD) \citep{Gretton2012AKT} and energy statistics \citep{Szkely2013EnergySA}. 
Details of the MMD and energy statistics metrics are contained in Appendix \ref{sec:setup_details}.
We also provide qualitative comparison by visualizing the \rev{distribution of} generated data.

\vspace{5pt}
\noindent \textit{Data and ResNet architecture.} 
In the examples of graph data,
the number of graph nodes ranges from 3 to 500. 
All graphs are undirected and unweighted, with inserted self-loops. 
Regarding the ResNet block layer type:
for non-graph data, we use 2 fully-connected hidden layers of 64 neurons in all ResNet blocks.
For graph data, in each residual block,
we replace the first hidden layer to be a GNN layer
(either ChebNet or L3Net layer);
the second layer is a shared fully-connected layer that applies channel mixing across all the graph nodes (which can be viewed as a GNN layer with identity spatial convolution).
The activation function is chosen as ELU \citep{ELU} or LipSwish \citep{ResFlow}, which have continuous derivatives. 
The network is trained end-to-end with the Adam optimizer \citep{Kingma2015AdamAM}. 
Hyperparameter selection and additional results with different choices are described in Appendix \ref{append_hyperparam}.

\vspace{-5pt}
\subsection{Simulated examples}\label{sec:simple_example}

We consider three simulated examples in this section. The first considers non-graph Euclidean data, and the second and the third consider graph data on small and large graphs. Detailed data-generating procedures can be found in Appendix \ref{sec:setup_details}. 

\begin{table}[b]
\vspace{-10pt}
\centering
\cprotect \caption{Relative inversion error $\mathbb{E}_X[\|F\rev{_{\theta}}^{-1}(F\rev{_{\theta}}(X))-X\|_2]$ on the solar ramping event test data. Generative quality and data details are described in Section \ref{sec:complex_example}.
\vspace{-5pt}
}
\label{W2_inv_err}
\resizebox{0.8\linewidth}{!}{%
\begin{tabular}{l | cccccc}
\toprule
$\gamma$ & 0 & 0.5 & 1 & 2 & 5 & 10 \\
\hline
Inversion error & 4.09e+04 & 2.74e-06 & 1.03e-06 & 3.14e-06 & 2.61e-06 & 1.60e-06 \\
\bottomrule
\end{tabular}
}
\end{table}

\begin{figure}[!t]
\centering
\begin{minipage}[b]{0.4\linewidth}
    \includegraphics[width=\linewidth]{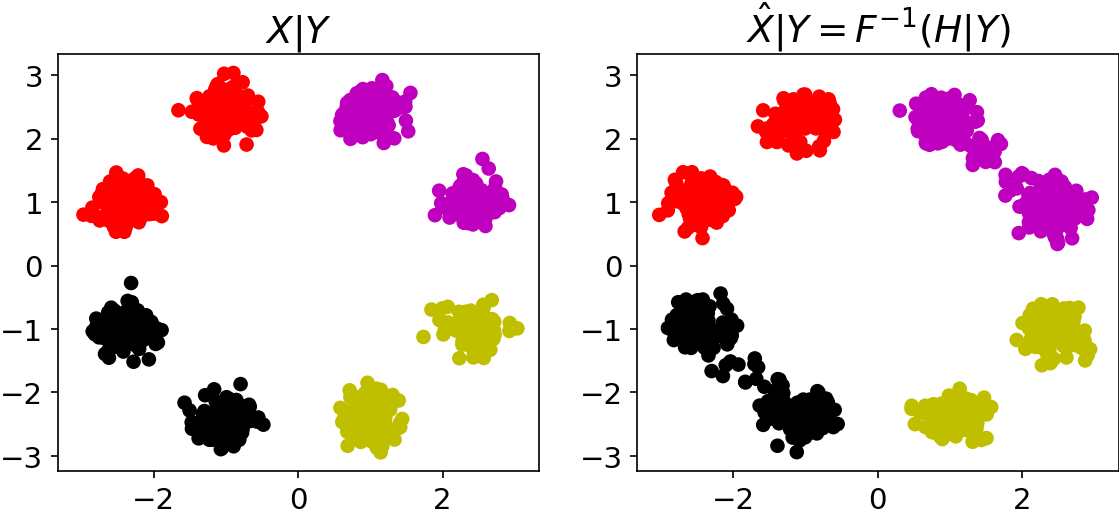}
    \vspace{-0.15in}
    \subcaption{$X|Y$ (left) and iGNN $\hat{X}|Y$ (right)}
\end{minipage}
\begin{minipage}[b]{0.175\linewidth}
    \includegraphics[width=\linewidth]{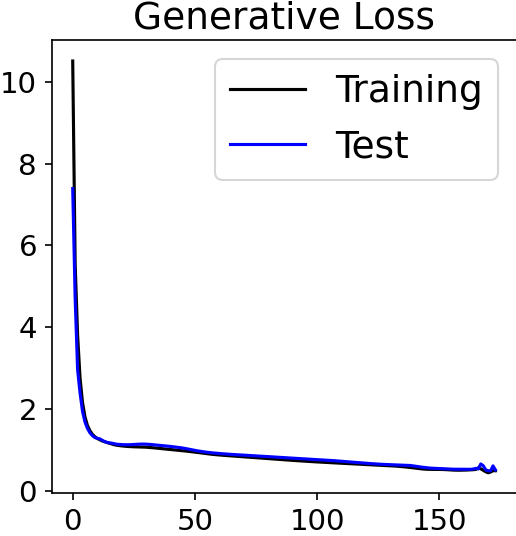}
    \vspace{-0.15in}
    \subcaption{iGNN loss}
\end{minipage}
\begin{minipage}[b]{0.19\linewidth}
    \includegraphics[width=\linewidth]{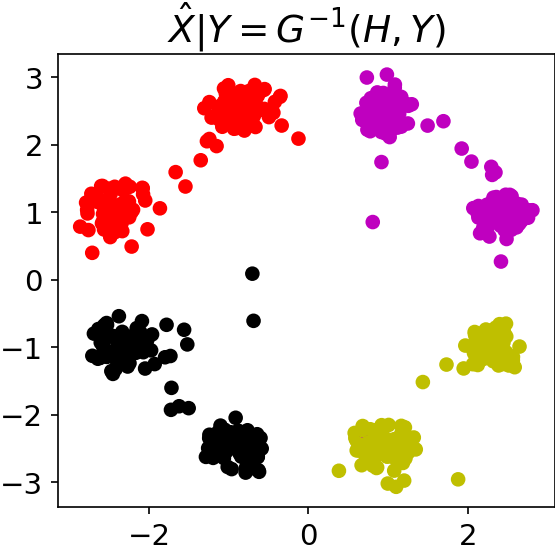}
    \vspace{-0.15in}
    \subcaption{cINN $\hat{X}|Y$}
\end{minipage}
\begin{minipage}[b]{0.175\linewidth}
    \includegraphics[width=\linewidth]{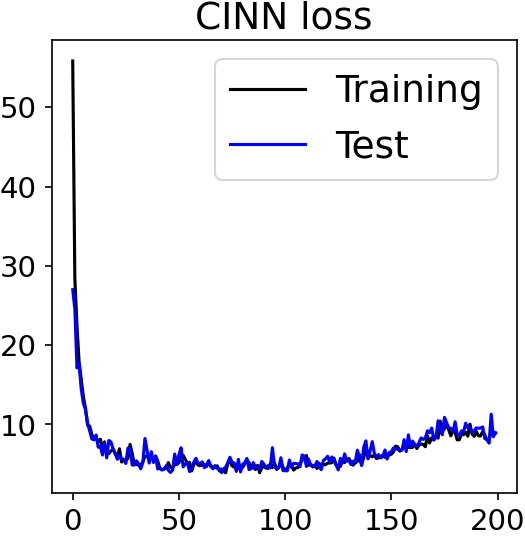}
    \vspace{-0.15in}
    \subcaption{cINN loss}
\end{minipage}
\caption{Compare iGNN vs. \ciNNI\ on simulated data in $\R^2$,
the data observes a mixture model having eight components but are attributed to four classes (indicated by color). The four classes of input data remain separated, cf. Assumption \ref{assump:same-flow}.
\vspace{-10pt}}
\label{8_gaussian}
\end{figure}

 \vspace{5pt}
\noindent \textit{1. Non-graph data in $\R^2$.} The dataset contains a Gaussian mixture of eight components with four classes, where each $X|Y$ is further divided into two \rev{separated} Gaussian distributions in $\R^2$. Fig. \ref{8_gaussian} compares the generative results of the proposed iGNN model with \ciNNI, where both methods can generate data that are reasonably close to $X|Y$ at each $Y$.

\vspace{5pt}
\noindent \textit{2. Data on a small graph.} 
We consider the three-node graph introduced in Example \ref{ex:cannot_learn}. 
At each node $v$, $Y_v \in \{0,1\}$ and features $X_v\in \R^2$,
thus the graph node label vector $Y\in \{0,1\}^3$. 
In this example, iGNN yields comparable generative performance with \ciNNI,
\rev{where both models use L3net GNN layers}.
See Fig. \ref{3node_nonconvex} and more results in appendix \ref{append:exp_result}.

\begin{figure}[!t]
\vspace{-10pt}
    \centering
    \begin{minipage}{0.15\linewidth}
        \includegraphics[width=\linewidth]{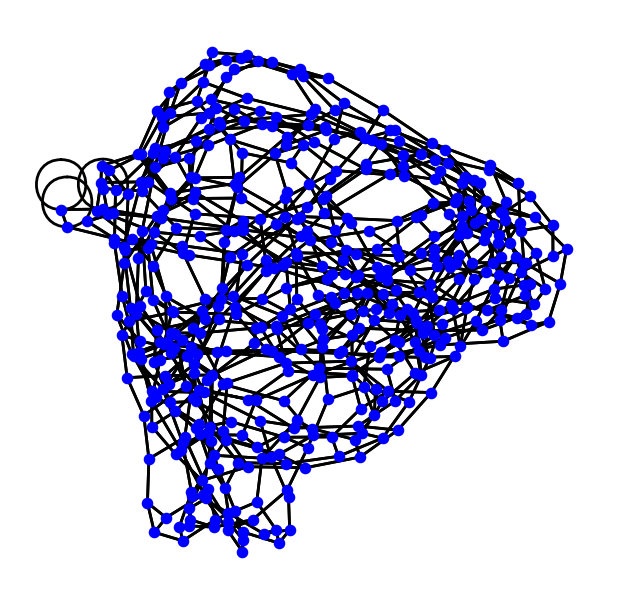}
        \subcaption{Graph}
    \end{minipage}
    \begin{minipage}{0.4\linewidth}
        \includegraphics[width=\linewidth]{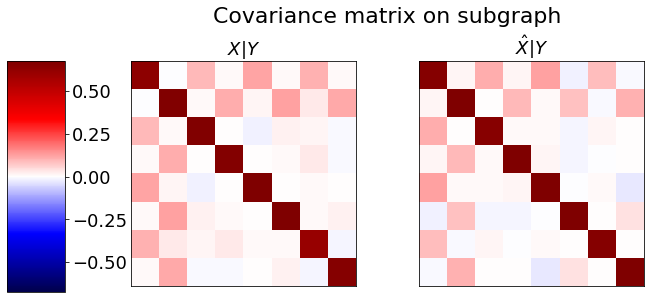}
        \subcaption{
       1-hop neighborhood of node 100 (4 nodes)}
    \end{minipage}
    \begin{minipage}{0.4\linewidth}
        \includegraphics[width=\linewidth]{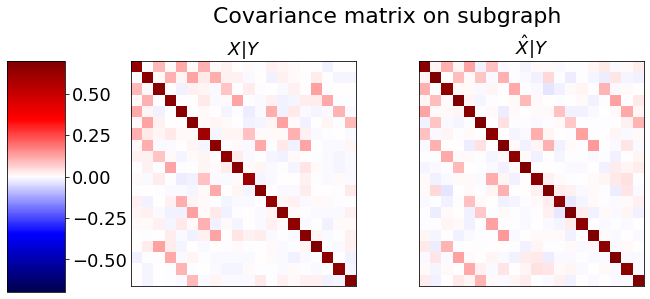}
        \subcaption{2-hop neighborhood of node 100 (10 nodes)}
    \end{minipage}

    \hspace{1in}
    \begin{minipage}{0.4\linewidth}
        \includegraphics[width=\linewidth]{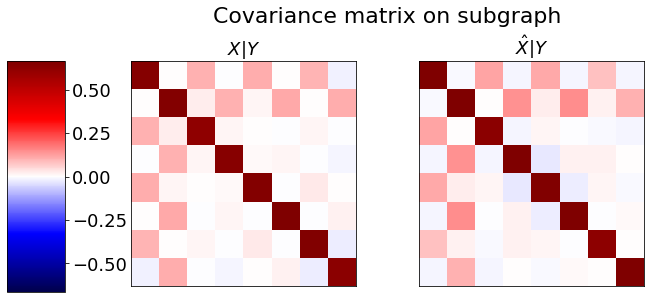}
        \subcaption{1-hop neighborhood of node 500 (4 nodes)}
    \end{minipage}
    \begin{minipage}{0.4\linewidth}
        \includegraphics[width=\linewidth]{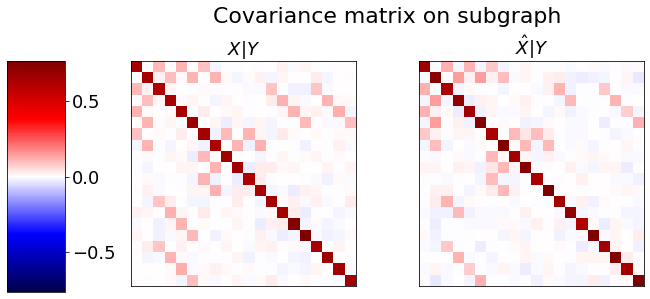}
        \subcaption{2-hop neighborhood of node 500 (10 nodes)}
    \end{minipage}
    \caption{
    Generating performance by iGNN of graph data $X|Y$ on a 503-node chordal cycle graph, where the node feature dimension $d' =2$, and the per-node class number $K=2$.
    To evaluate the conditional generation quality,
    we plot the
    covariance matrix of model-generated data $\hat{X}|Y$ 
    (right plot in (b)-(e))
    restricted to sub-graphs produced by 1 or 2-hop neighborhoods of a graph node
    in comparison with the ground truth
    (left plot in (b)-(e)).
     }
    \label{large_graph_cond_gen}
\end{figure}

 \vspace{5pt}
\noindent \textit{3. Data on a large graph.} 
 \rev{To demonstrate the scalability of our approach,} we consider a 503-node chordal cycle graph \citep{Lubotzky1994DiscreteGE}, which is an expander graph. We design binary node labels and let node features $X|Y \sim N(\mu_Y,\Sigma_Y)$, where $X_v \in \R^2$ and the mean $\mu_Y$ and covariance matrix $\Sigma_Y$ contain graph information. 
Because enumerating all values of $Y$ is infeasible, we randomly choose 50 values of outcome $Y$, each of which has 50\% randomly selected node labels to be 1. 
To visualize the generative performance of iGNN, we compare the covariance of true and generated data restricted to subgraphs. Specifically, we plot the covariance matrix of model-generated data $\hat{X}|Y$ and true data $X|Y$ on sub-graphs produced by 1 or 2-hop neighborhoods of a graph node. 
Fig. \ref{large_graph_cond_gen} shows the resemblance between learned and true covariance matrices on different neighborhoods on the graph.

\subsection{Real-data examples}\label{sec:complex_example}

We apply the iGNN model to two graph prediction data in real applications,
the solar ramping event data, 
and the traffic flow anomaly detection data.
The inverse prediction problem is formulated as a conditional generation task. 

\begin{figure}[!t]
\centering
\begin{minipage}[b]{0.2\linewidth}
    \includegraphics[width=\linewidth]{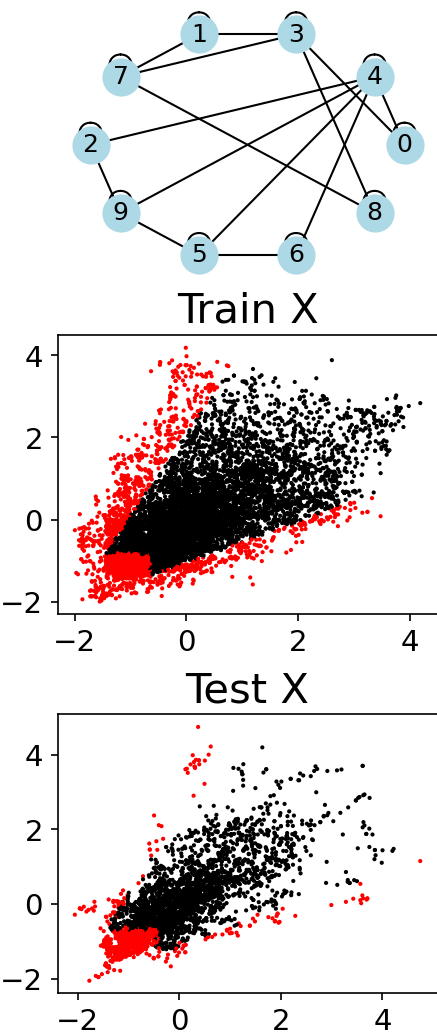}
\subcaption{Raw data}
\label{Solar_network}
\end{minipage}
\begin{minipage}[b]{0.34\linewidth}
    \includegraphics[width=\linewidth]{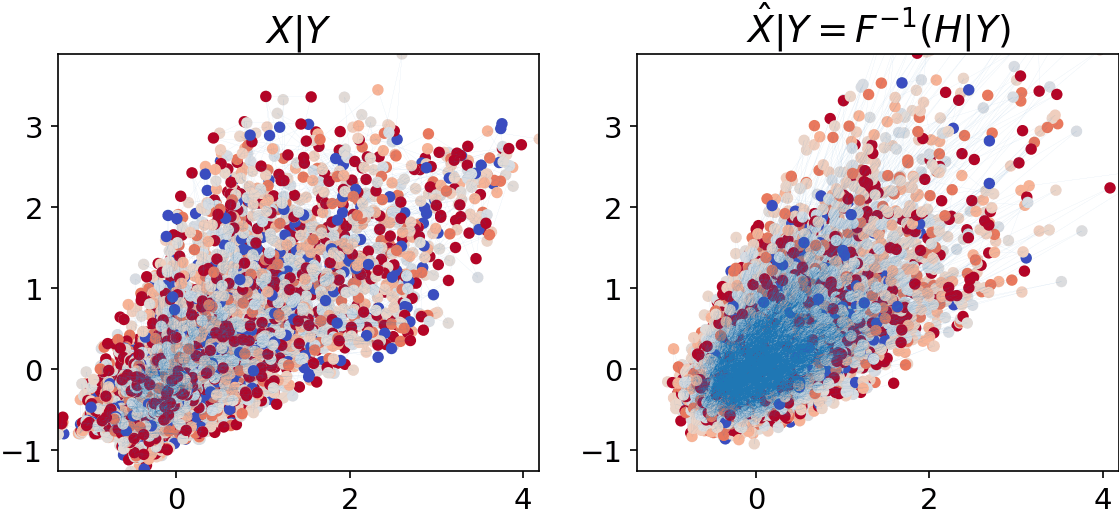}
    \includegraphics[width=\linewidth]{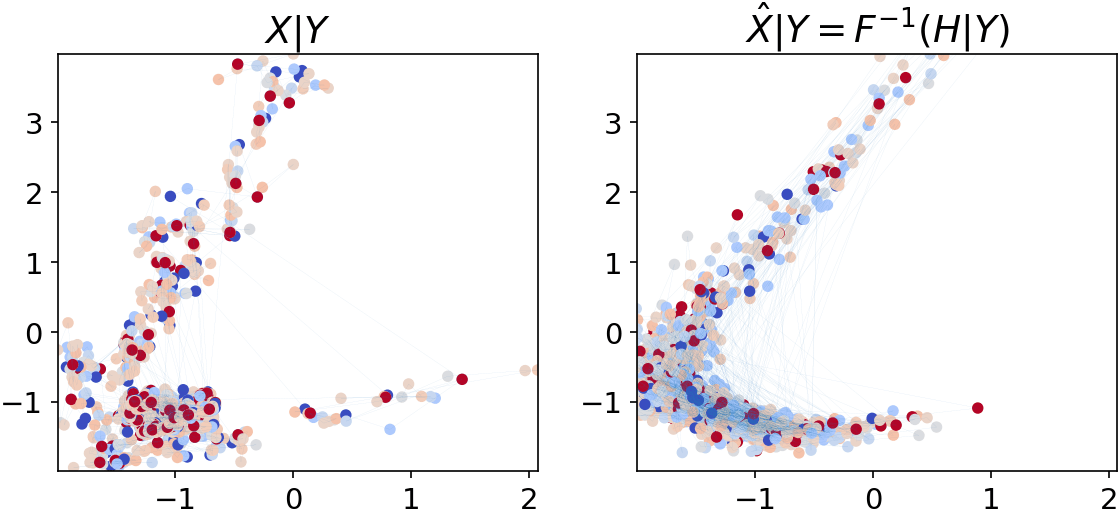}
    \includegraphics[width=\linewidth]{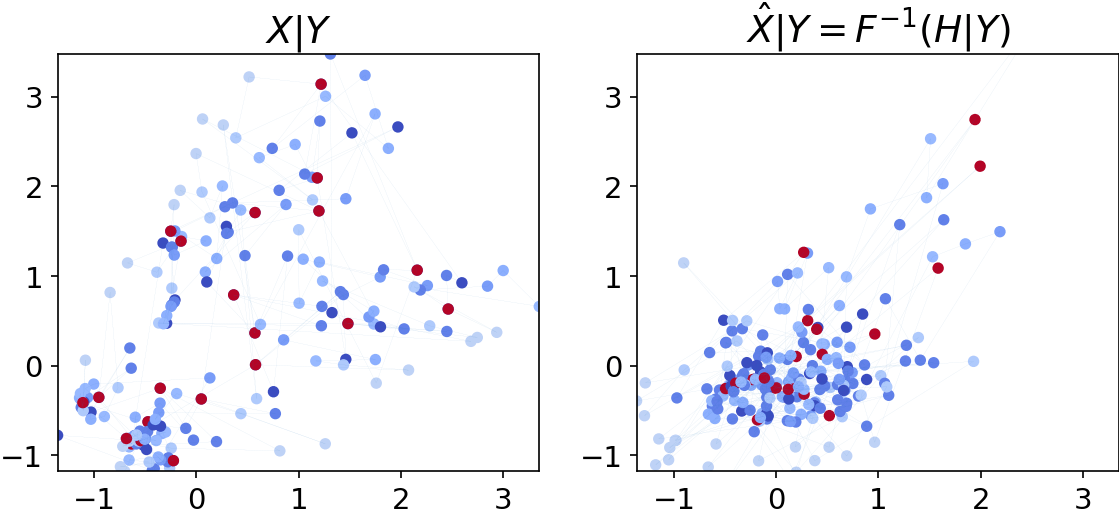}
    \vspace{-0.15in}
    \subcaption{$X|Y$ (left) and iGNN $\hat{X}|Y$ (right)}
\end{minipage}
\begin{minipage}[b]{0.158\linewidth}
    \includegraphics[width=\linewidth]{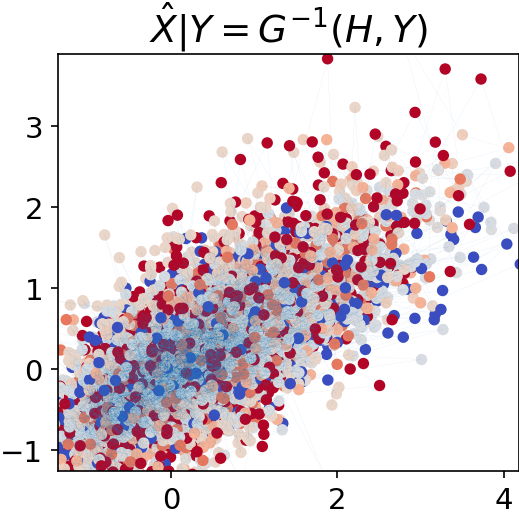}
    \includegraphics[width=\linewidth]{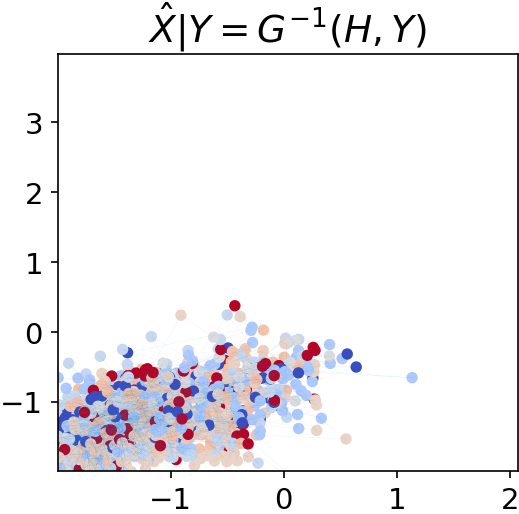}
    \includegraphics[width=\linewidth]{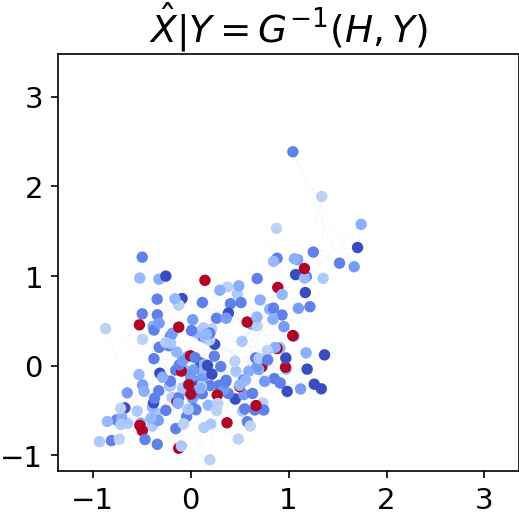}
    \vspace{-0.15in}
    \subcaption{cINN $\hat{X}|Y$}
\end{minipage}
\begin{minipage}[b]{0.233\linewidth}
    \includegraphics[width=\linewidth]{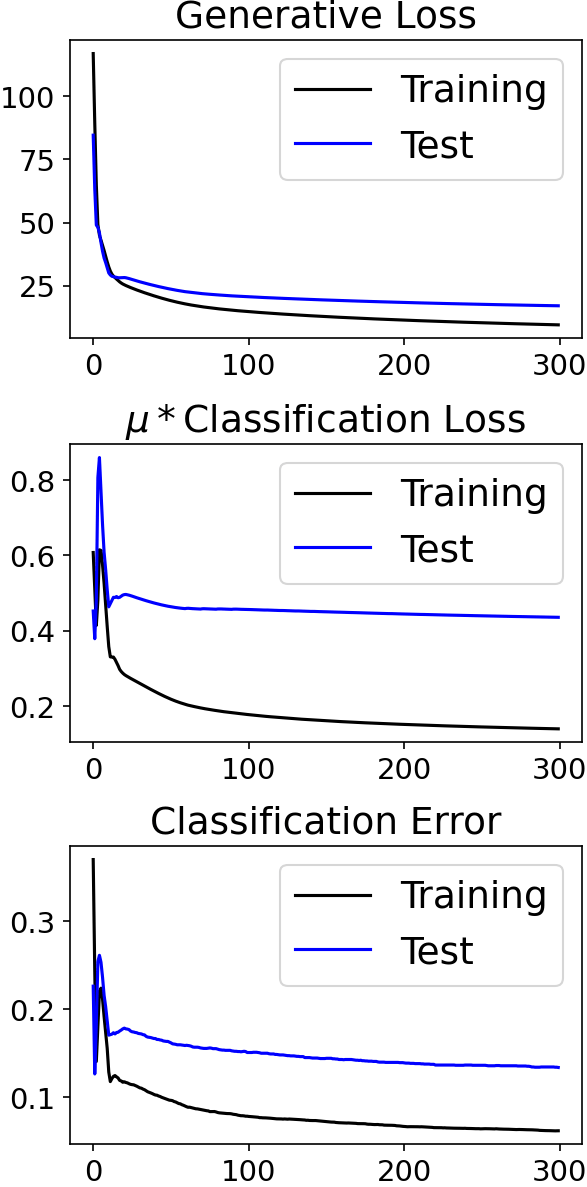}
    \vspace{-0.15in}
    \subcaption{iGNN loss}
\end{minipage}
\caption{
\vspace{-2pt}
Comparison of iGNN versus \ciNNI\ on solar ramping event data. The graph has 10 nodes with binary node response and node features in $\R^2$. \rev{(a) shows the scatter plot of nodal features across all nodes and all samples for training (upper panel) and testing data (lower panel), respectively.
(b)-(c) show the model generated $\hat{X}|Y$ in comparison to the ground truth $X|Y$, where three rows in the plot represent three chosen representative $Y$ (10-dimensional vector that contains nodal features over all nodes).
The scatter plots show the samples of two-dimensional nodal features, which are connected to each other by light-blue lines and are colored by empirical variances of each node. 
}
\vspace{-10pt}}
\label{Solar_training}
\end{figure}
\begin{figure}[!b]
\vspace{-5pt}
    \centering
    \begin{minipage}{0.32\linewidth}
        \includegraphics[width=\linewidth]{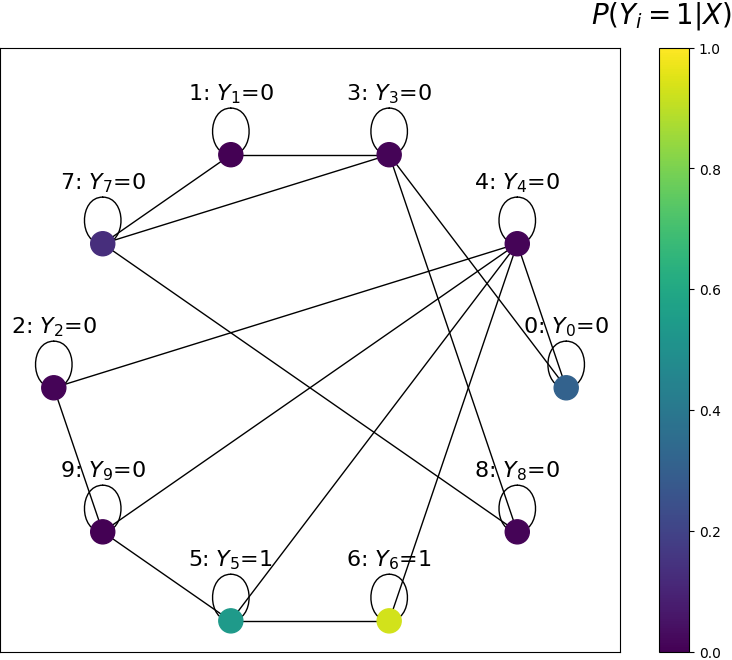}
    \end{minipage}
    \begin{minipage}{0.32\linewidth}
        \includegraphics[width=\linewidth]{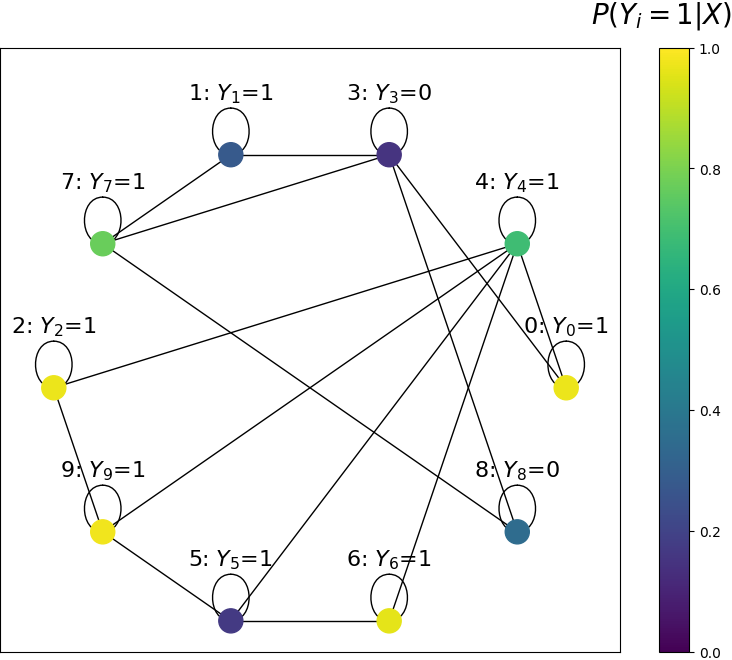}
    \end{minipage}
    \begin{minipage}{0.32\linewidth}
        \includegraphics[width=\linewidth]{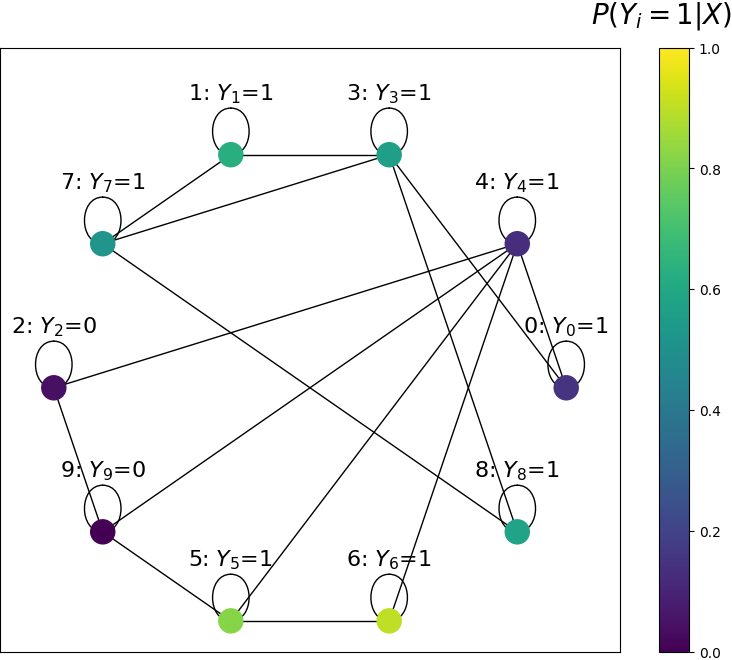}
    \end{minipage}
    \caption{
    Predicted probabilities of graph labels $Y$ by iGNN model
    where $Y$ takes different values on graph nodes on test data.
    Given a node feature matrix $X$, we compute $\mathbb{P}(Y_i=1|X)$ on each node using the linear classifier $f( \cdot; \theta_c^c)$ in $H$-$Y$ sub-network applied to the flow-mapped graph node feature $H=F_{\theta}(X)$. The true node label $Y_i$ is shown on top of each node in the plot.
    }
    \label{solar_prediction}
\end{figure}

\vspace{5pt}
\noindent \textit{
1. Solar ramping events data.} Consider the anomaly detection task on California solar data in 2017 and 2018, which were collected in ten downtown locations representing network nodes. Each node records non-negative bi-hourly radiation recordings measured in Global Horizontal Irradiance (GHI). After pre-processing, graph nodal features $X_t \in \R^{10\times 2}$ denote the average of raw radiation recordings every 12 hours in the past 24 hours, and response vectors $Y_t \in \{0,1\}^{10}$ contain the anomaly status of each city.
Fig. \ref{Solar_training} shows that the learned conditional distribution $\hat{X}|Y$ by iGNN model closely resembles that of the true data $X|Y$,
and outperforms the generation of \ciNNI. 
The quantitative evaluation is given in  Table \ref{real_data_metric},
which shows that iGNN has comparable or better performance than the alternative approaches (smaller test statistics indicate better generation). 
The table also shows that \ciNNII\ performs significantly worse than \ciNNI\ and iGNN on this example, 
which is consistent with the visual comparison of $\hat{X}|Y$ (not shown). 
Lastly, Fig. \ref{solar_prediction} shows the predictive capability of iGNN: given a test node feature matrix $X$, we can compute $\mathbb{P}(Y_i=1|X)$ for node $i$ using the trained linear classifier on $F_{\Theta}(X)$. 
The predicted probabilities learned by the model are consistent with the true nodal labels, and provide more information than the binary prediction output.

\begin{figure}[!t]
\centering
\begin{minipage}[b]{0.2\linewidth}
    \includegraphics[width=\linewidth]{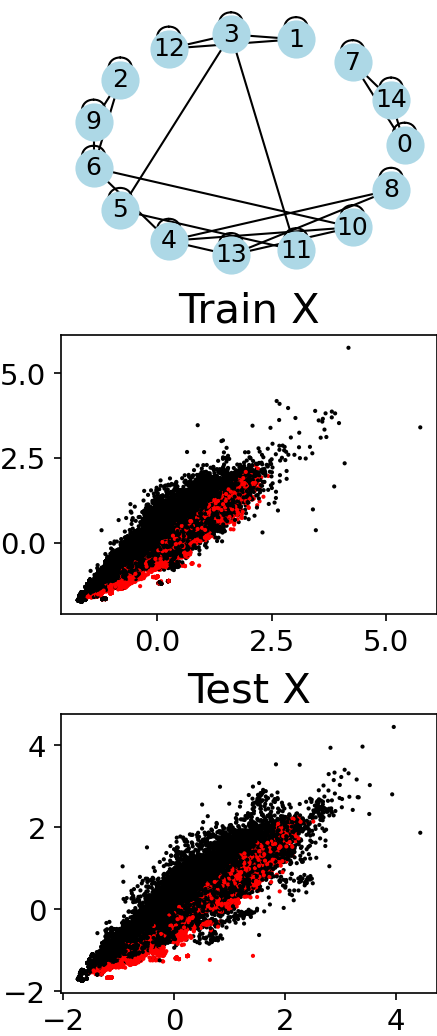}
\subcaption{Raw data}
\label{traffic_network}
\end{minipage}
\begin{minipage}[b]{0.34\linewidth}
    \includegraphics[width=\linewidth]{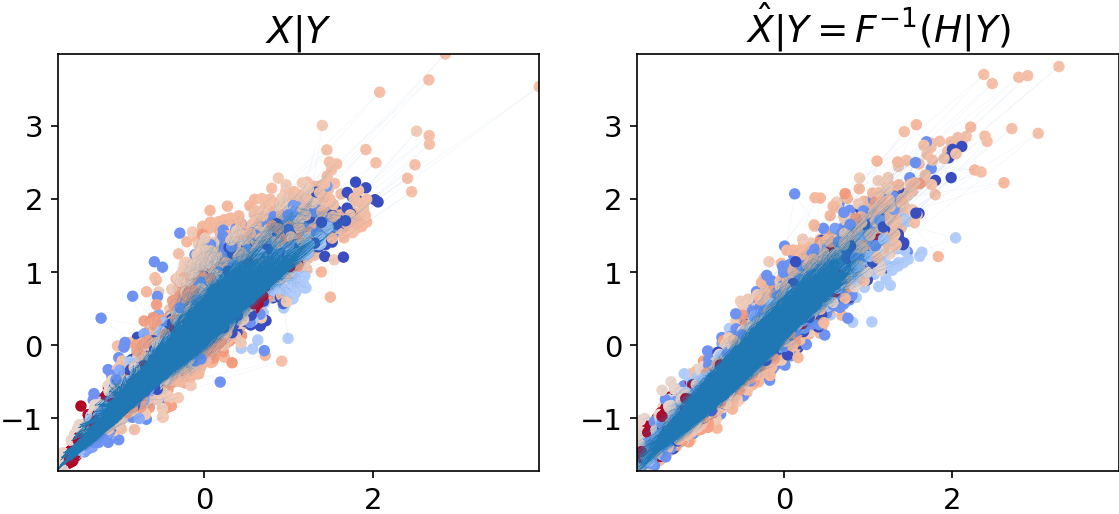}
    \includegraphics[width=\linewidth]{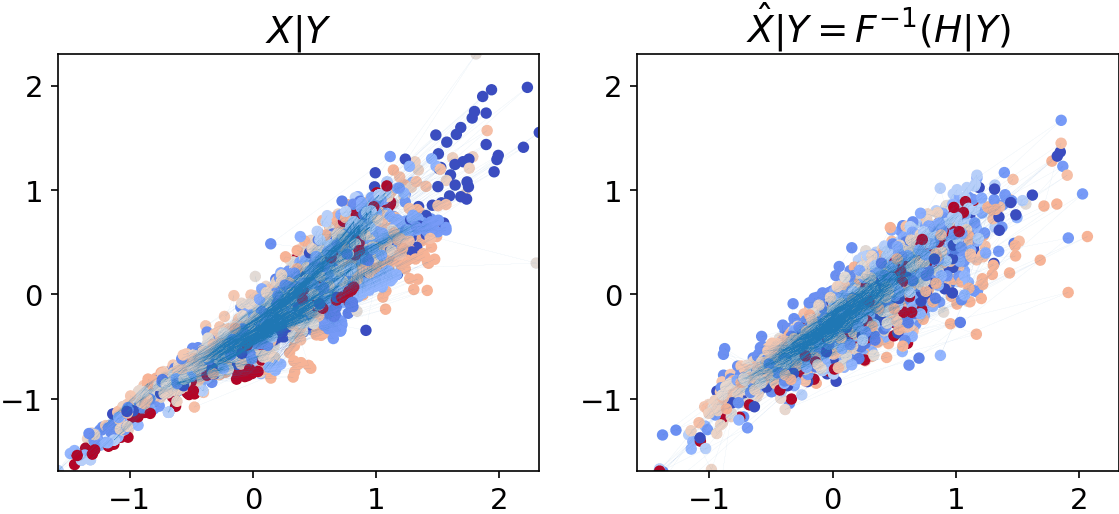}
    \includegraphics[width=\linewidth]{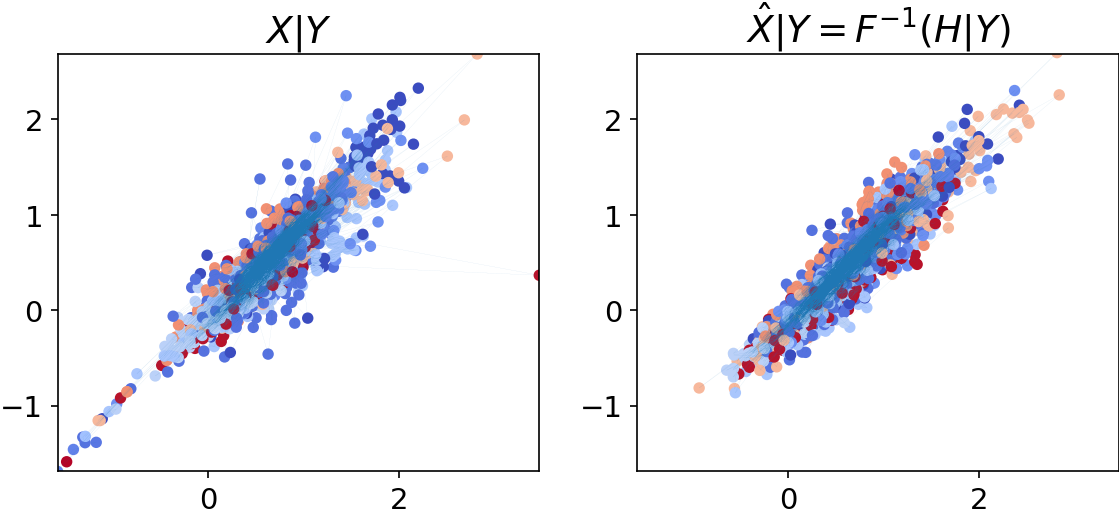}
    \vspace{-0.15in}
    \subcaption{$X|Y$ (left) and iGNN $\hat{X}|Y$ (right)}
\end{minipage}
\begin{minipage}[b]{0.158\linewidth}
    \includegraphics[width=\linewidth]{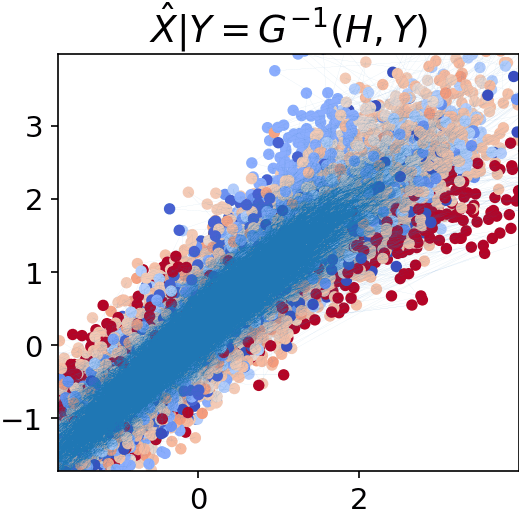}
    \includegraphics[width=\linewidth]{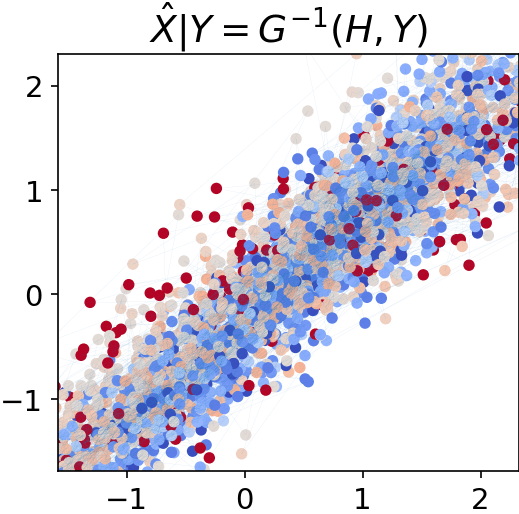}
    \includegraphics[width=\linewidth]{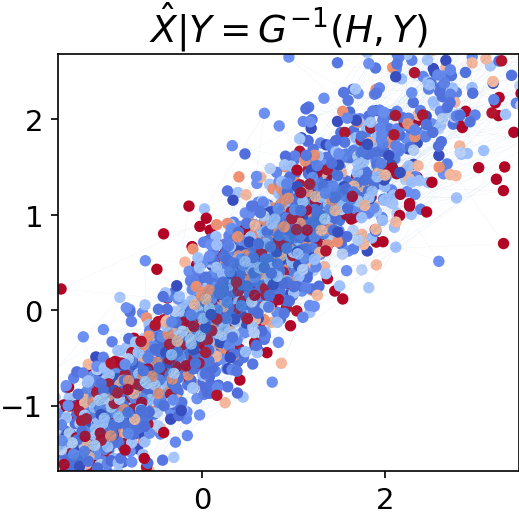}
    \vspace{-0.15in}
    \subcaption{cINN $\hat{X}|Y$}
\end{minipage}
\begin{minipage}[b]{0.233\linewidth}
    \includegraphics[width=\linewidth]{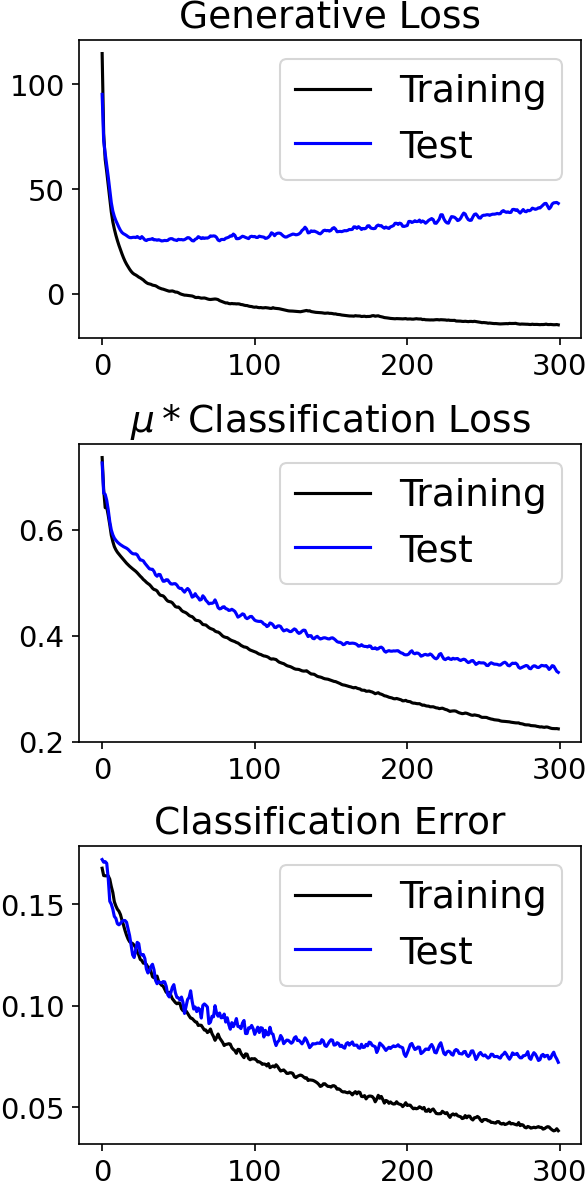}
    \vspace{-0.15in}
    \subcaption{iGNN loss}
\end{minipage}
\caption{
\vspace{-2pt}
Comparison of iGNN versus \ciNNI\ on traffic anomaly detection data.
Same plots \rev{arrangement} as in Fig. \ref{Solar_training}.
}
\label{traffic_training}
\end{figure}

 \vspace{5pt}
\noindent \textit{2. Traffic flow anomalies.} 
We study the anomaly detection task on Los Angeles traffic flow data from April to September 2019. The whole network has 15 sensors with hourly recordings. \rev{Graph nodal} features $X_t \in \R^{15 \times 2}$ denote the raw hourly recording in the past two hours, and response vectors $Y_t \in \{0,1\}^{15}$ contain the anomaly status of each traffic sensor. 
The graph topology is shown in Fig. \ref{traffic_network},
along with the raw input features in $\R^2$ (over all graph nodes).
The \rev{generated data distribution} by iGNN resembles the ground truth, as shown in Fig. \ref{traffic_training}(b), 
and \rev{the performance} is better than that of \ciNNI\ in (c). 
The quantitative evaluation metrics also reveal the better performance of iGNN over alternative baselines, cf. Table \ref{real_data_metric}.

\subsection{Comparison of GNN layers \rev{in iGNN models}}\label{sec:theory_connect}
\begin{table}[!b]
\centering
\cprotect 
\caption{Two-sample testing statistics on test data. We use the formula \eqref{weighted_MMD} for MMD statistics and \eqref{weighted_energy} for energy statistics.\vspace{-5pt}}
\label{real_data_metric}
\begin{center}
\resizebox{0.8\linewidth}{!}{%
\begin{tabular}{P{2.5cm}P{1.5cm}P{1.5cm}|P{2.5cm}P{1.5cm}P{1.5cm}}
\toprule
\textbf{Solar data} &  MMD &  Energy & {\textbf{Traffic data}} &  MMD  &  Energy \\
\midrule
iGNN &           0.062            &   \textbf{0.341} & iGNN   &           \textbf{0.128}  &   \textbf{0.537} \\
\ciNNI   &           \textbf{0.061} &   0.344 & \ciNNI   &           0.152 &   1.484 \\
\ciNNII &           0.402 &   3.488 & \ciNNII &           0.281 &   6.183 \\
cGAN        &           0.572 &   3.422  & cGAN        &           0.916 &   4.132 \\
\bottomrule
\end{tabular}
}
\end{center}
\end{table}

We examine the empirical performance of different GNN layers in learning the normalizing flow of graph data, for which the theoretical analyses appeared in Section \ref{sec:expressiveness}. 

\begin{figure}[t]
\vspace{-5pt}
\centering
\begin{minipage}[b]{0.20\linewidth}
    \includegraphics[width=\linewidth]{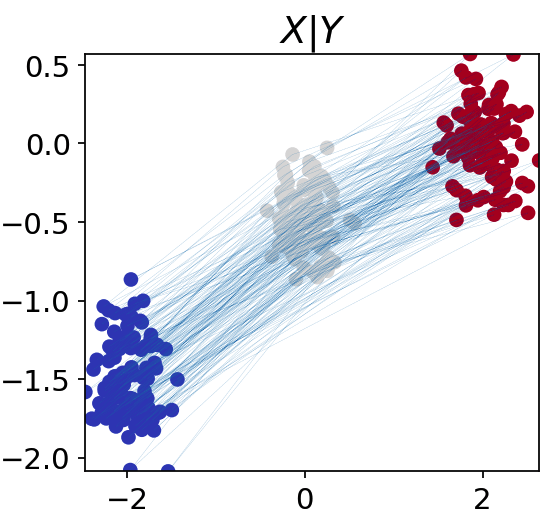}
    \includegraphics[width=\linewidth]{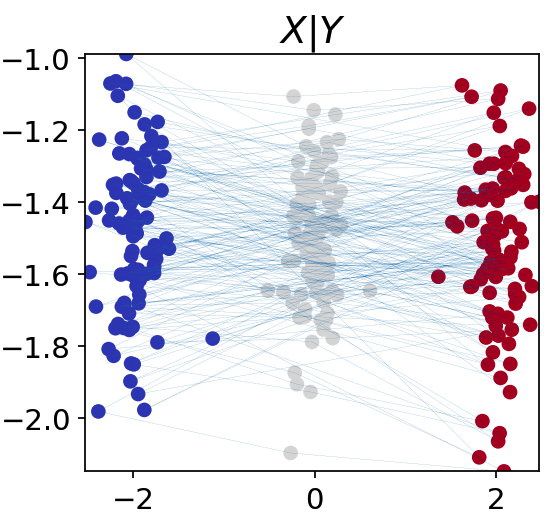}
    \subcaption{True data}
\end{minipage}
\begin{minipage}[b]{0.20\linewidth}
    \includegraphics[width=\linewidth]{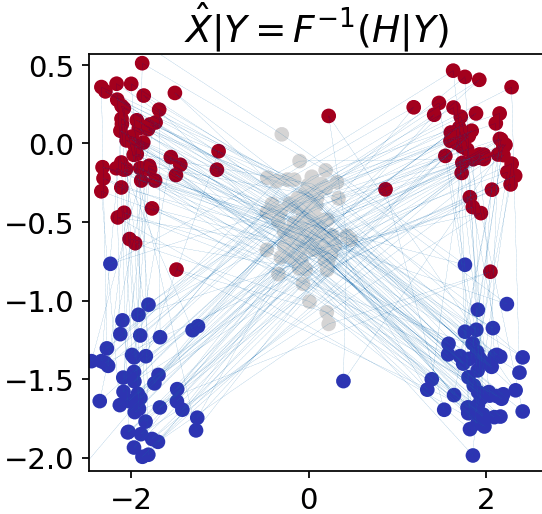}
    \includegraphics[width=\linewidth]{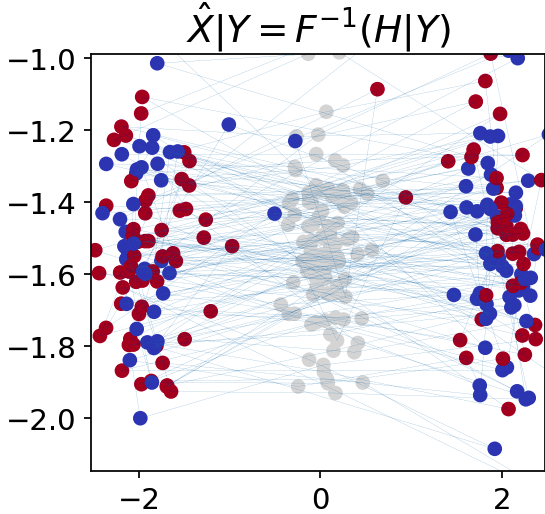}
    \subcaption{ChebNet}
\end{minipage}
\begin{minipage}[b]{0.184\linewidth}
    \includegraphics[width=\linewidth]{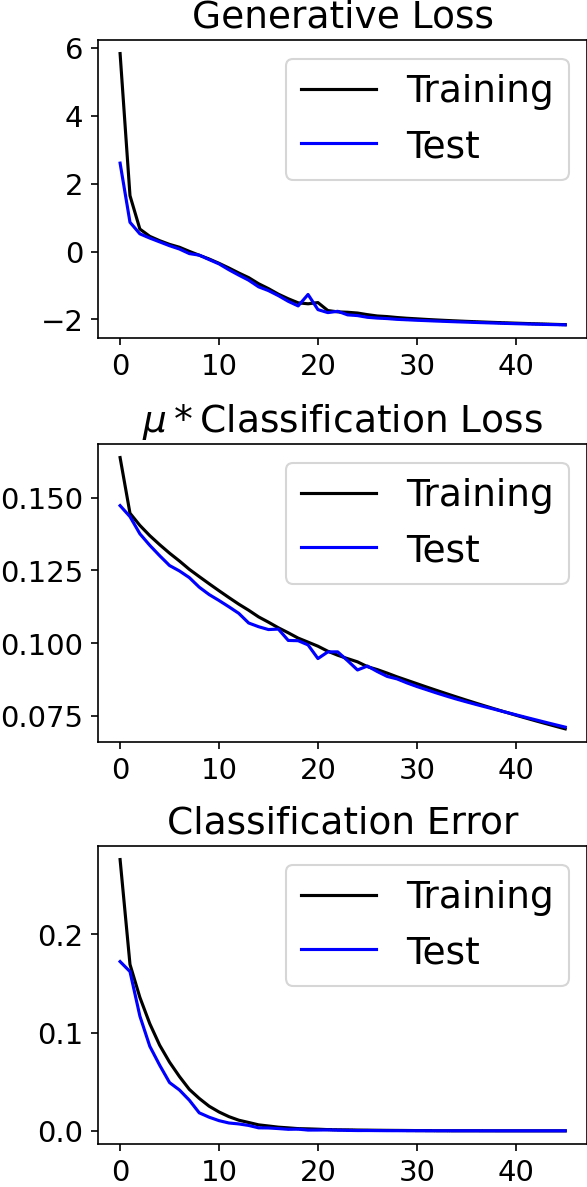}
      \subcaption{ChebNet losses}
\end{minipage}
\begin{minipage}[b]{0.20\linewidth}
    \includegraphics[width=\linewidth]{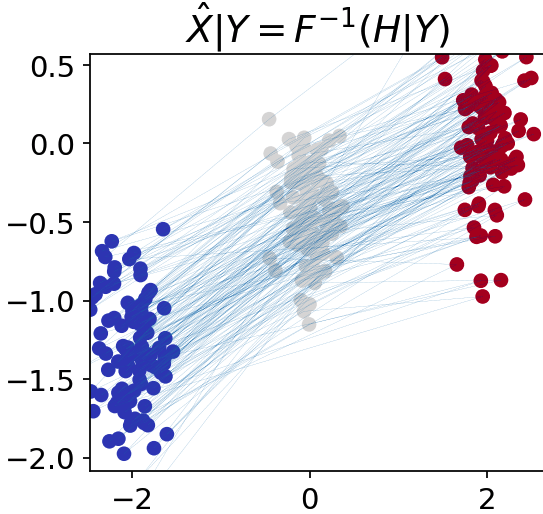}
    \includegraphics[width=\linewidth]{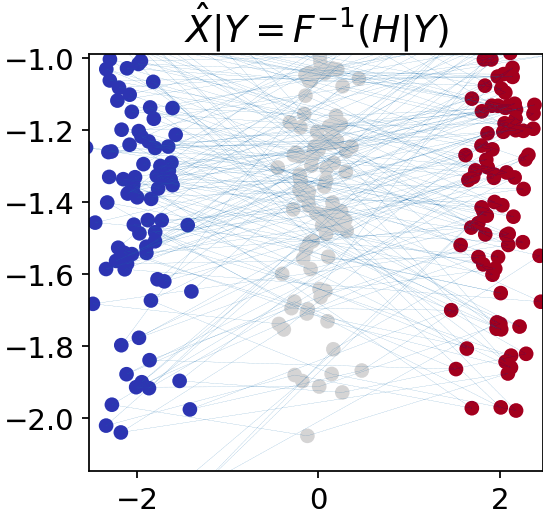}
    \subcaption{L3Net}
\end{minipage}
\begin{minipage}[b]{0.184\linewidth}
    \includegraphics[width=\linewidth]{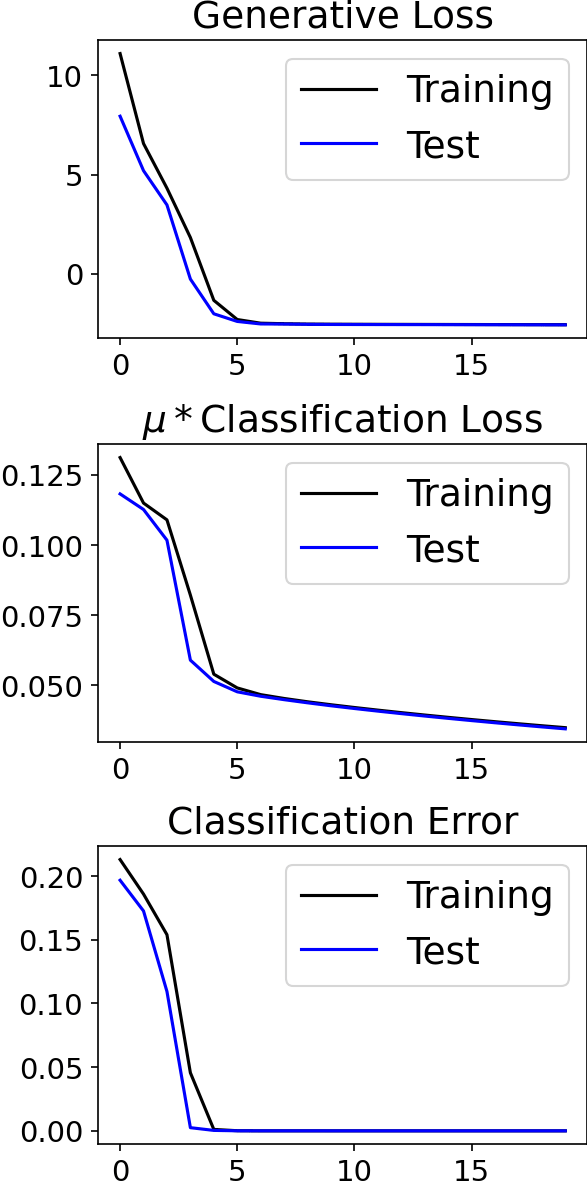}
      \subcaption{L3Net losses}
\end{minipage}
\caption{Comparison of using spectral and spatial GNN layer in iGNN model for conditional generation. Data in (a) are generated 
as two-dimensional graph node features lying on a three-node graph with binary node labels; \rev{colors indicate the node, and the gray lines connect three nodal features associated with the same data instance. The top row has $Y=[1,1,0]$ and the bottom row has $Y=[1,1,1]$}. We visualize samples generated by iGNN using the spectral GNN layer (ChebNet) in (b) and spatial GNN layer (L3Net) in (d), as well as the losses over training epochs in (c) and (e). 
\vspace{-10pt}}
\label{3node_convex_append}
\end{figure}

\vspace{2pt}
\noindent \textit{1. Simulated data \rev{on three-node graph}.} 
\rev{
To validate our theory, we study two simulated datasets
to show the possible insufficiency of spectral GNN layers, as has been explained in Section \ref{sec_compare_spectral_spatial}.}
\rev{The graph data are on a three-node graph:}
the first one is as in Example \ref{ex:cannot_learn},
where nodal feature dimension $d'=1$, and $K=1$;
the second one has nodal feature dimension $d'=2$,
and $K=2$, see more details in Appendix \ref{sec:setup_details}. 
We compare the generative performance of iGNN by using ChebNet and L3Net layers for both examples. 
The result on the $d'=2$ example is shown in Fig. \ref{3node_convex_append},
where iGNN with ChebNet layers fails to learn the conditional distribution $X|Y$, \rev{cf. plot (b).}
Meanwhile, iGNN with L3Net layers yields satisfactory performance by having sufficient model expressiveness for this example. 
\rev{The second experiment studies Example \ref{ex:cannot_learn} considered in Section \ref{sec_compare_spectral_spatial}.
The results are shown in Fig. \ref{example_one_verify},
where, similarly, iGNN with ChebNet layers fails to generate the graph data $X$ and switching to L3Net layers resolves the insufficiency of expressiveness. }

\vspace{5pt}
\noindent \textit{2. Simulated large-graph data.}
\rev{We also compare the GNN layers on a larger graph} using the same 503-node chordal cycle graph as in Fig. \ref{large_graph_cond_gen},
and simulate \rev{$X\sim \calN(0,\Sigma)$},
where $\Sigma^{-1}=\sum_{k=0}^2 a_k T_k(\widetilde{L})$, 
 $T_k$ is the $k$-th Chebyshev polynomial,
 \rev{$\widetilde{L} = I - D_A^{-1/2}AD_A^{-1/2}$,
 $A$ being the adjacency matrix and $D_A$ the degree matrix,}
and  $a_0=\rev{0.6},a_1=0,a_2=0.5$. 
Fig. \ref{Cheb_generation} (ChebNet) and \ref{Local_generation} (L3Net) in appendix \ref{append:exp_result} show that the iGNN model learns to generate samples close in distribution to the ground truth, as reflected in the resemblance of the correlation matrices. 
For this dataset, both spectral and spatial GNN layers have sufficient expressiveness to learn the data distribution.

\section{Discussion}\label{sec:conclude}

In this work, we developed the iGNN model, a conditional generative deep model based on invertible normalizing flow ResNet for inverse graph prediction problems. The model encodes the conditional distribution $X|Y$ into a parametric distribution $H|Y$ by a one-to-one mapping from $X$ to $H$, which allows forward prediction and generation of input data $X$ given an outcome $Y$ (including the construction of uncertainty sets for $X|Y$). The scalability of the iGNN model to large graphs is achieved by taking a factorized formula of $H|Y$ with a provable component separation guarantee, 
and the computational scalability is achieved via adopting GNN layers. 
In addition, the invertibility of the flow network is ensured by a Wasserstein-2 regularization, which can be computed efficiently in the forward pass of the flow network and is compatible with free-form neural network layer types in the residual blocks. 
Theoretically, we analyzed the existence of invertible flows and examined the expressiveness of GNN layers on graph data to express these normalizing flows. 
In experiments, we showed the improved performance of iGNN over alternatives on real data and the scalability of the iGNN model on large graphs.

There are several future directions to extend the work. 
On the theoretical side,
first, under the framework of our problem, more analysis of the $K >1$ case will be useful to go beyond the current Assumption \ref{assump:same-flow}. 
Second, it would be interesting to develop a full approximation result on how the theoretical continuous-time flow identified in Section \ref{sec:theory-flow} can be approximated by a deep residual network. 
Specifically, one can construct a discrete-time flow on $[0, T]$ with $L$ time stamps $t_l$ to approximate the continuous-time one. 
$T=1$ for the Benamou-Brenier flow,
and $T \sim \log ({1}/{\epsilon})$ for the Fokker-Planck flow (so as to achieve $\epsilon$-closeness to the normal density at time $T$). 
The smoothness of the velocity field in both cases, 
cf. Propositions \ref{prop:BB-flow} and \ref{prop:velocity-OU-smooth}, 
implies that at each time stamp $t_l$, $v(x, t_l)$ may be approximated by a shallow residual block with finite many trainable parameters based on universal approximation theory. Compositing the $L$ time steps constructs a neural network approximation of the flow with a provable approximate generation of the data density $p$. The precise analysis is left to future work. 
At last, one may be curious about the connection between the encoding scheme of iGNN model and the Gaussian channel. 
If we view $\mu_k$ as the input of the channel and the output as given by $\mu_k + Z$, for noise $Z\sim \mathcal N(0, \sigma^2 I_d)$,
then the separation of $K$ components can be viewed as a requirement to ensure the error probability of the linear decoder to be sufficiently small. A natural question arises in how many components can be encoded in this way and subsequently decoded (classified) successfully,
for instance, by considering a compact encoding domain. 
To further develop the methodology, one interesting question is to consider the regression problem, where the outcome $Y$ takes continuous (rather than categorical) values. 
Another related question is the inverse of the graph classification problem, that is, given an output label of a whole graph, generate the input graph data including changing graph topology and node/edge features. The current work tackles the node classification setting, and some of our techniques may extend to the graph classification setting. 

\section*{Acknowledgement}

 The work is supported by NSF DMS-2134037.  C.X. and Y.X. are supported by an NSF CAREER Award CCF-1650913, NSF DMS-2134037, CMMI-2015787, DMS-1938106, and DMS-1830210. X.C. is partially supported by NSF, NIH and the Alfred P. Sloan Foundation.

\bibliography{icml2022}
\bibliographystyle{plain}


\appendix
\setcounter{figure}{0} \renewcommand{\thefigure}{A.\arabic{figure}}
\setcounter{table}{0} \renewcommand{\thetable}{A.\arabic{table}}
\setcounter{equation}{0} \renewcommand{\theequation}{A.\arabic{equation}}
\setcounter{remark}{0} \renewcommand{\theremark}{A.\arabic{remark}}

\section{Proofs}\label{sec:proof}

\subsection{Proofs in Section \ref{sec:setup}}

\begin{proof}[Proof of Lemma \ref{lemma:separate-Rd}]
For any $k \in [K]$, and $k ' \neq k$, define  $v_{k,k'}:= \mu_{k'}- \mu_k$, $\| v_{k,k'} \| \ge 3\rho >0$. Define
\begin{equation}\label{eq:def-Skk'}
S_{k,k'} := \{ x \in \R^d,  \, (x- \mu_k)^T  \frac{v_{k, k'}}{\| v_{k, k'} \|}  <  \rho \},
\end{equation}
and we claim that 
\begin{equation}\label{eq:eps-tail-qk}
q_k(  S_{k,k'}^c ) \le \frac{\epsilon}{K}, \quad k \neq k'.
\end{equation}
If true, then let 
\[
\Omega_k:= \underset{ k' \neq k}{\cap} S_{k,k'},
\]
we have that 
\[
q_k(  \Omega_k^c )  = q_k(  \underset{ k' \neq k}{\cup} S_{k,k'}^c )  \le \sum_{k' \neq k} \frac{\epsilon}{K} \le \epsilon,
\]
thus $\Omega_k$ is an $\epsilon$-support of $q_k$ for all $k$. 
Meanwhile, for any $k \neq k'$, $x \in \Omega_k$ and $x' \in \Omega_{k'}$, 
we know that $x \in S_{k,k'}$ and $x' \in S_{k', k}$.
By construction \eqref{eq:def-Skk'}, we know that if we center the origin at $\mu_k$ and rotate the coordinates in $\R^d$ such that $v_{k,k'}$ is along $e_1$ the first coordinate,
then $x^T e_1 < \rho$, and $x'^T e_1 > \| v_{k,k'} \|-\rho \ge 2\rho$. 
This gives that 
\[
\| x - x' \|\ge \| (x-x')^T e_1 \| \ge \rho.
\]
Thus $d(  \Omega_k, \Omega_{k'}) \ge \rho$.

It remains to show \eqref{eq:eps-tail-qk} to finish the proof of the lemma. By construction, this is equivalent to show that for $X \sim \calN(0, I_d)$, 
\[
\Pr[ X^T e_1 \ge  r] \le \frac{\epsilon}{K}, \quad r := \sqrt{2 \log (K/\epsilon)},
\]
where we have that $r \ge \sqrt{2 \log 2} >1$ due to that $\epsilon < 1/2$.
Note that $X^T e_1 \sim \calN(0,1)$, thus 
\begin{align*}
\Pr[ X^T e_1 \ge  r]
& = \int_{r}^\infty \frac{1}{\sqrt{2\pi}} e^{-x^2/2} dx  \\
& \le  \frac{1}{ \sqrt{2\pi}} \int_{r}^\infty e^{-x^2/2} x dx    \quad \text{(by that $x \ge r >1$)} \\
& =  \frac{1}{ \sqrt{2\pi}} e^{-r^2/2} \\
& \le  e^{-r^2/2}  = \frac{\epsilon}{K},
\end{align*}
where the last equality is by the definition of $r$.
\end{proof}

\begin{proof}[Proof of Proposition \ref{prop:separate}]
Let $\epsilon = \varepsilon/(2N) < 1/2$,  then $\rho_N$ here equals $\rho$ in Lemma \ref{lemma:separate-Rd}, where the dimension $d$ in Lemma \ref{lemma:separate-Rd} is $d'$ here. 
Applying Lemma \ref{lemma:separate-Rd} to the $K$-component Gaussian mixture distribution $H_1|Y_1$ in $\R^{d'}$, 
there exist sets $\Omega_k \subset \R^{d'}$ such that $q_k( \Omega_k) \ge 1-\epsilon$, 
and 
\begin{equation}\label{eq:Omegak-Omegak'-sepa}
d_{\R^{d'}}(  \Omega_k, \Omega_{k'}) \ge \rho_N, \quad \forall k \neq k'.
\end{equation}
For each $Y$ from the $K^N$ possible graph labels, define 
\begin{equation}\label{eq:omeag-Y-prod}
\Omega_Y := \prod_{v=1}^N  \Omega_{Y_v}
\end{equation}
which is an $N$-way rectangle in $\R^{d'N}$. 
Due to the factorized form of $p(H|Y)$ as in \eqref{eq:H|Y-factored}, we know that 
\[
q_Y(  \Omega_Y ) = \prod_{v=1}^N q_{Y_v} (\Omega_{Y_v}) \ge (1-\epsilon)^N.
\]
By the elementary relation that 
$1-x \ge e^{-2 x}$ for  $0 < x < 1/2$,  we have that 
\[
(1-\epsilon)^N \ge e^{-2 \epsilon N} \ge 1-2\epsilon N = 1-\varepsilon. 
\]
This shows that $q_Y(  \Omega_Y ) \ge 1-\varepsilon $ and thus $\Omega_Y$ is an $\varepsilon$-support of $q_Y$ in $\R^{d'N}$. 

To prove the proposition, it remains to show that $d(  \Omega_Y, \Omega_{Y'}) \ge \rho_N$  for any $Y \neq Y'$.
Let $ X \in \Omega_Y$ and $X' \in \Omega_{Y'}$, 
by \eqref{eq:omeag-Y-prod}, 
\begin{align*}
X &= [ X_1, \cdots, X_N]^T, \quad X_v \in \Omega_{Y_v} \subset \R^{d'}, \\
X' &= [ X_1', \cdots, X_N']^T, \quad X_v' \in \Omega_{Y_v'}\subset \R^{d'}.
\end{align*}
Because $Y \neq Y'$, there exists a node $u \in [N]$ such that $Y_{u} \neq Y_u'$. Then 
\[
\| X  - X' \|_{\R^{d'N}}
\ge  \| X_u  - X_u' \|_{\R^{d'}}
\ge d_{\R^{d'}}(\Omega_{Y_u} , \Omega_{Y_u'}).
\]
Because $Y_u \neq Y_u'$,  together with \eqref{eq:Omegak-Omegak'-sepa},  we have that $ \| X  - X' \|_{\R^{d'N}} \ge \rho_N$.
This proves that $d(  \Omega_Y, \Omega_{Y'}) \ge \rho_N$.
\end{proof}

\subsection{Proofs in Section \ref{sec:theory-flow}}

\begin{proof}[Proof of Proposition \ref{prop:W2-reg}]
By definition, for each $l$, 
\[
W_2( \rho_{{l-1}}, \rho_{{l}})^2
= \inf_{ \tilde{T}, \, s.t., \tilde{T}_{\#} \rho_{l-1} = \rho_l }  \mathbb{E}_{x \sim \rho_{l-1} } \|  \tilde{T}(x) -x \|^2.
\]
The minimization of \eqref{eq:L+T} can be written as 
\begin{equation}\label{eq:L+T-both-Tl}
\min_{T_1, \cdots, T_l}  \mathcal{L}[ \rho_L ] + \frac{1}{\Delta t} \sum_{l=1}^L 
	 \inf_{ \tilde{T}_l, \, s.t., (\tilde{T}_l)_{\#} \rho_{l-1} = \rho_l }  \mathbb{E}_{x \sim \rho_{l-1} } \|  \tilde{T}_l(x) -x \|^2,  
\end{equation}
and the minimization in \eqref{eq:L+T-reduce} is with the extra constraint that $\tilde{T}_l = {T}_l$. 
It suffices to show that the minimization over ${T}_l$'s only (by requiring $\tilde{T}_l = {T}_l$ and eliminating the variables $\tilde{T}_l$'s) achieves the same minimum of minimizing over both ${T}_l$'s and $\tilde{T}_l$'s in \eqref{eq:L+T-both-Tl}. 
Suppose \eqref{eq:L+T-both-Tl} is minimized at ${T}_l$'s and $\tilde{T}_l$'s, and the minimum equals
\begin{equation}\label{eq:L+T-both-Tl-min}
\mathcal{L} [ \rho_L ] +  \frac{1}{\Delta t} \sum_{l=1}^L  \mathbb{E}_{x \sim \rho_{l-1} } \| x - \tilde{T}_l(x) \|^2,
\end{equation}
where $\rho_l = (T_l \circ \cdots \circ T_1)_{\#} p$, 
and we also have $(\tilde{T}_l)_{\#} \rho_{l-1} = \rho_l $ for $l=1,\cdots, L$.
This guarantees that 
\[
\rho_l = ( \tilde{T}_l \circ \cdots \circ \tilde{T}_1)_{\#} p, \quad l=1, \cdots, L,
\]
that is, if we replace $T_l$ with $\tilde{T}_l$ in \eqref{eq:L+T-both-Tl-min} the value of the equation remains the same. 
This shows that the minimum of  \eqref{eq:L+T-both-Tl} can be achieved by the objective of \eqref{eq:L+T-reduce} at a set of $L$ transport maps $\tilde{T}_l$'s. 
\end{proof}

\begin{proof}[Proof of Proposition \ref{prop:BB-flow}]
The Wasserstein-2 optimal transport $T = \nabla \varphi$, 
and because both $p$ and standard normal $q $ are smooth and have all finite moments in $\R^d$,
$\varphi $ is also smooth \cite{caffarelli1996boundary}.
The optimal $v(x,t)$ has the expression that 
\begin{equation}\label{eq:v-BB-explicit}
 v( \cdot ,t) = (T- {\rm Id}) \circ T_t^{-1},
 \end{equation}
where $T_t = (1-t) {\rm Id} + t T$  is the  displacement interpolation for $0 \le t \le 1$ \cite{villani2021topics}.  
Because both $T$ and $T_t$ are smooth, 
 $v(x,t)$ is smooth on $\R^d \times [0,1]$.
For any bounded domain $\Omega \subset \R^d$, $v(x,t)$ has finite $x$-Lipschitz constant on $\Omega \times [0,1]$,
which suffices for the well-posedness of the IVP \citep{Sideris2013OrdinaryDE}.
\end{proof}
\begin{proof}[Proof of Proposition \ref{prop:velocity-OU-smooth}]
The solution $\rho(x,t)$  of \eqref{eq:fokker-planck} has the explicit expression as 
\begin{equation}\label{eq:rhoxt-OU}
\rho(x,t) = \int K_t( x,y) \rho_0(y) dy,
\quad
K_t(x,y)  := \frac{1}{(2\pi \sigma_t^2)^{d/2}} e^{- \frac{| x - e^{-t} y |^2}{ 2 \sigma_t^2}}, 
\quad \sigma_t^2:= 1-e^{-2t}.
\end{equation}
By the smoothness of $K_t(x,y)$,  $\rho(x,t)$ is smooth over $\R^d \times (0,\infty)$.  
By \eqref{eq:f-jko-flow}, for any $t > 0$,
$ v (x,t) =  - x-  {\nabla \rho(x,t)}/{\rho(x,t)}$,
and thus $v (x,t)$ is also smooth over $\R^d \times (0,\infty)$.  
Same as in the proof of Proposition \ref{prop:BB-flow},
the global smoothness of $v(x,t)$ suffices for the well-posedness of the IVP on $\R^d \times (0,T)$ for any $T>0$,
\end{proof}

\subsection{Proofs in Section \ref{sec:expressiveness}}

\begin{proof}[Proof of Lemma \ref{lem:force}]
By the expression of $\rho(x,t)$ of the OU process \eqref{eq:rhoxt-OU}, we know that 
\[
x(t) | x(0) \sim \calN (  e^{-t} x(0), \sigma_t^2 I ), \quad \sigma_t^2 =1- e^{-2t}.
\]
Because $x(0) \sim \calN(0,\Sigma)$,
one can directly verify that the marginal distribution of $x(t)$ is a gaussian with mean $0$ and variance $\Sigma_t$  defined as 
\begin{equation}\label{sigmat}
    \Sigmat :=(1- e^{-2t}) I+ e^{-2t} \Sigma.
\end{equation}
This gives that 
\[
\rho(x,t) \propto e^{- \frac{1}{2}x^T \Sigma_t^{-1} x},
\]
and as a result, 
\[
\nabla \log \rho(x,t) = - \Sigma_t^{-1} x. 
\]
This proves \eqref{force_general_form} by definition \eqref{eq:f-jko-flow}. 
\end{proof}
\begin{proof}[Proof of Lemma \ref{lem:force-BB}]
The Wasserstein-2 optimal transport between two Gaussians has closed form solution \cite{takatsu2011wasserstein}.
From source  density $p = \calN(0,\Sigma)$ to target density $q = \calN(0,I)$,
the optimal transport map
\[
T = \Sigma^{-1/2},
\]
and then $T_t = (1-t) I + t \Sigma^{-1/2} $. 
By \eqref{eq:v-BB-explicit}, 
\begin{equation}
 v(x ,t) = (T- {\rm Id}) \circ T_t^{-1}x
  =  - (I - \Sigma^{-1/2}) ((1-t) I +  t\Sigma^{-1/2})^{-1}  x,
 \end{equation}
 which is equivalent to \eqref{eq:v-BB-lemma}. 
\end{proof}
\begin{proof}[Proof of Lemma \ref{lemma:fs-gs-approx}]
To prove (i), note that 
\begin{align}
g_s (x) 
& = \frac{ x }{  (1-s) + s  x }  \nonumber \\
& = \frac{x}{ (1-s) + s b } \left(  1 + \frac{  s (x-b) }{  (1-s) + s b}  \right)^{-1}  \nonumber  \\
& = \frac{x}{ (1-s) + s b } \sum_{k=0}^\infty \left( - \frac{  s (x-b) }{  (1-s) + s b}  \right)^k. 
\label{eq:gs-expand-1}
\end{align}
Because $b$ is the midpoint of the interval $[ 1/a, a ]$ where $x$ lies on, we have that for all $s \le 1/2$,
\begin{equation}\label{eq:bound-rho-gs}
\left|  \frac{ s (x-b) }{  (1-s) + s b} \right| 
 \le \frac{  s b}{  (1-s) + s b}
 = \frac{  b}{   b +  {( 1-s )}/{s } } 
 \le  \frac{  b}{   b +  1},  
\end{equation}
where in the last inequality we have use that $s \le 1/2$. 
We now truncate to the expansion \eqref{eq:gs-expand-1} up to $k =n$, and define
\[
Q_s^{(n+1)}(x):= \frac{x}{ (1-s) + s b } \sum_{k=0}^n \left( - \frac{  s (x-b) }{  (1-s) + s b}  \right)^k,
\]
the polynomial $Q_s^{(n+1)}$ is a polynomial of degree $(n+1)$ for $ s >0$ and reduces to $x$ when $s =0$. 
By \eqref{eq:bound-rho-gs}  and that the number $(1-s) + s b \ge 1$ because $b \ge 1$, 
we have that for all $x \in [ 1/a, a ]$,
\begin{align*}
| g_s(x)  - Q_s^{(n+1)}(x) | 
&\le  \frac{|x|}{ (1-s) + s b } \sum_{k>n} \left| \frac{  s (x-b) }{  (1-s) + s b}  \right|^k   \\
&\le  a  \sum_{k>n}  \left(\frac{  b}{   b +  1}\right)^k  \\
&= a  b \left(\frac{  b}{   b +  1}\right)^{n},
\end{align*}
which holds uniformly for all $s \in [0, 1/2]$.

To prove (ii), 
use the expansion of $f_s(x)$ as
\begin{align}
f_s(x) 
& = \frac{ 1 }{    s + (1-s) x   }  \nonumber \\
& = \frac{1}{ s + (1-s) b } \sum_{k=0}^\infty \left( - \frac{   (1-s)  (x-b) }{   s + (1-s) b}  \right)^k, 
\label{eq:fs-expand-1}
\end{align}
and define $P_s^{(n)}$ as the truncated summation up to $k = n$, which is a polynomial of degree $n$. 
Similarly, we can bound the residual as
\begin{align}
| f_s(x)  - P_s^{(n)}(x) | 
&\le  \frac{1}{   s + (1-s) b } \sum_{k>n} \left| \frac{   (1-s) (x-b) }{   s + (1-s) b}  \right|^k  \nonumber \\
&\le   \sum_{k>n}  \left(\frac{  b}{   b +  1}\right)^k   
=  b \left(\frac{  b}{   b +  1}\right)^{n},
\label{eq:bound-fs-1}
\end{align}
where we use that $\forall x \in  [ 1/ a, a ]$ and $s \ge 1/2$,
\begin{equation}
\left|  \frac{  (1-s) (x-b) }{   s + (1-s) b} \right| 
 \le \frac{   (1-s) b}{   s +  (1-s) b}
 = \frac{  b}{   b +  { s }/ { (1-s )} } 
 \le  \frac{  b}{   b +  1}.   
\end{equation}
The bound \eqref{eq:bound-fs-1} holds uniformly for all $x \in  [ 1/a, s]$ and $s \ge 1/2$.
\end{proof}
\begin{proof}[Proof of Theorem \ref{thm:spectral_sigma}]
Under the assumption of the theorem, Lemma \ref{lemma:fs-gs-approx} applies with $s = 1-e^{-2t}$. 

\vspace{5pt}
Proof of (i):
We set $q_t (x) = Q_s^{(n+1)} ( q  (x))$, which is a polynomial of degree at most $n_0+n+1$ since the degree of $q$ is at most $n_0$.
Recall that $\tilde{S} = q  (\tilde{L})$ satisfies (C1)(C2), and by \eqref{eq:Tt-by-fs-gs}, 
\begin{align*}
{ T}_t'  -   q_t( \tilde{L}) 
&= g_s(S) - Q_s^{(n+1)} ( \tilde{S} ) \\
&= ( g_s(S) - g_s(\tilde{S}) ) + ( g_s(\tilde{S})  -   Q_s^{(n+1)} ( \tilde{S} )) =: \textcircled{1} + \textcircled{2}.
\end{align*}
We bound $\| \textcircled{1}\|_{op} $ and $\| \textcircled{2}\|_{op} $ respectively. 

By definition of $g_s$ as in \eqref{eq:def-fs-gs}, for any invertible real symmetric matrix $A$, 
denote $\alpha : =  s $, $\beta := 1-s$,
\begin{align*}
g_s(A) 
& = \left( \alpha I  + \beta A^{-1} \right)^{-1} \\
& = A ( \beta I + \alpha A)^{-1}
= ( \beta I + \alpha A)^{-1} A.
\end{align*}
By that 
$\textcircled{1}  =  g_s(S)  - g_s( \tilde{S})  
= ( \beta I + \alpha S)^{-1} S - \tilde{S}  ( \beta I + \alpha \tilde{S})^{-1}$,
one can verify that 
\begin{align*}
( \beta I + \alpha S)  \textcircled{1} ( \beta I + \alpha \tilde{S}) 
&  =  S  ( \beta I + \alpha \tilde{S})- ( \beta I + \alpha S) \tilde{S}    
= \beta( S - \tilde{S}),  
\end{align*}
and thus
\begin{align*}
\|  \textcircled{1}   \|_{op}
& = \| ( \beta I + \alpha S)^{-1 }  \beta( S - \tilde{S}) ( \beta I + \alpha \tilde{S})^{-1} \|_{op} \\
& \le  \beta \| S - \tilde{S}\|_{op} \| ( \beta I + \alpha S)^{-1 } \|_{op} \| ( \beta I + \alpha \tilde{S})^{-1} \|_{op}.
\end{align*}
By (C2), $\| \tilde{S} - S \|_{op} \le \delta \sqrt{\kappa}$;
By \eqref{eq:spec-Sig-S}, $ \min {\rm spec}(S)  \ge 1/\sqrt{\kappa} \ge 0$, and then
\[
\| ( \beta I + \alpha S)^{-1 } \|_{op}  
\le \frac{1}{ \min \{  {\rm spec} ( \beta I + \alpha S) \}}
\le \frac{1}{ \beta};
\]
Meanwhile, $ \min {\rm spec}(\tilde{S})  \ge 1/\sqrt{\kappa} \ge 0$ by (C1), and then similarly,  
$ \| ( \beta I + \alpha \tilde{S})^{-1 } \|_{op}  \le 1/\beta$.
Putting together, 
\begin{equation}\label{eq:bound-circle1-(i)}
\|  \textcircled{1}   \|_{op}
\le  \beta \delta \sqrt{\kappa}  (\frac{1}{\beta})^2
= \frac{1}{\beta}  \delta \sqrt{\kappa} 
\le 2  \delta \sqrt{\kappa},
\end{equation}
where the last inequality is by that 
$\beta = e^{-2t} \ge e^{-2t_0} = 1/2$.

To bound $\| \textcircled{2}\|_{op} $, 
we introduce the following lemma which can be verified directly by definition (in below).

\begin{lemma}\label{lemma:matrix-bounded-by-func-approx}
Let $A$ be a real symmetric matrix and ${\rm spec}(A)\subset [a,b]$. For real-valued functions $f$ and $g$ on $[a,b]$,
\[
\| f(A) - g(A)\|_{op} \le \sup_{x \in [a,b]} | f(x) - g(x) |.
\]
\end{lemma}

By definition of $\textcircled{2}$ and that $  {\rm spec}( \tilde{S} ) \subset [ 1/\sqrt{\kappa}, \sqrt{\kappa} ]$ by (C1), 
\begin{align}
\| \textcircled{2}\|_{op} 
& = \| g_s(\tilde{S})  -   Q_s^{(n+1)} ( \tilde{S} ) \|_{op} \nonumber \\
& \le \sup_{x \in [ 1/\sqrt{\kappa}, \sqrt{\kappa} ]} | g_s(x) -  Q_s^{(n+1)}(x) |   
	\quad \text{ (by Lemma \ref{lemma:matrix-bounded-by-func-approx} )}  \nonumber \\
& \le \sqrt{\kappa} b  \left(\frac{b}{b+1} \right)^n, 
	\quad \text{(by Lemma \ref{lemma:fs-gs-approx}(i))} \label{eq:bound-circle2-(i)-1}
\end{align}
where $a = \sqrt{\kappa}$ and $b = ( 1/a + a )/2$.
Because $b \ge 1$, by the elementary inequality
$\frac{b}{b+1}  = 1 - \frac{1}{b+1} \le e^{-1/(b+1)}$, 
we have
\begin{equation}\label{eq:bound-factor-by-exp}
\left(\frac{b}{b+1} \right)^n 
\le e^{-n/(b+1)}
\le e^{-n/(\sqrt{\kappa}+1)}
\end{equation}
by that $b \le \sqrt{\kappa}$. 
Back to \eqref{eq:bound-circle2-(i)-1}, we have
\begin{equation}\label{eq:bound-circle2-(i)-2}
\| \textcircled{2}\|_{op} 
 \le  \sqrt{\kappa} b  e^{-n/(\sqrt{\kappa}+1)} 
 \le  \kappa  e^{-n/(\sqrt{\kappa}+1)}.
\end{equation}
Putting together \eqref{eq:bound-circle1-(i)} and \eqref{eq:bound-circle2-(i)-2} proves \eqref{eq:bound-short-time} by triangle inequality
for all $t \le t_0$.

\vspace{5pt}
Proof of (ii):
We set $p_s (x) = P_s^{(n)} ( p  (x))$, which is a polynomial of degree at most $n_0+n$.
Recall that $\tilde{\Sigma} = p  (\tilde{L})$ satisfies (C1)(C2), and then
\begin{align*}
T_t' -   p_t( \tilde{L}) 
&= f_s(\Sigma) - P_s^{(n)} ( \tilde{\Sigma} ) \\
&= ( f_s(\Sigma) - f_s(\tilde{\Sigma}) ) + ( f_s(\tilde{\Sigma})  -   P_s^{(n)} ( \tilde{\Sigma} )) =: \textcircled{1} + \textcircled{2}.
\end{align*}

To bound $\| \textcircled{1}\|_{op}$, 
by definition of $f_t$ as in \eqref{eq:def-fs-gs}, the constants $\alpha$ and $\beta$ as before, we have
\[
f_s(A)  = \left( \alpha I  + \beta A \right)^{-1},
\]
and then 
\begin{align*}
 \textcircled{1} 
&  =   \left( \alpha I  + \beta \Sigma \right)^{-1} - \left( \alpha I  + \beta \tilde{\Sigma} \right)^{-1} 
 =\left( \alpha I  + \beta \Sigma \right)^{-1} \beta (  \tilde{\Sigma} -  {\Sigma} )\left( \alpha I  + \beta \tilde{\Sigma} \right)^{-1}.
\end{align*}
By that $ \min {\rm spec}(\Sigma),  \min {\rm spec}(\tilde{\Sigma})   \ge 1/\sqrt{\kappa} \ge 0$,
and also with that $\alpha = 1- e^{-2t} \ge 1- e^{-2t_0} = 1/2$ for $t \ge t_0$,
\[
\| \left( \alpha I  + \beta \Sigma \right)^{-1} \|_{op} 
\le \frac{1}{ \min \{  {\rm spec} ( \alpha I + \beta \Sigma) \}}
\le \frac{1}{ \alpha} 
\le 2,
\]
and same with $\| \left( \alpha I  + \beta \tilde{\Sigma} \right)^{-1} \|_{op} $.
Together with $\| \tilde{\Sigma} - \Sigma \|_{op} \le \delta \sqrt{\kappa}$ By (C2), we have
\begin{equation}\label{eq:bound-circle1-(ii)}
\|  \textcircled{1}   \|_{op}
\le  \beta \delta \sqrt{\kappa}  \cdot 2^2
\le 2  \delta \sqrt{\kappa},
\end{equation}
where we have used that $\beta = e^{-2t} \le e^{-2t_0} = 1/2$ when $t  \ge t_0$. 

To bound $\| \textcircled{2}\|_{op} $, by that $  {\rm spec}( \tilde{\Sigma} ) \subset [ 1/\sqrt{\kappa}, \sqrt{\kappa} ]$ by (C1), 
\begin{align}
\| \textcircled{2}\|_{op} 
& = \| f_s(\tilde{S})  -   P_s^{(n)} ( \tilde{\Sigma} ) \|_{op} \nonumber \\
& \le \sup_{x \in [ 1/\sqrt{\kappa}, \sqrt{\kappa} ]} | f_s(x) -  P_s^{(n)}(x) |   
	\quad \text{ (by Lemma \ref{lemma:matrix-bounded-by-func-approx} )}  \nonumber \\
& \le b  \left(\frac{b}{b+1} \right)^n
	\quad \text{(by Lemma \ref{lemma:fs-gs-approx}(ii))}  \nonumber \\
& \le  \sqrt{\kappa}  e^{-n/(\sqrt{\kappa}+1)}.
	\quad \text{(by \eqref{eq:bound-factor-by-exp} and that $b \le a=\sqrt{\kappa}$)}
\label{eq:bound-circle2-(ii)}
\end{align}
Putting together \eqref{eq:bound-circle1-(ii)} and \eqref{eq:bound-circle2-(ii)} proves \eqref{eq:bound-long-time} for all $t \ge t_0$.
\end{proof}

\begin{proof}[Proof of Lemma \ref{lemma:matrix-bounded-by-func-approx}]
Suppose $A$ is $N$-by-$N$.
Let $A = U \Lambda U^T$ be the eigen-decomposition, where $U$ is an orthogonal matrix consisting of columns $U_i$, $i=1,\cdots, N$,
and $\lambda_i$ are the diagonal entries of $\Lambda$ (the associated eigenvalues). 
Then
\[
f(A) - g(A) = \sum_{i=1}^N ( f(\lambda_i) - g(\lambda_i) )U_i U_i^T,
\]
and thus
\[
\| f(A) - g(A) \|_{op}  = \max_{ 1 \le i \le N} | f(\lambda_i) - g(\lambda_i) | \le \sup_{x \in [a,b]} | f(x) - g(x) |
\]
due to that all $\lambda_i$ lie inside $[a,b]$.
\end{proof}
\begin{proof}[Proof of Theorem \ref{thm:local_sigma}]
To prove (i), let
\[
B_k = \tilde{S}^k, \quad k=0,\cdots, n+1, 
\]
where $\tilde{S}$ is $v$-local and satisfies (C1)(C2). As a result, $B_k$ is ($kv$)-local.
We set the coefficients $c_k(t)$ such that
\[
\sum_{k=0}^{n+1} c_k(t) B_k 
= \sum_{k=0}^{n+1} c_k(t) \tilde{S}^k
=  Q_s^{(n+1)} ( \tilde{S} ), \quad s:= 1-e^{-2t}.
\]
The proof of Theorem \ref{thm:spectral_sigma}Theorem \ref{thm:spectral_sigma}(i) that bounds $\| g_s(S) - Q_s^{(n+1)} ( \tilde{S} )\|_{op}$ 
by the r.h.s. of \eqref{eq:bound-short-time} only uses the properties (C1)(C2) of $ \tilde{S}$, 
and thus also applies there. 

Statement (ii) can be proved similarly: let $B_k = \tilde{\Sigma}^k$, $k=0,\cdots, n$, $B_k$ is ($kv$)-local,
and set the coefficients $c_k(t)$ such that
\[
\sum_{k=0}^{n} c_k(t) B_k 
= \sum_{k=0}^{n} c_k(t) \tilde{\Sigma}^k
=  P_s^{(n)} ( \tilde{\Sigma} ).
\]
Since $\tilde{\Sigma}$ satisfies (C1)(C2), 
the rest of proof is the same as in Theorem \ref{thm:spectral_sigma}(ii) 
which  bounds $\| f_s(\Sigma) - P_s^{(n)} ( \tilde{\Sigma} )\|_{op}$ by the r.h.s. of \eqref{eq:bound-long-time}.
\end{proof}

\section{Additional experimental details}\label{sec:exp_append}

\begin{figure}[!t]
\centering
\includegraphics[width=0.65\linewidth]{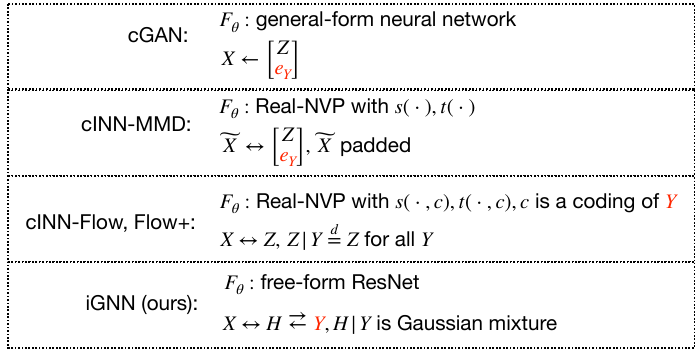}
    
    \caption{
     Detailed comparison of conditional generative networks in cGAN (top), cINN (middle two), and iGNN (bottom). 
    The random code $Z \sim \calN(0,I_d)$ has the same dimension as input data $X\in \R^d$. 
    First, cGAN \citep{cGAN} uses an general-form generative NN,
    where the input concatenates $Z$ and one-hot-encoded layer $Y$. Second, \ciNNI \ \citep{cinn1} constructs an invertible network between the padded input $\widetilde{X}$ and the concatenated input of $Z$ and $e_Y$, and uses Real-NVP layers to ensure network invertibility. 
    Third, \ciNNII\ \citep{cinn2} and \ciNNIIplus\ \citep{cinn2+} also use Real-NVP layers to construct an invertible network between $X$ and $Z$, where each layer contains encoded information of the label $Y$. 
    Lastly, the proposed iGNN model builds an invertible network between the intermediate feature $H$ and input data $X$, where $H|Y$ denotes a Gaussian mixture in $\R^d$, and constructs the invertible network using free-form residual blocks.
  }
    \label{fig:comparison_appendix}
\end{figure}

\subsection{Experimental set-up}\label{sec:setup_details}

\subsubsection{Computation of $\log\det$}

To compute the log determinant in \eqref{Nflow_adaptive}, we adopt the following unbiased log determinant approximation technique as proposed in \citep{ResFlow}. Let $F_{\theta}(x)=x+f_{\theta}(x)$ denote the output from a generic ResNet block with parameter $\theta$. First, observe that for any input $x$, $\log|\det J_{F_{\theta}}(x)|=\text{tr}(\log J_{F_{\theta}}(x))$ because the matrix $J_{F_{\theta}}(x)$ is non-singular. We thus have $\text{tr}(\log J_{F_{\theta}}(x))=\text{tr}(\log (I+J_{f_{\theta}}(x)))$. As a result, the trace of the matrix logarithm can be expressed as
\begin{equation}\label{matrix_log}
    \text{tr}(\log (I+J_{f_{\theta}}(x))) = \text{tr}\left(\sum_{k=1}^{\infty}  \frac{(-1)^{k+1}}{k}[J_{f_{\theta}}(x)]^k\right).
\end{equation}
Based on \eqref{matrix_log}, which takes infinite time to compute, we can obtain an unbiased estimator in finite time based on the ``Russian roulette`` estimator approach \citep{russian_roulette,ResFlow}. For a ResNet as a concatenation of $L$ ResNet blocks, the approximation is applied to each block and summed over all blocks.
Lastly, to speed up gradient computation of the approximation, we further adopt memory efficient backpropagation through the early computation of gradients \citep{ResFlow}.

\begin{figure}[!t]
    \begin{minipage}[b]{0.48\linewidth}
        \includegraphics[width=\linewidth]{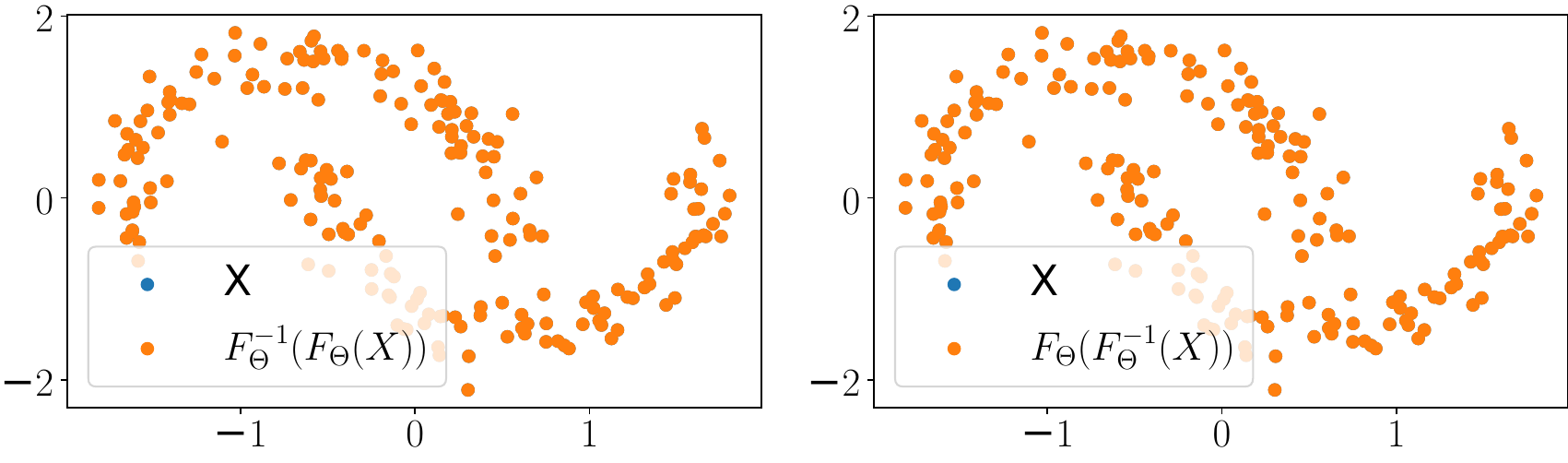}
        \subcaption{In-distribution sample $X$}
    \end{minipage}
    \hspace{0.25in}
    \begin{minipage}[b]{0.48\linewidth}
        \includegraphics[width=\linewidth]{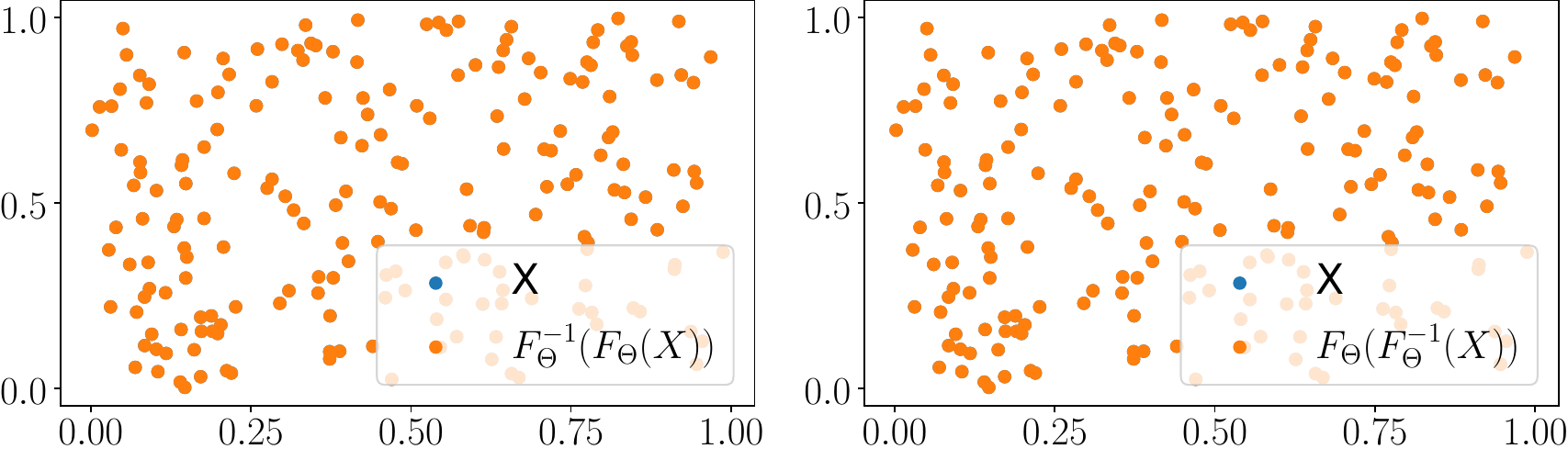}
        \subcaption{Out-of-distribution sample $X$}
    \end{minipage}
    \caption{The invertibility of iGNN on non-graph data in $\R^2$. We visualize $F^{-1}_{\theta}(F_{\theta}(X))$ (forward then invert) and $F_{\theta}(F^{-1}_{\theta}(X))$ (invert then forward) on in-distribution data (i.e., $X$ as a part of two-moon data) and out-of-distribution data (i.e., $X$ having random $U[0,1)$ entries). \rev{Note that the scatter plots of $X$ and $F_{\theta}(F^{-1}_{\theta}(X))$ completely overlap because of small invertibility errors.}}
    \label{check_inv}
\end{figure}
\subsubsection{Model evaluation metrics}
\begin{figure}[!b]
\centering
    \includegraphics[width=0.5\linewidth]{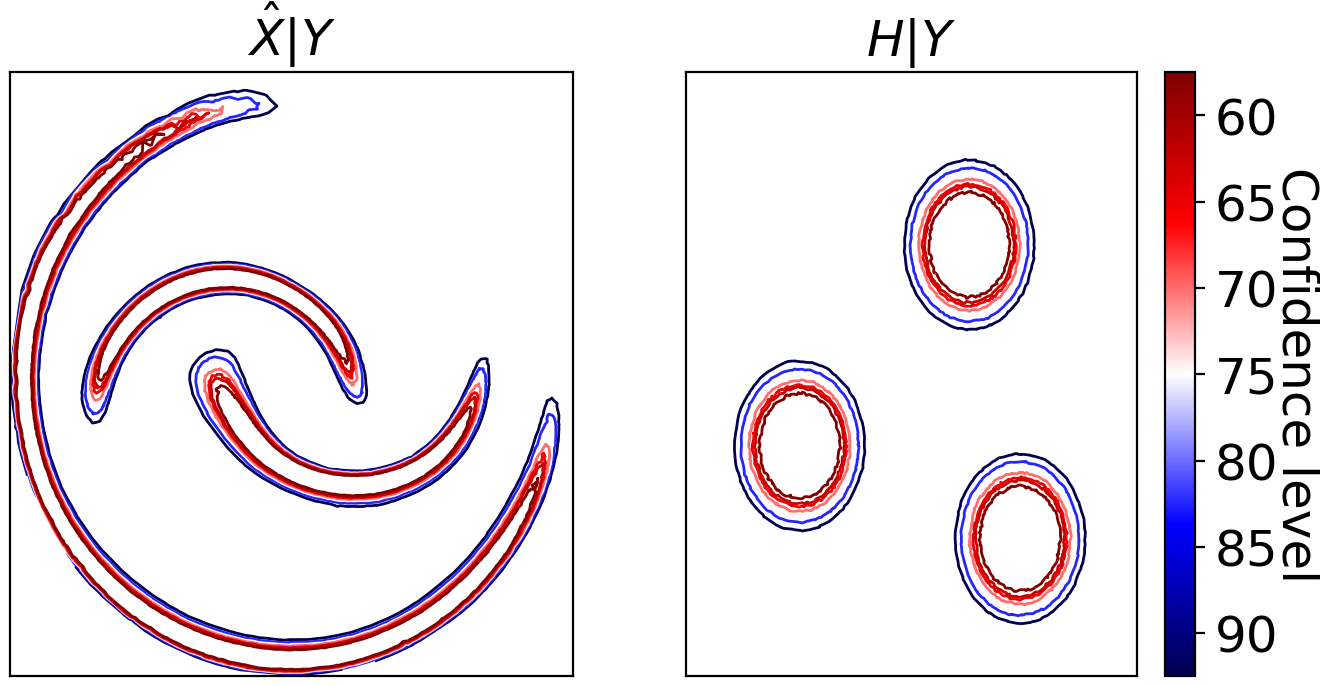}
\caption{
Confidence region of three moons by \rev{iGNN}. The setup is identical to that in Fig. \ref{three_moon}, where we visualize the confidence region of $X|Y$ based on that of $H|Y$. The confidence region in the input space of $X$ can be computed from that in the feature space $H$ based on the parametric mixture model of $H$. 
}
\label{three_moon-confidence}
\end{figure}
\vspace{0.1in}
\noindent \textit{MMD statistics metric.} 
Given two sets of samples $\boldsymbol{X}=\{x_1,\ldots,x_n\},\boldsymbol{X'}=\{x'_1,\ldots,x'_n\}$ of same sample size $n$, 
the MMD two-sample statistic between $\boldsymbol{X}$ and $\boldsymbol{X'}$ is defined as 
\begin{equation}\label{MMD}
    \text{MMD}(\boldsymbol{X},\boldsymbol{X'}):=\frac{1}{n^2}\sum_{i=1}^n \sum_{j=1}^n k(x_i,x_j) + \frac{1}{n^2}\sum_{i=1}^n \sum_{j=1}^n k(x'_i,x'_j)
    - \frac{2}{n^2}\sum_{i=1}^n \sum_{j=1}^n k(x_i,x'_j),
\end{equation}
where we use the radial basis kernel $k(x,x')=\exp(-\alpha\|x-x'\|^2)$ with $\alpha=0.1$.

For the $K$-class conditional distribution $\boldsymbol{X}|Y$, where we denote by  $\{ \boldsymbol{X}|Y=k \}$ the set of samples $\{x_i:y_i=k\}_{i=1}^n$,  the overall MMD statistic is defined using \eqref{MMD} as 
\begin{equation}\label{weighted_MMD}
    {\rm MMD} = \sum_{k=1}^K w_k  
    \text{MMD}(  \{ \boldsymbol{X}|Y=k \} ,   
        \{ \boldsymbol{X'}|Y=k \}),
\quad w_k = \frac{\sum_{i=1}^n \textbf{1}(y_i=k)}{n}.
\end{equation}
Note that on graph data where $Y$ concatenates all nodal labels, 
the summation is over all types of $Y$ (up to $K^N$ many). 

\begin{figure}[!t]
\centering
\begin{minipage}[b]{0.5\linewidth}
    \includegraphics[width=\linewidth]{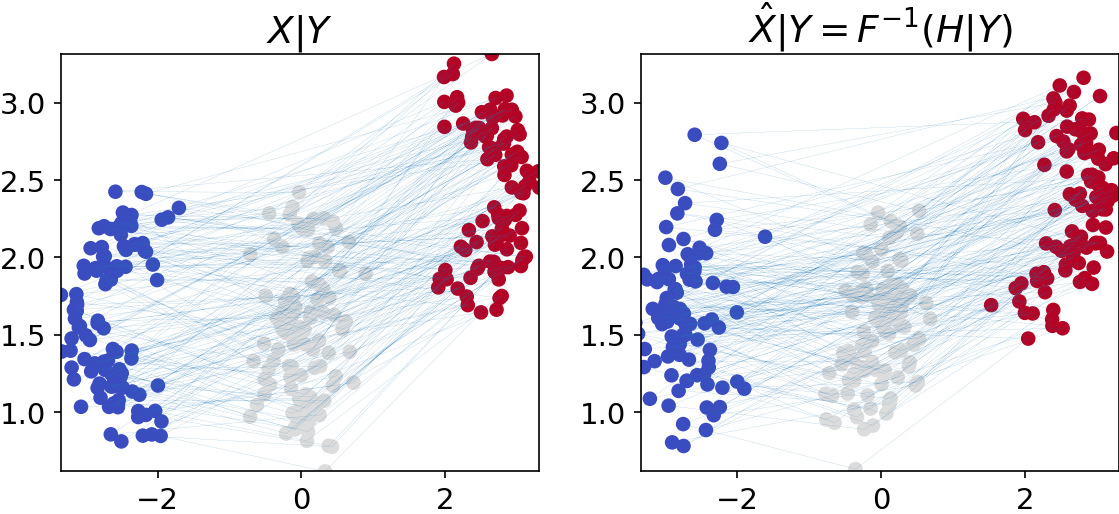}
    \includegraphics[width=\linewidth]{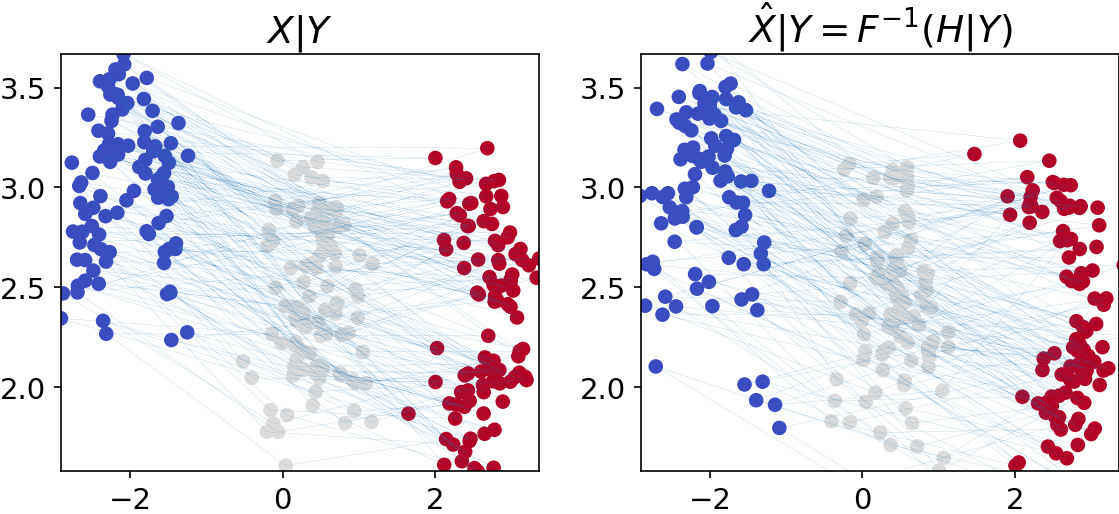}
    \vspace{-0.15in}
    \subcaption{$X|Y$ \hspace{0.7in} iGNN $\hat{X}|Y$}
\end{minipage}
\begin{minipage}[b]{0.2331\linewidth}
    \includegraphics[width=\linewidth]{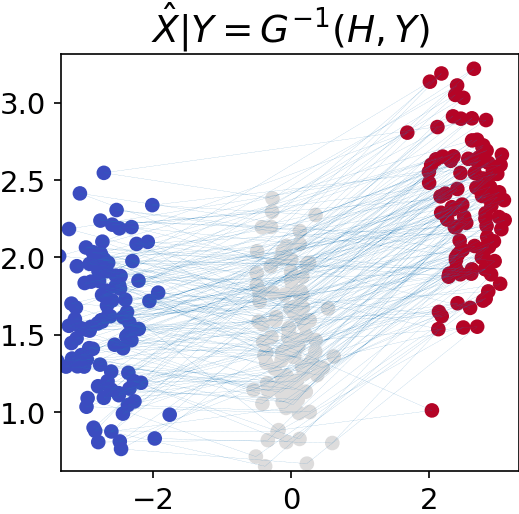}
    \includegraphics[width=\linewidth]{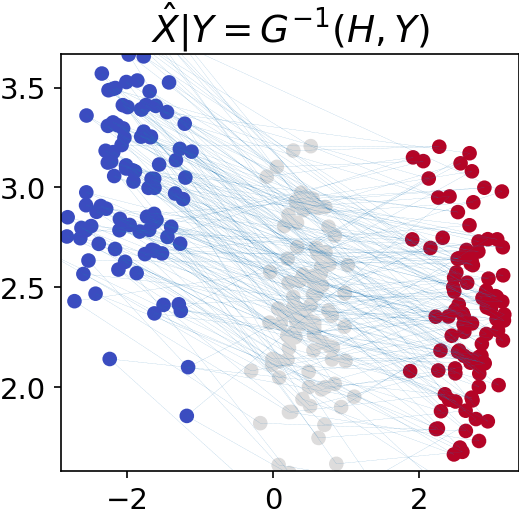}
    \vspace{-0.15in}
    \subcaption{\ciNNI\ $\hat{X}|Y$}
\end{minipage}
\begin{minipage}[b]{0.225\linewidth}
    \includegraphics[width=\linewidth]{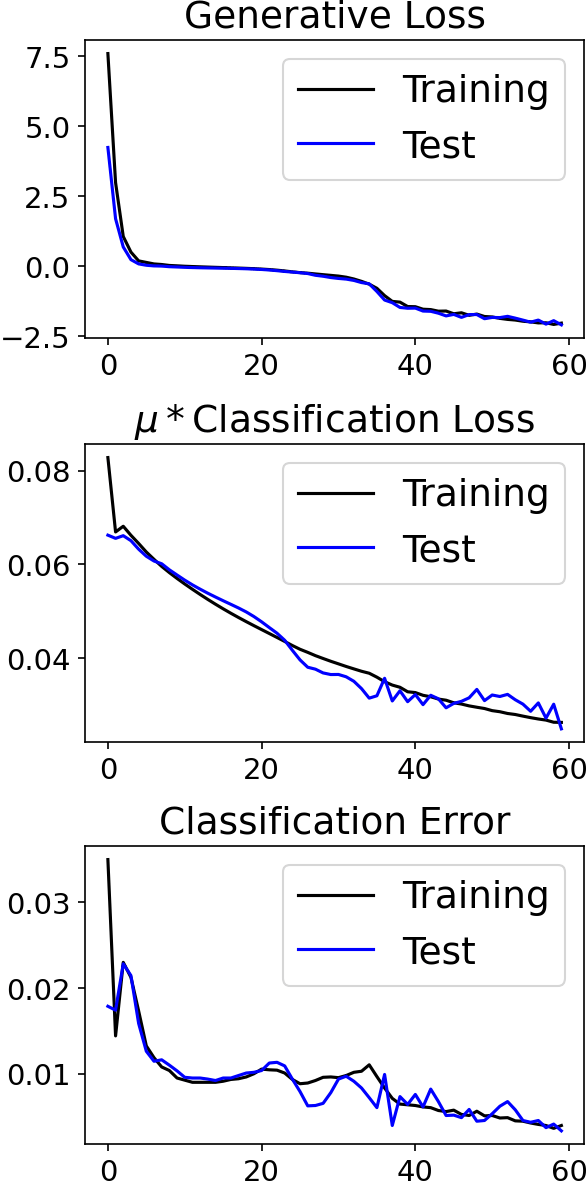}
    \vspace{-0.15in}
    \subcaption{iGNN losses}
\end{minipage}
\caption{Compare iGNN  vs. \ciNNI\ on generating two-dimensional graph node features on the three-node graph. Color indicates node index and rows \rev{in (a) and (b)} are determined by two different values of $Y\in \{0,1\}^3$. We connect the two-dimensional node
features belonging to the same 3-by-2 feature matrix $X$ by light blue lines to illustrate the distribution of $X|Y$.
\rev{Both models use L3net GNN layers.}
}
\label{3node_nonconvex}
\end{figure}

\vspace{0.1in}
\noindent \textit{Energy statistic metric.}  Given two sets of samples $\boldsymbol{X}=\{x_1,\ldots,x_n\},\boldsymbol{X'}=\{x'_1,\ldots,x'_n\}$, The energy statistic under $\ell_2$ norm is defined as
\begin{equation}\label{energy}
    \text{Energy}(\boldsymbol{X},\boldsymbol{X'}):=\frac{2}{n^2}\sum_{i=1}^n \sum_{j=1}^n \|x_i-x'_j\|_2 - \frac{1}{n^2}\sum_{i=1}^n \sum_{j=1}^n \|x_i-x_j\|_2- \frac{1}{n^2}\sum_{i=1}^n \sum_{j=1}^n \|x'_i-x'_j\|_2,
\end{equation}
For $K$-class conditional distribution $\boldsymbol{X}|Y$, the weighted energy statistics is defined using \eqref{energy} as
\begin{equation}\label{weighted_energy}
    {\rm Energy} = \sum_{k=1}^K w_k  \text{Energy}( \{\boldsymbol{X}|Y=k\} ,   \{ \boldsymbol{X'}|Y=k\} ),
\quad w_k = \frac{\sum_{i=1}^n \textbf{1}(y_i=k)}{n}.
\end{equation}
Computation of the weighted statistics on graph data is identical to that of the weighted MMD statistics on graph.

\vspace{0.1in}
\noindent \textit{Model invertibility error.} 
We see from Fig. \ref{check_inv} that iGNN under the Wasserstein-2 regularization ensures model invertibility up to very high accuracy.

\begin{figure}[!t]
\centering
\begin{minipage}[b]{0.28\linewidth}
    \includegraphics[width=\linewidth]{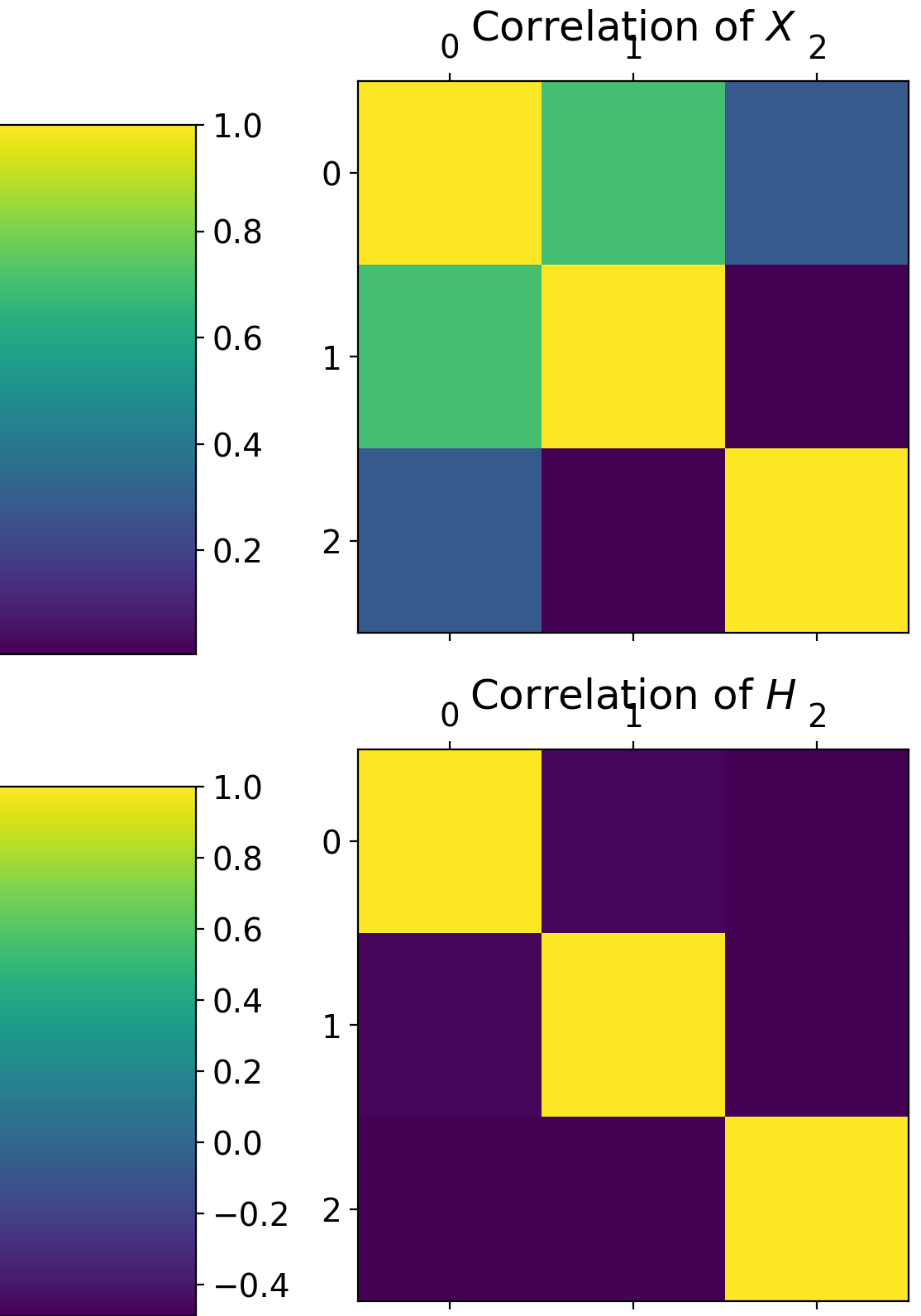}
    \subcaption{
    True data}
\end{minipage}
\begin{minipage}[b]{0.185\linewidth}
    \includegraphics[width=\linewidth]{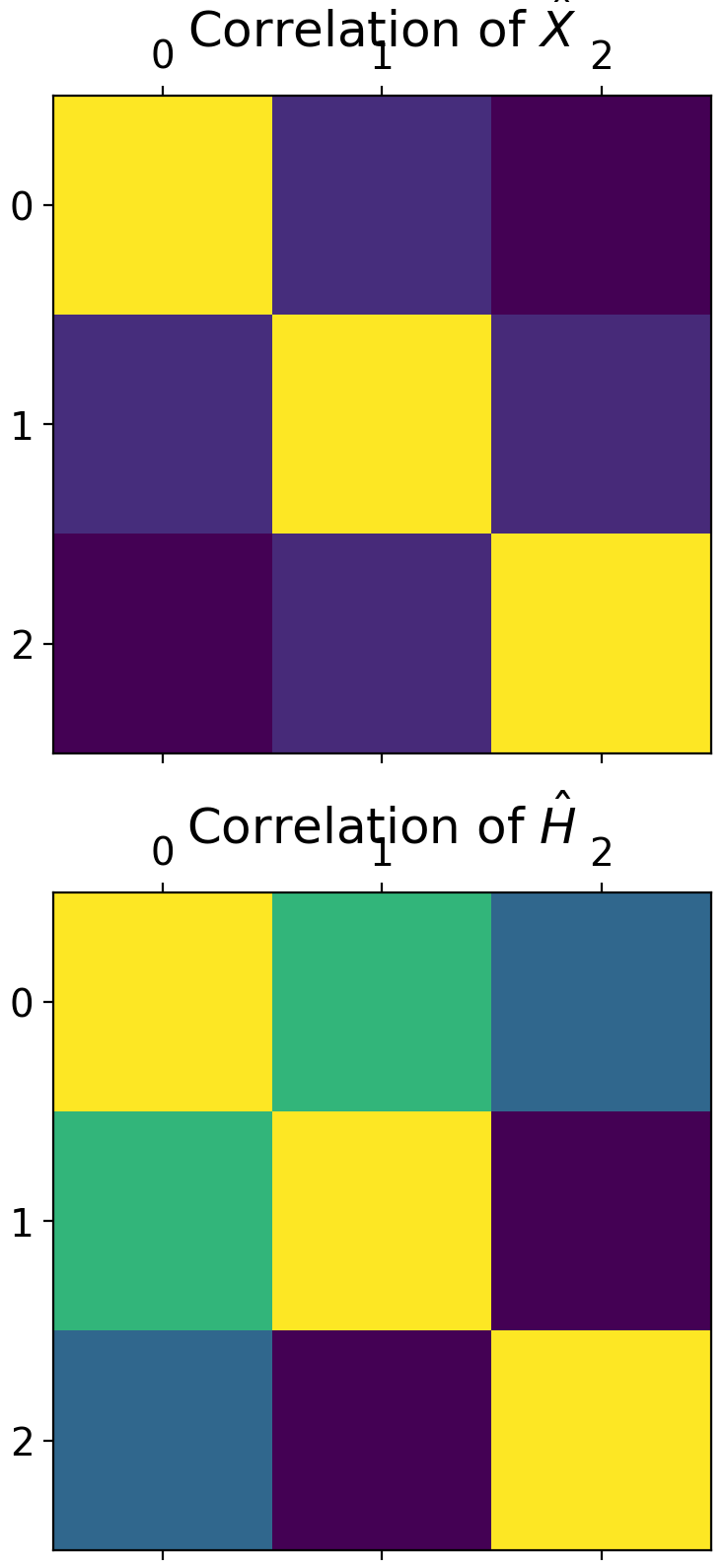}
    \subcaption{ChebNet}
\end{minipage}
\begin{minipage}[b]{0.183\linewidth}
    \includegraphics[width=\linewidth]{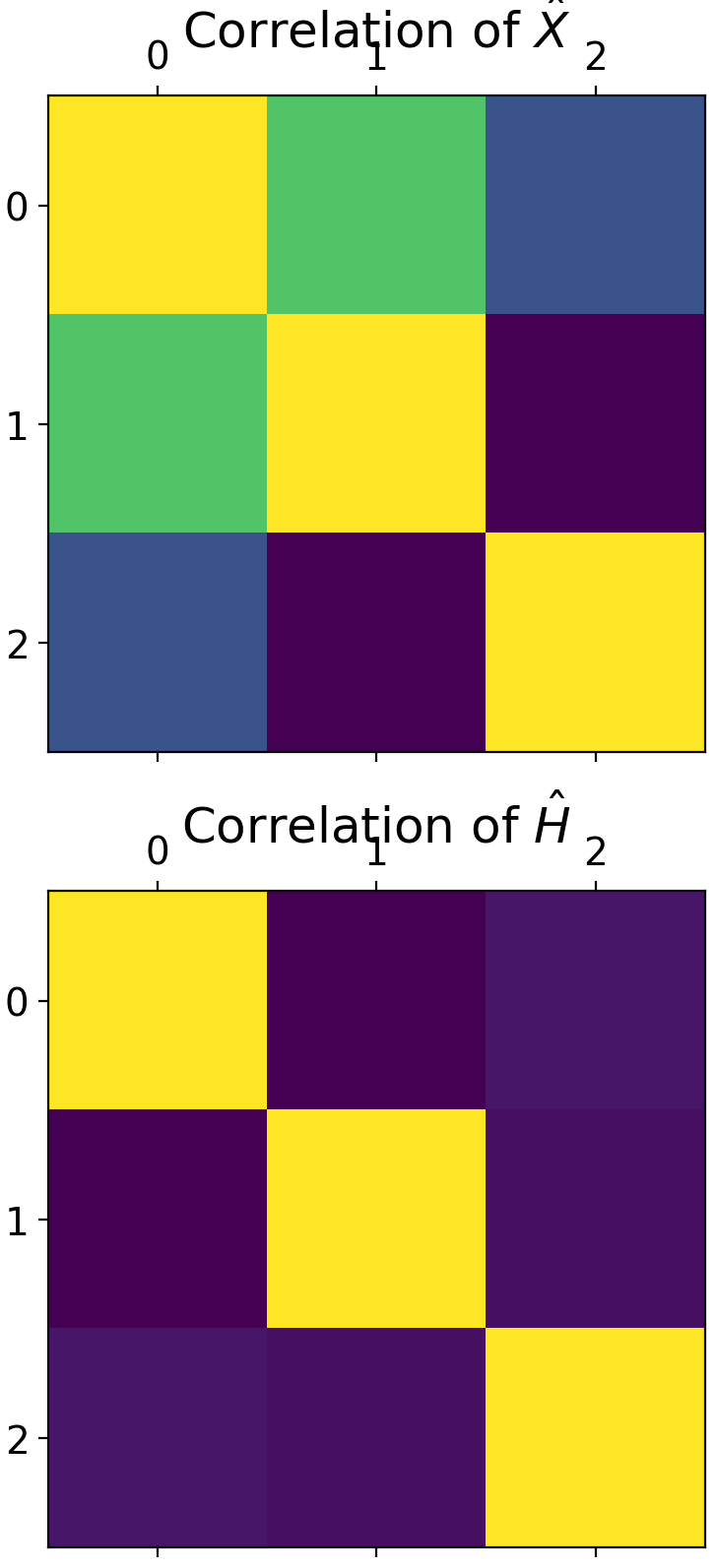}
    \subcaption{L3Net}
\end{minipage}
\begin{minipage}[b]{0.23\linewidth}
    \includegraphics[width=\linewidth]{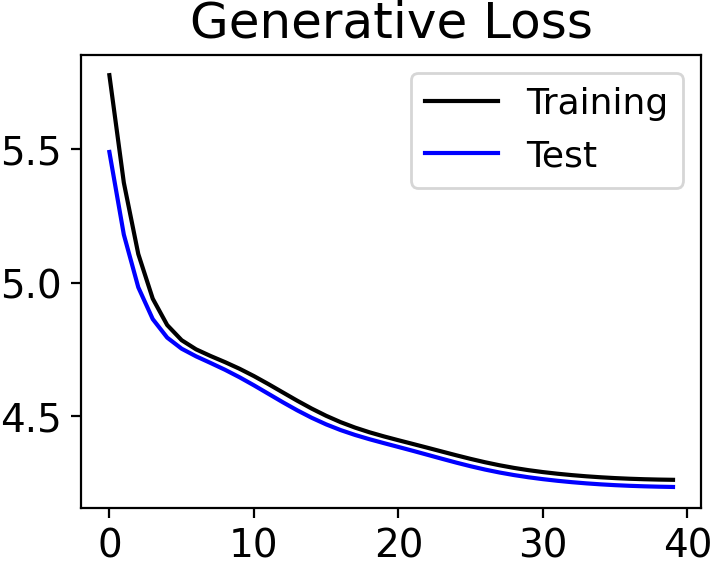}
    \includegraphics[width=\linewidth]{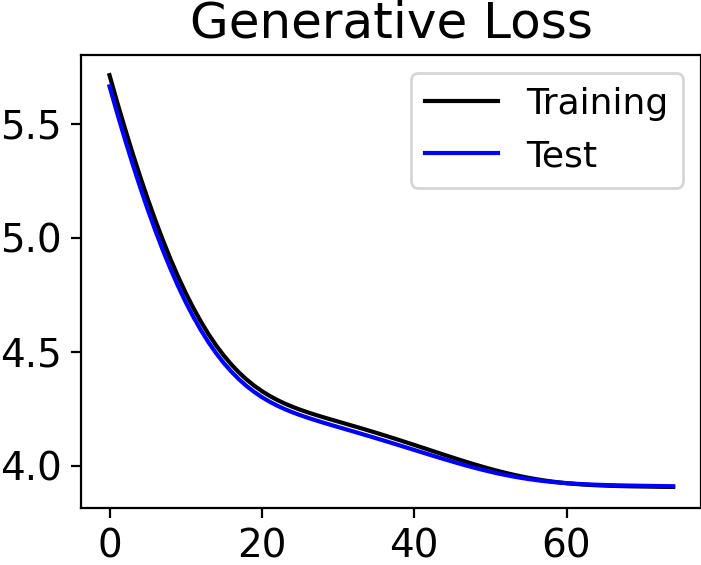}
      \subcaption{Chebnet (top) and L3net (bottom) losses}
\end{minipage}
\caption{
\rev{
Comparison of using spectral and spatial GNN layers in iGNN model 
for generating graph data as in Example \ref{ex:cannot_learn},
where the data samples are one-dimensional graph node features lying on a three-node graph,
$\rho=0.6$ and $\rho_1=-0.4$. 
We visualize correlation of samples generated by iGNN using the spectral GNN layer (ChebNet) in (b) and spatial GNN layer (L3Net) in (c),
in comparison with the ground truth in (a).
The losses over training epochs are shown in (d).
}}
\label{example_one_verify}
\end{figure}
\subsubsection{Construction of simulated graph data}

We first describe the construction of simulated graph data on small graphs (corresponding to Figures \ref{3node_nonconvex} and \ref{3node_convex_append}).
Each node has a binary label so that $Y\in \{0,1\}^3$ is a binary vector. Conditioning on a specific binary vector $Y$ out of the eight choices, the distribution of $X|Y$ is defined as 
\[X|Y:=P_A\left(Z|Y+\begin{bmatrix}
-4 & 0 & 4\\
0 & 0 & 0
\end{bmatrix}\right),\]
where $P_A:=D_A^{-1}A$ is the graph averaging matrix. 
\rev{The distribution of $Z|Y$ is specified as:
\begin{itemize}
    \item  In Figure \ref{3node_convex_append}, we let 
    \begin{equation}\label{mixture_small}
        Z_v|Y_v \sim \begin{cases} 
        \mathcal N((0,1.5)^T, 0.1I_2) & \text{if }Y_v=0 \\
        \mathcal N((0,-1.5)^T, 0.1I_2) & \text{if } Y_v=1
        \end{cases}.
    \end{equation}
    \item  In Figure \ref{3node_nonconvex}, $Z_v|Y_v$ are non-identically distributed over nodes $v=0,1,2$. When $v=1$, $Z_v|Y_v$ follow the same distribution as \eqref{mixture_small}. When $v=0 \text{ or } 2$, $Z_v|Y_v$ are rotated and shifted noisy two moons, whose positions depend on the binary value of $Y_v$. 
\end{itemize}
}

On large graphs (corresponding to Fig. \ref{large_graph_cond_gen}), each node has a binary label so that $Y\in \{0,1\}^{503}$ is a binary vector. Conditioning on a specific binary vector $Y$ out of the eight choices, the distribution of $X|Y$ is defined as
\[X|Y:=R((1-\delta)I+\delta P_A)Z|Y \text{ s.t. } Z_v|Y_v \sim \begin{cases} 
\mathcal N((0,12)^T, I_2) & \text{if }Y_v=0 \\
\mathcal N((0,0)^T, I_2) & \text{if } Y_v=1
\end{cases},\]
where $R$ denotes a counter-clockwise rotation matrix for 90 degrees \rev{(applied node-wise to each two-dimensional nodal feature)}
and $P_A:=D_A^{-1}A$ is the graph averaging matrix\rev{, where $D_A$ is the degree matrix}. We choose $\delta=0.2$ so that a soft graph averaging is applied to the hidden variables $Z|Y$.

\subsection{Additional experimental results}\label{append:exp_result}

\begin{figure}[!t]
\centering
\begin{minipage}[b]{0.19\linewidth}
    \includegraphics[width=\linewidth]{ChebNet_Sigma_graph_large.png}
    \vspace{-0.15in}
    \subcaption{503-node graph}
    \label{Cheb_network}
    \includegraphics[width=\linewidth]{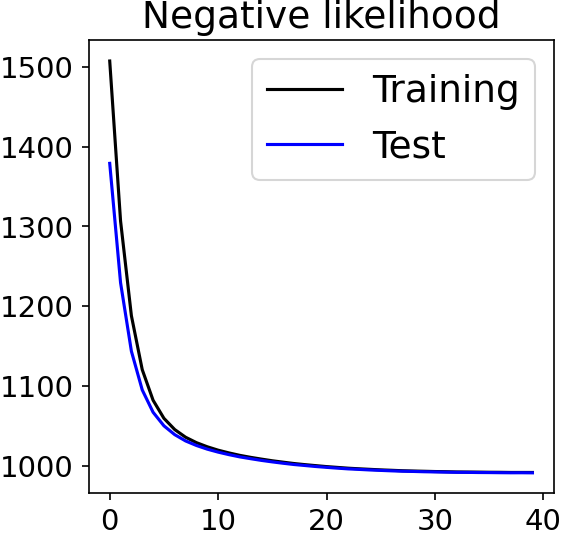}
    \vspace{-0.15in}
    \subcaption{iGNN loss}
    \label{Cheb_loss}
\end{minipage}
\vspace{0.15in}
\begin{minipage}[b]{0.79\linewidth}
    \includegraphics[width=\linewidth]{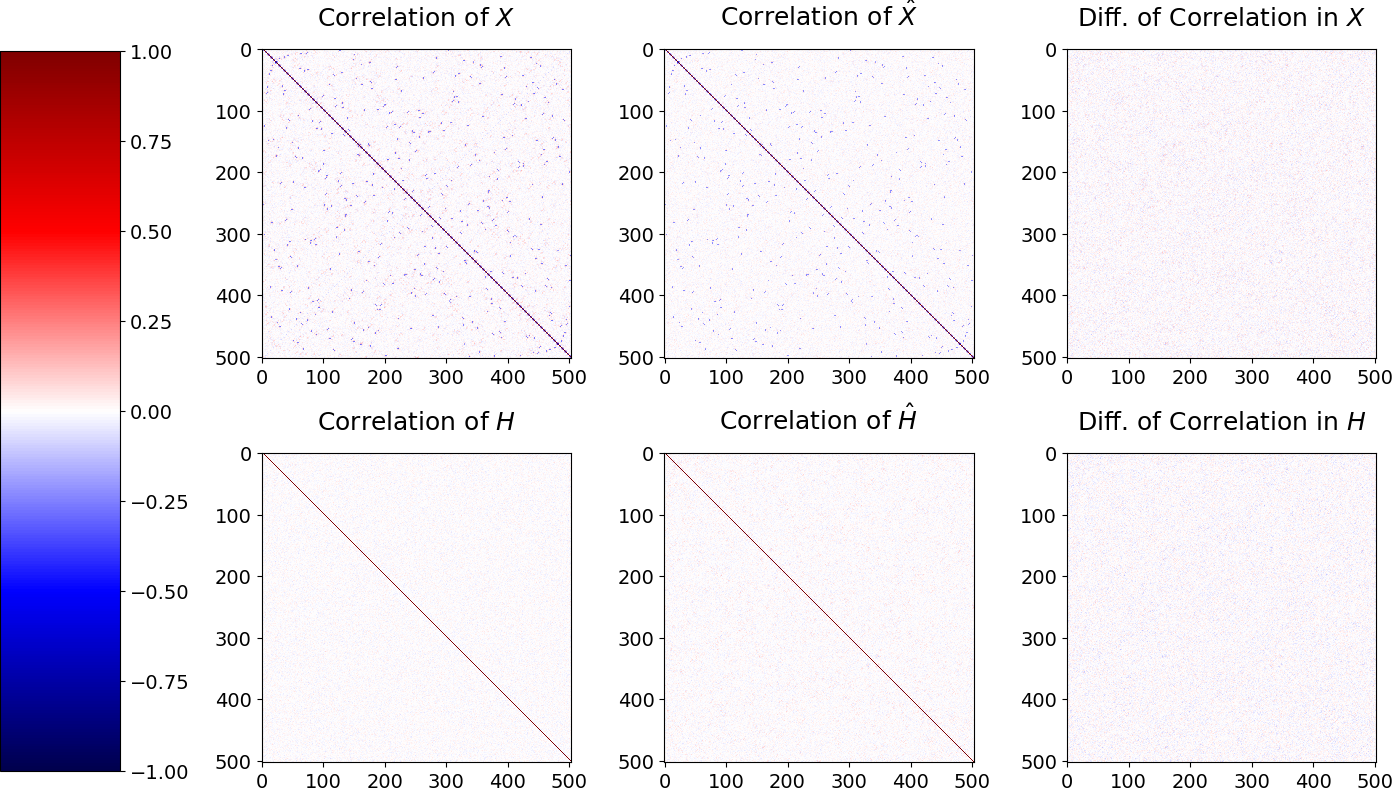}
    \vspace{-0.15in}
    \subcaption{True correlation matrices (left column), learned correlation matrices (middle column), and difference in true and learned correlation matrices (right column)}
    \label{Cheb_generation}
\end{minipage}
\vspace{-0.15in}
\caption{
Unconditional generation performance on graph data by iGNN using spectral GNN (ChebNet). 
Data $X$ lie on a 503-node chordal cycle graph 
\rev{and}
the node feature dimension $d' = 1$. To examine the generation quality, we plot the covariance matrix of model generated data $\hat{X}$ in comparison with that under the ground truth data $X$.
}
\end{figure}

\begin{figure}[!t]
\centering
\begin{minipage}[b]{0.19\linewidth}
    \includegraphics[width=\linewidth]{ChebNet_Sigma_graph_large.png}
    \vspace{-0.15in}
    \subcaption{500-node graph}
    \includegraphics[width=\linewidth]{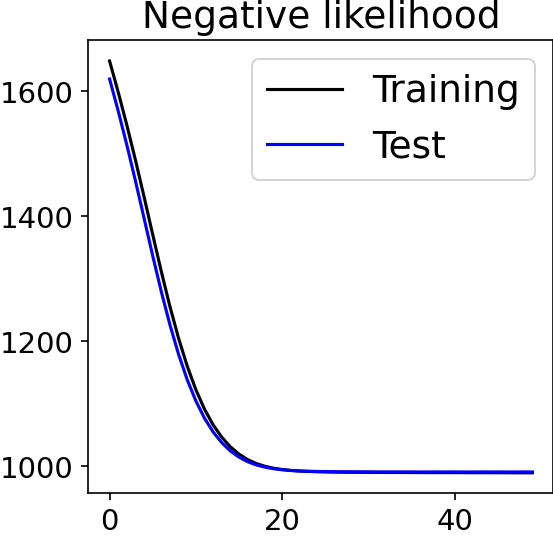}
    \vspace{-0.15in}
    \subcaption{iGNN loss}
    \label{Local_loss}
\end{minipage}
\vspace{0.15in}
\begin{minipage}[b]{0.79\linewidth}
    \includegraphics[width=\linewidth]{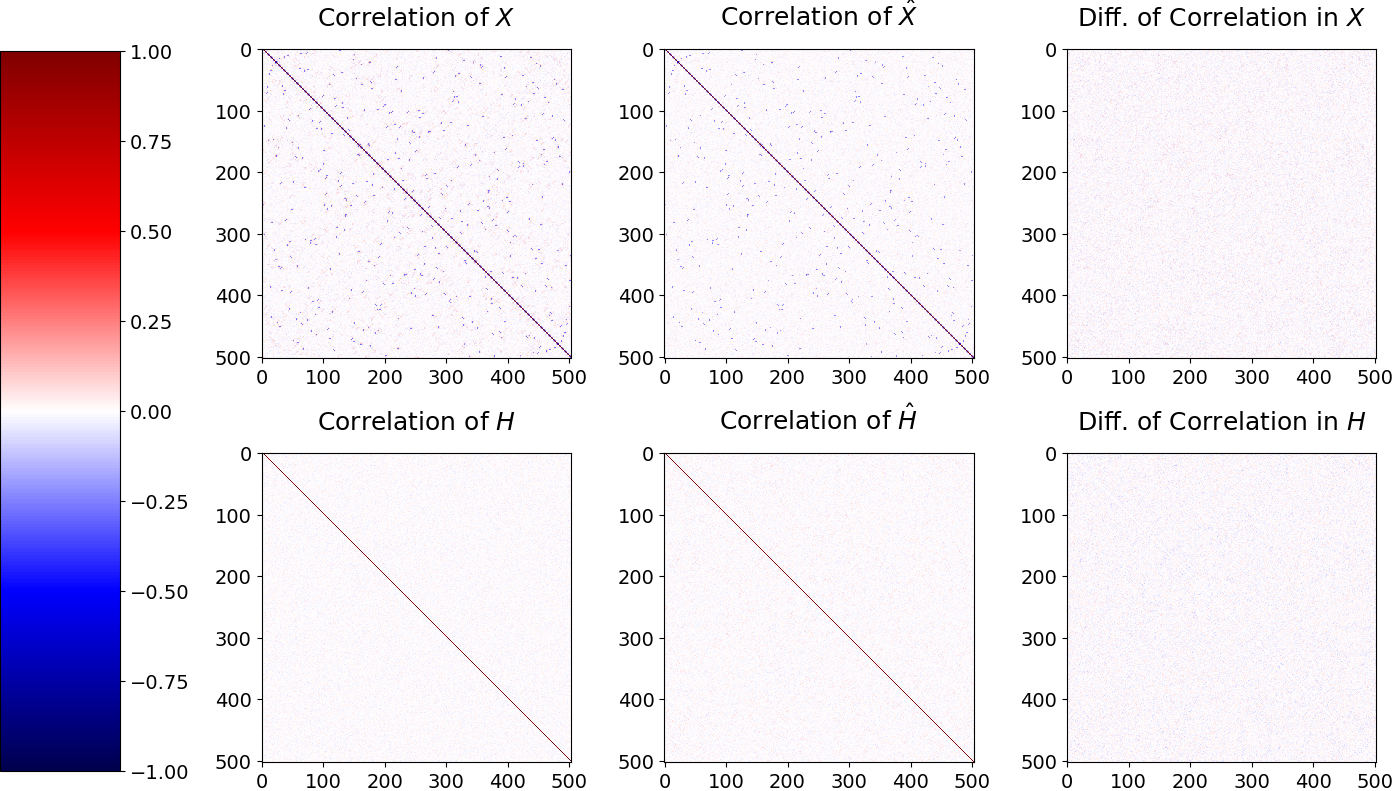}
    \vspace{-0.15in}
    \subcaption{True correlation matrices (left column), learned correlation matrices (middle column), and difference in true and learned correlation matrices (right column)}
    \label{Local_generation}
\end{minipage}
\vspace{-0.15in}
\caption{
Unconditional generation performance on graph data by iGNN using spatial GNN (L3Net). Data $X$ lie on a 503-node chordal cycle graph with spatial-based covariance $\Sigma$, where the node feature dimension $d' = 1$. To examine the generation quality, we plot the covariance matrix of model generated data $\hat{X}$ in comparison with that under the ground truth data $X$.}
\end{figure}

This subsection contains experimental results to augment those in the main text. In particular,
\begin{itemize}
    \itemsep 0em
    \item We show the confidence region on the three-moon non-graph conditional generation data in Fig. \ref{three_moon-confidence}. The generative quality is presented in Fig. \ref{three_moon} in the main text.
    \item We compare the generative quality of iGNN and \ciNNI\ on simulated three-node graph conditional generation data in Fig. \ref{3node_nonconvex}, to augment the results in Section \ref{sec:simple_example}.
    \item \rev{We numerically verify Example \ref{ex:cannot_learn} based on simulated three-node graph unconditional generation data. The generative results are presented in Fig. \ref{example_one_verify} for iGNN using Chebnet and L3net.}
    \item We demonstrate the unconditional generation results of iGNN using GNN layers in Fig. \ref{Cheb_generation} (spectral GNN) and Fig. \ref{Local_generation} (spatial GNN). Empirical results are presented in Section \ref{sec:theory_connect} and theoretical analyses are presented in Section \ref{sec:expressiveness}.
\end{itemize}

\subsection{Hyperparameter selection}\label{append_hyperparam}

We first summarize the hyperparameters used in our experiments and then verify that iGNN is insensitive to alternative hyperparameter choices.
\begin{itemize}
    \item Mixture distribution in $\R^2$ with 8 components (cf. Fig. \ref{8_gaussian}): we let iGNN contain 40 ResNet blocks, each of which is built with fully-connected layers with two hidden layers. We fix the learning rate at 5e-4 and train with a batch size of 1000. We fix the $\gamma$ regularization factor of the $W_2$ loss as 1.
    \item Conditional generation on small graph  (cf. \rev{Fig. \ref{3node_convex_append} and } Fig. \ref{3node_nonconvex}): we let iGNN contain 40 ResNet blocks, where the first hidden layer of each block is an L3Net layer. We fix the learning rate at 5e-4 and train with a batch size of 100. We fix the $\gamma$ regularization factor of the $W_2$ loss as 1.
    \item Conditional generation on large graph (cf. Fig. \ref{large_graph_cond_gen}): we let iGNN contain 5 ResNet blocks, where the first hidden layer of each block is an L3Net layer. We fix the learning rate at 1e-3 and train with a batch size of 100. We fix the $\gamma$ regularization factor of the $W_2$ loss as 0.375/503.
    \item Solar ramping event (cf. Fig. \ref{Solar_training}): 
    we let iGNN contain 40 ResNet blocks, where the first hidden layer of each block is a Chebnet layer. We fix the learning rate at 1e-4 and train with a batch size of 150. We fix the $\gamma$ regularization factor of the $W_2$ loss as 0.375/503.
    \item Traffic anomaly detection (cf. Fig. \ref{traffic_training}): we let iGNN contain 40 ResNet blocks, where the first hidden layer of each block is an L3Net layer. We fix the learning rate at 1e-4 and train with a batch size of 200. We fix the $\gamma$ regularization factor of the $W_2$ loss as 1.
    \item Unconditional generation on large graph (cf. Fig. \ref{Cheb_generation} and Fig. \ref{Local_generation}): we let iGNN contain 40 ResNet blocks, each of which contains a ChebNet or an L3Net layer. We fix the learning rate at 5e-4 and train with a batch size of 400. We fix the $\gamma$ regularization factor of the $W_2$ loss as 1.
\end{itemize}

\begin{figure}[!t]
    \centering
    \begin{minipage}{0.45\linewidth}
        \includegraphics[width=\linewidth]{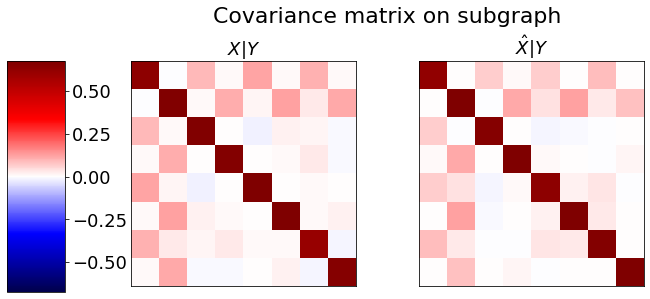}
        \subcaption{1-hop neighborhood of node 100 (4 nodes)}
    \end{minipage}
    \begin{minipage}{0.45\linewidth}
        \includegraphics[width=\linewidth]{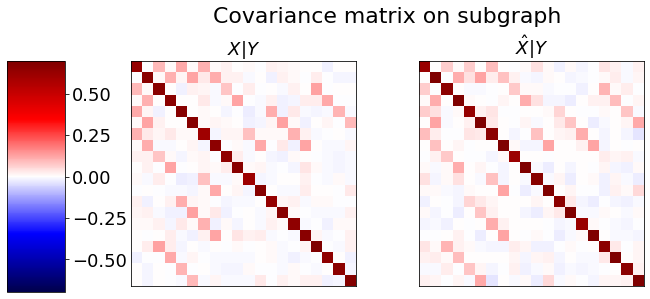}
        \subcaption{2-hop neighborhood of node 100 (10 nodes)}
    \end{minipage}
    \caption{Insensitivity of iGNN to the choice of $\gamma$ on conditional generation of large graph data. We increase $\gamma$ by 503X and keep all other choices the same as we did in Fig. \ref{large_graph_cond_gen}.}
    \label{lagre_graph_cond_appendix}
\end{figure}

\begin{figure}[!t]

\centering
\begin{minipage}[b]{0.24\linewidth}
    \includegraphics[width=\linewidth]{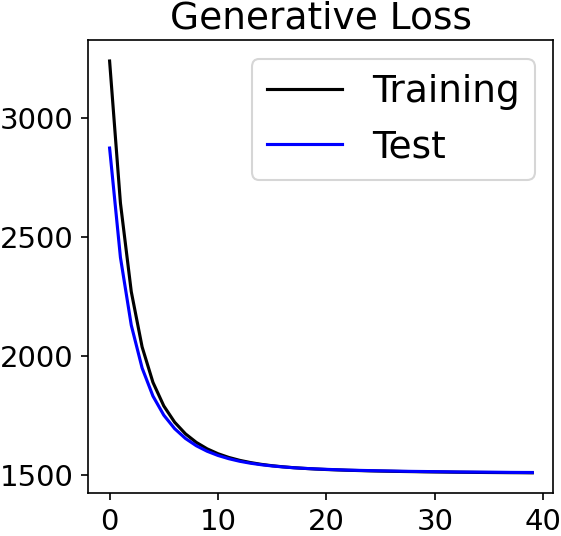}
    \subcaption{Increase $\gamma$ for $W_2$ regularization from 1 to 5}
\end{minipage}
\begin{minipage}[b]{0.24\linewidth}
    \includegraphics[width=\linewidth]{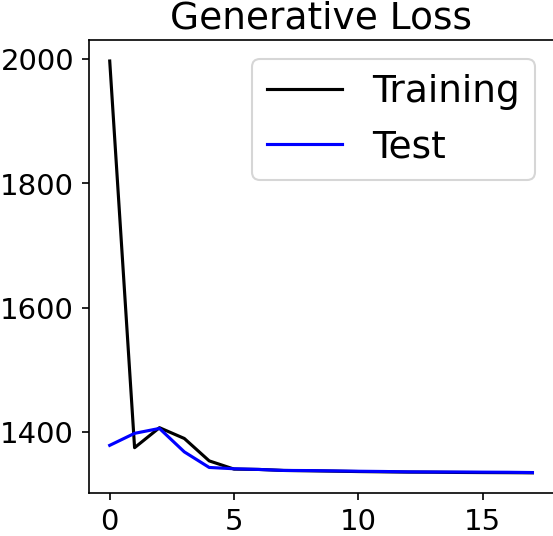}
    \subcaption{Increase the learning rate from 5e-4 to 5e-3}
\end{minipage}
\begin{minipage}[b]{0.24\linewidth}
    \includegraphics[width=\linewidth]{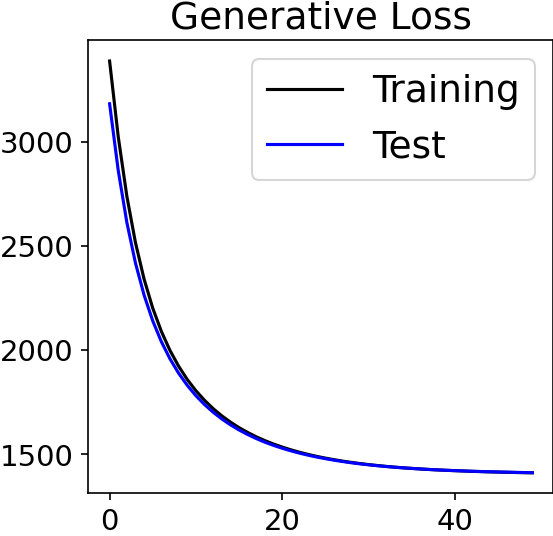}
    \subcaption{Reduce the number of blocks from 40 to 20}
\end{minipage}
\begin{minipage}[b]{0.24\linewidth}
    \includegraphics[width=\linewidth]{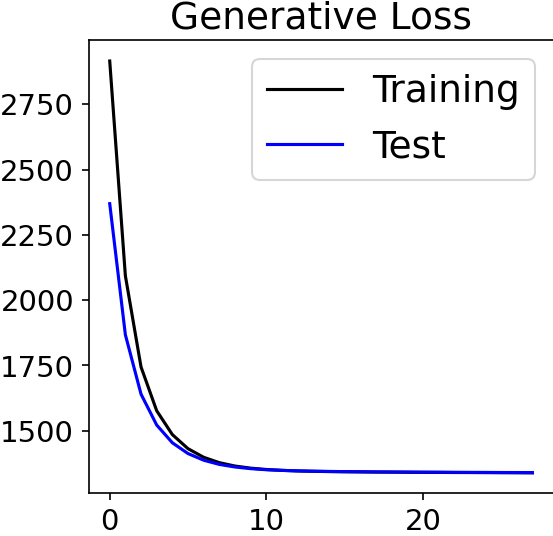}
    \subcaption{Reduce the batch size from 400 to 200}
\end{minipage}
\caption{Insensitivity of iGNN to the choice of various hyperparameters on unconditional generation of large graph data. We change one hyperparameter at a time and keep all other choices the same as we did in Fig. \ref{Cheb_generation}.}
\label{spectral_Cheb_alternative_hyperparam}
\end{figure}

\noindent \textit{Insensitivity of iGNN to $\gamma$ for $W_2$ regularization.} We verify that as long as the ResNet is invertible under the choice of $\gamma$, the value of $\gamma$ only affects the training efficiency but not the final generative quality. This is because larger $\gamma$ values restrict more the amount of movement by the ResNet, which thus potentially takes longer training epochs before transporting the distribution $X|Y$ to $H|Y$ at each $Y$. Fig. \ref{lagre_graph_cond_appendix} shows iGNN performance on conditional generation of data on large graph, where we increase the $\gamma$ factor by 503 times. The performance is nearly identical to that in Fig. \ref{large_graph_cond_gen}. On the other hand, Fig. \ref{spectral_Cheb_alternative_hyperparam} visualizes the generative loss of iGNN on unconditional generation of data on large graph, where we increase the $\gamma$ factor by 5 times.
Comparing to the generative loss in Fig. \ref{Cheb_loss}, the loss takes more epochs to converge, but eventual generative performance are nearly identical.

\vspace{0.1in}
\noindent \textit{Insensitivity of iGNN to other hyperparameters.} We verify that iGNN is insensitive to other hyperparameter choices such as learning rate, batch size, and the number of residual blocks. Figures \ref{spectral_Cheb_alternative_hyperparam} shows the loss trajectories of iGNN under alternative choices of these hyperparameters, where the setup is identical to that in Fig. \ref{Cheb_generation}. We see that training losses under these choices all converge reasonably fast. We omit comparing the covariance matrices because the generative quality barely differs from those in Fig. \ref{Cheb_generation}.

\end{document}